\documentclass[10pt]{article}
\usepackage[margin=1in]{geometry}

\usepackage{amsmath,amsthm,amssymb}
\usepackage{commath,mathtools}
\usepackage{algorithm,algorithmic}
\usepackage{hyperref}
\usepackage{color}
\usepackage{diagbox}
\usepackage{bm}
\usepackage{graphicx}
\usepackage{subfigure}
\usepackage{booktabs}       
\usepackage{appendix}
\usepackage{multirow}

\title{ParticleWNN: a Novel Neural Networks Framework for Solving Partial Differential Equations}
\author{Yaohua Zang
\footnote{School of Engineering and Design, Technische Universität München, München, Germany. Email: \texttt{yaohua.zang@tum.de}.}
\and
Gang Bao
\footnote{School of Mathematical Sciences, Zhejiang University, Hangzhou, Zhejiang, China. Email: \texttt{baog@zju.edu.cn}.}
}
\date{}

\begin{document}
\maketitle

\begin{abstract}
Deep neural networks (DNNs) have been widely used to solve partial differential equations (PDEs) 
in recent years. In this work, a novel deep learning-based framework named Particle Weak-form 
based Neural Networks (ParticleWNN) is developed for solving PDEs in the weak form. 
In this framework, the trial space is defined as the space of DNNs, while the test space 
consists of functions compactly supported in extremely small regions, centered around particles. 
To facilitate the training of neural networks, an R-adaptive strategy is designed to adaptively 
modify the radius of regions during training.
The ParticleWNN inherits the benefits of weak/variational formulation, requiring less regularity 
of the solution and a small number of quadrature points for computing integrals. Additionally, 
due to the special construction of the test functions, ParticleWNN enables parallel implementation 
and integral calculations only in extremely small regions. This framework is particularly desirable 
for solving problems with high-dimensional and complex domains. The efficiency and accuracy of 
ParticleWNN are demonstrated through several numerical examples, showcasing its superiority over 
state-of-the-art methods.
The source code for the numerical examples presented in this paper is available 
at \url{https://github.com/yaohua32/ParticleWNN}.
\bigskip

\smallskip
\end{abstract}

\section{Introduction}
\label{sec:introduction}
In recent decades, deep learning has emerged as a prominent field of study, 
capturing sustained attention from researchers.
Its successful application spans various domains, 
including natural language processing \cite{vaswani2017attention}, 
computer vision \cite{voulodimos2018deep}, and complex tasks like 
protein structure prediction \cite{jumper2021highly}. 
Notably, deep neural networks have gained significant traction 
in the field of numerical PDEs in recent years.
In comparison to traditional techniques like the finite element method (FEM), 
finite difference method (FDM), and finite volume method (FVM), 
DNN-based methods offer several advantages. 
These include their mesh-free nature, the capacity to surmount 
the curse of dimensionality \cite{han2018solving,sirignano2018dgm,zang2020weak}, 
robustness in handling noisy data \cite{bao2020numerical,both2021deepmod}, 
and intrinsic regularization properties \cite{yan2019robustness,mowlavi2023optimal}. 
Consequently, DNN-based approaches are particularly well-suited for tackling 
a wide array of PDEs and related problems, ranging from high-dimensional PDE problems 
to PDE-based optimal control and inverse problems.
Generally, the PDE-based forward and inverse problems are governed by the following PDE:
\begin{subequations}\label{eq:general}
\begin{align}
    \mathcal{A}[u(t,\bm{x});\bm{\gamma}] &=0,\quad \text{in}\ (0,T]\times\Omega, \label{eq:general_pde} \\
    \mathcal{B}[u(t,\bm{x});\bm{\gamma}] &= g(t,\bm{x}),\quad \text{on}\ [0,T]\times\partial\Omega, \label{eq:general_bd} \\
    \mathcal{I}[u(0,\bm{x});\bm{\gamma}] &= h(\bm{x}), \quad \text{on}\ \Omega, \label{eq:genreal_init}
\end{align}
\end{subequations}
where $\Omega$ is a bounded domain in $\mathbb{R}^d$ with boundary $\partial\Omega$, $T>0$, 
and $u(t,\bm{x})$ is the solution of the PDE.
In this context, $\mathcal{A}$ is the forward operator, representing the type of PDE, 
which can be parabolic, hyperbolic, or elliptic.
The operators $\mathcal{B}$ and $\mathcal{I}$ correspond to the boundary condition 
and initial condition, respectively.
The function $\bm{\gamma}$ denotes parameters in the PDE; for example, they could represent 
physical parameters, coefficients, or other relevant quantities.
The forward problem involves solving the PDE given the function $\bm{\gamma}$ and 
appropriate boundary conditions \eqref{eq:general_bd} and/or initial conditions \eqref{eq:genreal_init}. 
Conversely, the inverse problem aims to determine the function $\bm{\gamma}$ from 
additional boundary or interior measurements of the solution.

Among the existing DNN-based PDE solvers, one popular framework is based on the strong form of PDEs.
The most notable one is the physics-informed neural networks (PINN) \cite{raissi2019physics}. 
PINN approximates the PDE solutions with DNNs and formulates the loss function as a combination of 
the strong-form PDE residuals and data mismatch.
However, the strong-form methods usually require a massive amount of collocation points, 
leading to a high training cost \cite{jagtap2021extended}. 
Moreover, the non-convex optimization problem obtained by strong-form methods may 
result in many local minima, making learning complex physical phenomena more 
challenging\cite{jagtap2020conservative,krishnapriyan2021characterizing}. 
%
Different from the strong-form methods, the weak-form methods formulate the loss 
based on the weak/variational form of the PDEs, as demonstrated in various works 
\cite{yu2018deep,liao2019deep,li2019d3m,khodayi2020varnet,kharazmi2021hp,zang2020weak,
bao2020numerical,kharazmi2019variational}.
This approach offers advantages suh as requiring less smoothness in the solution, 
a reduced number of quadrature points for  integral computations, and the ability 
to facilitate local learning through domain decomposition \cite{kharazmi2019variational}.
Usually, these methods use DNNs to parameterize the trial functions.
The difference lies in choosing different types of test functions, including global 
or piecewise polynomials \cite{kharazmi2019variational,khodayi2020varnet}, 
localized non-overlapping high-order polynomials \cite{kharazmi2021hp}, 
and DNNs \cite{zang2020weak,bao2020numerical}.
However, existing weak-form methods define the test functions in the whole domain or 
sub-domain, which requires accurate integration methods and a large number of integration 
points to reduce integral approximation errors.
Although domain decomposition \cite{li2019d3m,kharazmi2021hp} can alleviate the problem in 
low-dimensional cases, high-dimensionality makes domain decomposition much more challenging.
%

In this paper, we introduce ParticleWNN, a novel deep learning-based framework designed 
for solving PDEs based on the weak form.
In this approach, the trial space is defined using DNNs, while the test space is 
constructed from functions compactly supported in exceedingly small regions, 
with particles as their centers, serving as local support centers for the test functions.
Constructing test functions in this manner offers several advantages. 
Firstly, the locally defined test functions allow the local training of DNNs and 
enable parallel implementation.
Secondly, since the particles are randomly sampled in the domain and the radius can be 
extremely small, the proposed method is desirable for high-dimensional and complex domain 
problems, and avoids the need to calculate integrals over large domains.
To enhance the training of the neural network model, we also introduce an R-adaptive strategy, 
dynamically reducing the upper bound of the compact support radius with each iteration.
The efficacy of ParticleWNN is verified through a series of numerical examples. 
Our primary contributions can be summarized as follows:
\begin{itemize}
    \item We propose a novel weak-form framework for solving PDEs by using DNNs, where the 
    test space is constructed by functions compactly supported in extremely small regions 
    whose centers are particles. 
    \item We develop several training techniques for the proposed framework to improve 
    its convergence and accuracy.
    \item We demonstrate the efficiency of the proposed framework through numerical 
    experiments, showcasing its superiority over state-of-the-art methods, particularly 
    in addressing complex problems like PDEs with intricate solutions and 
    inverse problems involving noisy data.
\end{itemize}
The remainder of the paper is organized as follows.
The related works are discussed in Section \ref{sec:related_works}.
In Section \ref{sec:method}, the ParticleWNN method is proposed for 
solving PDEs based on the weak formulation with deep neural networks.
Section \ref{sec:train} introduces some important implementation details 
of the ParticleWNN method and techniques to improve algorithm training, 
and summarizes the algorithm.
In Section \ref{sec:experiments}, the accuracy and efficiency of the 
proposed method are examined through a series of numerical experiments.
Finally, conclusions and discussions are drawn in Section \ref{sec:conclusion}.
\section{Related Works}
\label{sec:related_works}
According to different frameworks, deep learning-based PDE methods can be roughly divided 
into four categories: the strong-form methods, the weak-form methods, the hybrid methods 
of traditional methods and deep learning, and methods based on neural operators.
Typically, the strong-form methods utilize DNNs to approximate the solution of the equation.
Collocation points are generated on the domain and the domain boundary to evaluate 
the strong-from residuals of the governing equations in the domain and the mismatches of 
solutions on the boundary, respectively.
Then, the network parameters are learned by minimizing the weighted combination of 
residuals and mismatches.
Representative methods of this type include the deep Galerkin method (DGM) 
\cite{sirignano2018dgm} and the physical-informed neural networks (PINN) method 
\cite{raissi2019physics}.
Building upon the PINN framework, many methods have been proposed to address 
specific PDEs or enhance PINN's capabilities.
These include methods for solving fractional advection-diffusion 
equations (fPINN) \cite{pang2019fpinns}, 
handling Navier-Stokes equations (NSFnets) \cite{jin2021nsfnets}, 
addressing PDE-based inverse problems (B-PINN) \cite{yang2021b}, 
and dealing with PDEs in irregular domain (PhyGeoNet) \cite{gao2021phygeonet};
Various improvements to PINN have also been suggested,
such as adaptive collocation points strategies \cite{anitescu2019artificial,lu2021deepxde}, 
domain decomposition techniques \cite{jagtap2020conservative,jagtap2021extended},
loss modification approaches \cite{patel2022thermodynamically,yu2022gradient}, 
and new training strategies \cite{krishnapriyan2021characterizing}.
For additional information on strong-from methods, refer to 
\cite{karniadakis2021physics}, which provides a comprehensive exploration of these techniques

The weak-form methods, on the other hand, 
formulate the loss based on the weak or variational form of the PDEs.
Based on the variational principle, the Deep Ritz method (DeepRitz) \cite{yu2018deep} solves PDEs
by minimizing an energy function, which was further extended to PDEs with essential boundary
conditions in the deep Nitsche method \cite{liao2019deep}.
In \cite{li2019d3m}, a deep domain decomposition method (D3M) was proposed based on the variational
principle.
In \cite{zang2020weak,bao2020numerical}, the weak adversarial network (WAN) was proposed 
based on the weak form to convert the problem into an operator norm minimization 
problem.
The variational neural networks (VarNet) method \cite{khodayi2020varnet} is 
based on the Petrov-Galerkin method. In this method, the trial function is approximated by DNN, 
and the test functions are a set of piecewise polynomials that have compact support in the domain. 
The variational formulation of PINN (VPINN) \cite{kharazmi2019variational} 
has a similar formulation, except that the test functions are chosen to be 
polynomials with globally compact support over the domain.
Combined with domain decomposition, the hp-VPINN \cite{kharazmi2021hp} was developed
based on the sub-domain Petrov-Galerkin method. It divides the domain into 
multiple non-overlapping subdomains (h-refinement) and sets test functions as 
localized non-overlapping high-order polynomials (p-refinement).

The hybrid methods combine traditional methods with DNN-based methods to overcome 
the shortcomings of pure DNN-based methods.
In \cite{tompson2017accelerating}, the classical Galerkin method was combined 
with neural networks to solve initial boundary value problems.
In \cite{ranade2021discretizationnet}, a DiscretizationNet method was developed to combine 
the finite volume discretization with a generative CNN-based encoder-decoder 
for solving the Navier-Stokes equations.
The coupled automatic-numerical differentiation PINN (CAN-PINN) was developed in \cite{chiu2022can}, 
where the numerical differentiation-inspired method was coupled with the 
automatic differentiation to define the loss function.
Furthermore, DNN-based methods were integrated with the finite difference method in 
\cite{ren2022phycrnet,wandel2020learning} 
and the finite element method in \cite{yao2020fea}.

%
Recent advancements in solving PDEs have seen the emergence of operator learning methods 
\cite{li2020neural,lu2021learning,li2020fourier,li2020multipole,cai2021deepm}, which
focus on learning mappings between function spaces.
These methods differ from other approaches discussed earlier, as they rely on 
high-precision numerical PDE methods, typically classical methods like the FEM
and FVM, to generate the training data. 
The advantage of these methods lies in providing rapid numerical solution predictions 
for a specific class of PDEs.
In \cite{lu2021learning}, the DeepONet method was proposed to learn continuous nonlinear operators.
It is composed of two sub-networks: a branch net for the input fields and a trunk net for 
the locations of the output field.
Based on the DeepONet, the DeepM\&Mnet framework was developed for simulating multiphysics and multiscale problems 
in \cite{cai2021deepm}.
In \cite{li2020neural}, the authors introduced the concept of the neural operator and instantiated it through graph kernel networks designed to
approximate mappings between infinite-dimensional spaces.
Inspired by the classical multipole methods, this method was then generalized to the multi-level case in \cite{li2020multipole} 
to capture interaction at all ranges.
In \cite{li2020fourier}, the Fourier Neural Operators (FNOs) method was proposed using convolution-based integral kernels
within neural operators, where kernels can be efficiently evaluated in the Fourier space.

\section{Methodology}
\label{sec:method}
\subsection{The ParticleWNN framework}
To illustrate the proposed method, we consider the classical Poisson 
equation with the Dirichlet boundary condition:
\begin{equation}\label{eq:poisson}
    \begin{cases}
         -\Delta u(\bm{x})= f(\bm{x}), \quad \text{in}\  \Omega\subset \mathbb{R}^d, \\
        u(\bm{x}) = g(\bm{x}), \quad \text{on}\  \partial\Omega.
    \end{cases}
\end{equation}
The weak formulation of Poisson's equation \eqref{eq:poisson} involves finding a function in 
$\{u\in H^1(\Omega)| u|_{\partial\Omega}=g\}$ such that, for all test functions $\varphi\in H^1_0(\Omega)$, 
the following equation holds:
\begin{equation}\label{eq:weak_form}
    \int_{\Omega} \nabla u\cdot \nabla \varphi\ d\bm{x} = \int_{\Omega}f\varphi\ d\bm{x}, 
\end{equation}
where $H^1(\Omega)$ denotes the Sobolev space of functions with square-integrable derivatives 
and $H^1_0(\Omega)$ contains $H^1(\Omega)$ functions with zero boundary conditions.
Under appropriate conditions for $f$ and $g$, the weak form \eqref{eq:weak_form} allows a unique solution $u$, 
referred to as the \textit{weak solution} \cite{evans2022partial}.
Generally, weak-form DNN-based methods approximate the function $u$ with a 
neural network $u_{NN}(\bm{x};\theta)$, which usually consists of $l$ hidden layers with 
$\mathcal{N}_i$ neurons in each layer and activation function $\sigma(\cdot)$ that 
takes the following form:
\begin{equation}
u_{NN}(\bm{x};\theta) = T^{(l+1)}\circ T^{l}\circ T^{(l-1)}\circ\cdots\circ T^{(1)}(\bm{x}).
\end{equation}
Here, the linear mapping $T^{(l+1)}:\mathbb{R}^{\mathcal{N}_l}\rightarrow\mathbb{R}$ indicates the output layer, 
and $T^{(i)}(\cdot) =\sigma(\bm{W}_i\cdot +\bm{b}_i),\ i=1,\cdots l$ are nonlinear mappings 
with weights $\bm{W}_i\in\mathbb{R}^{\mathcal{N}_i\times\mathcal{N}_{i-1}}$ and 
biases $\bm{b}_i\in\mathbb{R}^{\mathcal{N}_i}$. The network parameters are collected in 
$\theta=\{\bm{W}_i, \bm{b}_i\}^{l+1}_{i=1}$.
Usually, the network is trained by minimizing a loss function, typically defined as 
the root mean square (RMS) or mean squared error (MSE) of the weak-form residuals, 
along with some penalty terms.
Following \cite{kharazmi2019variational,kharazmi2021hp}, we denote $\mathcal{R}$ as 
the weak-form residual in \eqref{eq:weak_form}:
\begin{equation}\label{eq:residual}
    \mathcal{R}(u_{NN};\varphi) = \int_{\Omega} \nabla u_{NN}\cdot \nabla \varphi\ d\bm{x}  - \int_{\Omega}f\varphi\ d\bm{x}.
\end{equation}
Different choices of the test functions vary in different weak-form methods.
For instance, the VarNet \cite{khodayi2020varnet} defines the test functions as a set of piecewise polynomials 
with compact support over the entire domain, the VPINN \cite{kharazmi2019variational} selects 
test functions as polynomials with globally compact support, the hp-VPINN \cite{kharazmi2021hp} employs 
localized non-overlapping high-order polynomials, and the WAN \cite{zang2020weak} 
represents the test space with DNNs.
The special feature of our work is choosing test functions to be compactly supported functions 
defined in small neighborhoods $B(\bm{x}^c, R)\subset \Omega$, where $\bm{x}^c$ is a particle in 
$\Omega$ and $R$ is the radius of the neighborhood.
This choice offers advantages such as enabling the DNN to focus on 
extremely local regions, avoiding integration over the entire domain, 
and facilitating parallelization.
Specifically, we choose the compactly supported radial basis functions (CSRBFs) as test functions in this work. 
Usually, the CSRBFs defined in $B(\bm{x}^c, R)$ have the following form:
\begin{equation}\label{eq:CSRBFs}
    \varphi(r) =
    \begin{cases}
        \varphi_{+}(r),\quad r(\bm{x})\leq 1, \\
        0,\quad r(\bm{x})>1,
    \end{cases}
\end{equation}
where $r(\bm{x}) = \|\bm{x}-\bm{x}^c\|_2/R$.
In fact, any functions in $H^1(\Omega)$ that are compactly supported in $B(\bm{x}^c, R)$ 
can be used as test functions.
To improve training efficiency, we use multiple test functions to formulate the loss function.
We generate $N_p$ particles $\{\bm{x}^c_i\}^{N_p}_i$ and the corresponding $\{R_i\}^{N_p}_i$ randomly or 
with predefined rules in the domain\footnote{To ensure that $B(\bm{x}^c, R)\subset \Omega$, we generate $R$ first, 
and then sample $\bm{x}^c$ in $\tilde{\Omega} = \{\bm{x}\in\Omega| \text{dist}(\bm{x},\partial\Omega)\geq R\}$.}, 
and then define $N_p$ CSRBFs $\{\varphi_i\}^{N_p}_i$ in each small neighbourhood $B(\bm{x}^c_i, R_i)$.
Therefore, we obtain the MSE of the weak-form residuals:
\begin{equation}\label{eq:loss_int}
\mathcal{L}_{\mathcal{R}} = \frac{1}{N_p}\sum^{N_p}_{i=1}|\mathcal{R}(u_{NN};\varphi_i)|^2.
\end{equation}
For the boundary condition (and/or initial condition), we can treat it as a penalty term:
\begin{equation}\label{eq:loss_bd}
\mathcal{L}_{\mathcal{B}} = \frac{1}{N_{bd}}\sum^{N_{bd}}_{j=1}|\mathcal{B}[u_{NN}(\bm{x}_j)] - g(\bm{x}_j)|^2,
\end{equation}
where $\{\bm{x}_j\}^{N_{bd}}_{j=1}$ are sampled points on the $\partial\Omega$.
Finally, we formulate our loss function as:
\begin{equation}\label{eq:loss}
\mathcal{L}(\theta) = \lambda_{\mathcal{R}}\mathcal{L}_{\mathcal{R}} + \lambda_{\mathcal{B}}\mathcal{L}_{\mathcal{B}},
\end{equation}
where $\lambda_{\mathcal{R}}$ and $\lambda_{\mathcal{B}}$ are weight coefficients in the loss function.

\subsection{Calculation of the loss}
To evaluate the loss in \eqref{eq:loss}, we need to calculate $N_p$ integrals that are defined 
in $B(\bm{x}^c_i, R_i),\ i=1,\cdots,N_p$.
A straightforward way to evaluate integrals is to use the Monte Carlo integration.
Unfortunately, it requires an immense sample size to ensure admissible integration errors.
An alternative is the quadrature rule method. This method works efficiently in 
low-dimensional case or when the integrand is simple.
However, in the case of high dimensional problems and complicated integrands, one needs to 
further increase the quadrature points, thus greatly increasing the computational cost.
Other numerical techniques, such as sparse grids \cite{novak1996high} and 
quasi-Monte Carlo integration \cite{morokoff1995quasi}, can also be employed.
In our framework, thanks to the special construction of the test functions, we only need 
to evaluate integrals in the small region $B(\bm{x}^c_i, R_i)$ rather than the entire domain.
As $u_{NN}$ is typically nearly constant within a small region, 
the integrand simplifies. 
Consequently, we can achieve similar accuracy with significantly fewer 
integration points compared to integrating over the entire domain or sub-domain.
By applying a simple coordinate transformation, we convert the calculation of integrals over 
$N_p$ small regions into computations over a standard region $B(\bm{0},1)$.
The coordinate transformation $\bm{x}=\bm{s}R_i+\bm{x}^c_i$ can be applied in 
\eqref{eq:residual}, resulting in:
\begin{equation}
\begin{aligned}
    \mathcal{R}(u_{NN};\varphi_i) 
    &= \int_{B(\bm{x}^c_i,R_i)} \nabla_{\bm{x}} u_{NN}\cdot \nabla_{r} \varphi_i \cdot \nabla_{\bm{x}} r  \ d\bm{x}  - \int_{B(\bm{x}^c, R_i)}f\varphi_i\ d\bm{x}, \\
    &= R^d_i\bigg(\int_{B(\bm{0},1)}\nabla_{\bm{x}} u_{NN}(\bm{x})\cdot \nabla_{r} \varphi_i \cdot \nabla_{\bm{x}} r \ d\bm{s} - \int_{B(\bm{0},1)}f(\bm{x})\varphi_i\ d\bm{s}\bigg),
\end{aligned} 
\end{equation}
where we use $\nabla_{\bm{x}} \varphi_i = \nabla_{r} \varphi_i \cdot \nabla_{\bm{x}} r$.
This allows us to generate one set of integration points in $B(\bm{0},1)$ to calculate $N_p$ integrals.
For example, assume that $\{\bm{s}_k\}^m_{k=1}$ are $K_{int}$ integration points generated from $B(\bm{0},1)$ and 
$\{w_k\}^{m}_{k=1}$ are the corresponding weights.
Then, we denote $\bm{x}^{(i)}_k := \bm{s}_k*R_i+\bm{x}^c_i$ and approximate $\mathcal{R}(u_{NN};\varphi_i)$ by:
\begin{equation}\label{eq:integral_approx}
\mathcal{R}(u_{NN};\varphi_i)\approx \frac{R_i^d\mathcal{V}_{B(\bm{0},1)}}{K_{int}}
    \sum^{K_{int}}_{k=1}w_k\bigg(\nabla_{\bm{x}} u_{NN}(\bm{x}^{(i)}_k)\cdot\nabla_r \varphi_i(\bm{x}^{(i)}_k)\cdot \nabla_{\bm{x}}r(\bm{x}^{(i)}_k) - 
    f(\bm{x}^{(i)}_k)\varphi_i(\bm{x}^{(i)}_k)
    \bigg),
\end{equation}
where $\mathcal{V}_{B(\bm{0},1)}$ indicates the volume of $B(\bm{0},1)$.
Consequently, the loss \eqref{eq:loss_int} can be approximated by:
\begin{equation}\label{eq:loss_approx_int}
\mathcal{L}_\mathcal{R}\approx \frac{\mathcal{V}^2_{B(\bm{0},1)}}{N_pK^2_{int}}\sum^{N_p}_{i=1}R_i^{2d}
\bigg(
    \sum^{K_{int}}_{k=1}w_k\bigg( \nabla_{\bm{x}} u_{NN}(\bm{x}^{(i)}_k)\cdot\nabla_r \varphi_i(\bm{x}^{(i)}_k)\cdot \nabla_{\bm{x}}r(\bm{x}^{(i)}_k) 
    -f(\bm{x}^{(i)}_k)\varphi_i(\bm{x}^{(i)}_k)\bigg)
\bigg)^2.
\end{equation}
From \eqref{eq:loss_approx_int}, we can see that the value of $\mathcal{L}_\mathcal{R}$ depends on $R^{2d}_i$.
To avoid the training failure, we remove $R_i^{2d}$ in \eqref{eq:loss_approx_int}.
We also remove the fixed term $\mathcal{V}^2_{B(\bm{0},1)}$, which does not affect the training of the model.
We denote $\tilde{\mathcal{L}}_\mathcal{R}$ as the approximation of 
$\mathcal{L}_\mathcal{R}$ with $R_i^{2d}$ and $\mathcal{V}^2_{B(\bm{0},1)}$ removed.
Then, the loss function \eqref{eq:loss} is modified and approximated as:
\begin{equation}\label{eq:loss_new}
\begin{aligned}
\tilde{\mathcal{L}}(\theta)
    &= \frac{\lambda_{\mathcal{R}}}{N_pK^2_{int}}\sum^{N_p}_{i=1}
    \bigg(
        \sum^{K_{int}}_{k=1}w_k\bigg(\nabla_{\bm{x}} u_{NN}(\bm{x}^{(i)}_k)\cdot\nabla_r \varphi_i(\bm{x}^{(i)}_k)\cdot \nabla_{\bm{x}}r(\bm{x}^{(i)}_k) 
        -f(\bm{x}^{(i)}_k)\varphi_i(\bm{x}^{(i)}_k)\bigg)
    \bigg)^2 \\
    &\quad + \frac{\lambda_{\mathcal{B}}}{N_{bd}}\sum^{N_{bd}}_{j=1}\bigg(\mathcal{B}[u_{NN}(\bm{x}_j)] - g(\bm{x}_j)\bigg)^2.
\end{aligned}
\end{equation}

\section{The Implementation Details}
\label{sec:train}
In this section, we discuss some implementation details and training techniques that 
will improve the performance of the proposed ParticleWNN.

\paragraph{Selection of test functions.} 
In our framework, while the construction of the compact support of the test functions is determined, 
the types of test functions themselves can be very diverse.
In fact, any functions in the $H^1_0$ space can be considered test functions 
providing an infinite number of possible choices.
For different integration methods, the selection of test functions have varying effects 
on the integration error, ultimately influencing the accuracy of the method.
In the numerical examples presented in this paper, we opt for meshgrid integration points 
and choose Compactly Supported Radial Basis Functions (CSRBFs) as test functions.
There are various types of CSRBFs to choose from, such as the Bump function \cite{fry2002smooth},
Wendland's function \cite{wendland1995piecewise}, Wu's function \cite{wu1995compactly},
Buhmann's function \cite{buhmann2001new}, among others. 
Typically, we consider the following Wendland's type CSRBFs:
\begin{equation}
    \phi_{d,2}(r) =
    \begin{cases}
         \frac{(1-r)^{l+2}}{3}[(l^2+4l+3)r^2+(3l+6)r+3], \quad r<1,\\
         0, \quad r\geq 1,
    \end{cases}
\end{equation}
where $l=\lfloor d/2\rfloor  + 3$, $d$ is the dimension of the domain, and $\lfloor\cdot\rfloor$ 
indicates the flooring function.
\paragraph{R-adaptive strategy.}
In the ParticleWNN framework, the parameter $R$ plays a crucial role in determining 
the size of the interest area of the network model $u_{NN}$.
If $R$ is too large, there will be increased overlaps between compact supports, 
leading to the loss becoming insensitive to changes in particles.
Additionally, a large $R$ often implies a more complex integrand, resulting in 
a larger approximation error for the integral.
However, $R$ should not be too small, as it may cause precision loss in floating-point numbers.
In this work, we dynamically generate $R_i$ for each particle $\bm{x}^c_i$ from a range $[R_{min}, R_{max}]$
where $R_{min}$ is a small fixed number and $R_{max}(\geq R_{min})$ is another small number 
that varies with iterations.
Generally, there are three common ways for $R_{max}$ to change with iterations:
\begin{itemize}
    \item \textbf{R-ascending}: $R_{max}$ gradually increases with iterations until it reaches a specified upper bound. 
    \item \textbf{R-fixed}: $R_{max}$ remains unchanged. 
    \item \textbf{R-descending}: $R_{max}$ gradually decreases with iterations until a given lower bound is reached.
\end{itemize}
In the experiment, we observed that the \textbf{R-descending} strategy yielded the best performance 
(refer to details in the appendix \ref{sec:app_R_adaptive}).
Therefore, we adopt the \textbf{R-descending} strategy for training the ParticleWNN.
\paragraph{Adaptive selection of particles.}
A common way to improve the training of ParticleWNN 
is to select the particles with some smart rules.
In this paper, we found that the $topK$ technique can improve the 
performance of the proposed ParticleWNN (see the appendix \ref{sec:app_topK} for details).
The implementation of the $topK$ technique is as follows: initially, we randomly sample $N_p$ 
particles in the domain. Subsequently, we evaluate the square residuals for each particle with 
\eqref{eq:integral_approx}. Finally,
we choose the $topK(\leq N_p)$ particles with the largest residuals to calculate the loss.

\paragraph{Summary of the algorithm.}
We summarize the ParticleWNN algorithm in Algorithm \ref{alg:ParticleWNN}.
It is worth noting that, although the implementation of parallelization 
is not included in Algorithm \ref{alg:ParticleWNN}, 
achieving parallelization is straightforward.
The steps involve distributing the computation of $N_p$ weak residuals 
and their gradients to $M \leq N_p$ machines for simultaneous processing, 
thus achieving parallelization.
\begin{algorithm}[t!]
\caption{The ParticleWNN Algorithm}
\label{alg:ParticleWNN}
\begin{algorithmic}[1]
\STATE \textbf{Input:} 
{$N_p$, $N_{bd}$, $K_{int}$, $R_{min}$, $R_{max}$, 
$\lambda_{\mathcal{R}}$, $\lambda_{\mathcal{B}}$, $topK$, learning rate $\tau_\theta$, and 
maximum iterations $maxIter$.}
\STATE \textbf{Initialize:} Network architecture $u_{NN}(\bm{x};\theta):\Omega \to \mathbb{R}$, where $\theta$ indicates network parameters.
\WHILE{iterations $<maxIter$ }
\STATE {Generate radius $\{R_i\}^{N_p}_{i=1} \sim \text{Unif}[R_{min}, R_{max}]$, particles $\{\bm{x}^c_i \in \tilde{\Omega} : i\in[N_p]\}$, 
data points $\{\bm{x}_j \in \partial \Omega : j\in[N_{bd}]\}$, and integral points $\{\bm{s}_k\in B(\bm{0},1):k\in[K_{int}]\}$.}
\STATE {Get integral points for each region $B(\bm{x}^c_i, R_i)$: $\bm{x}^{(i)}_k= \bm{s}_kR_i + \bm{x}^c_i, k\in[K_{int}], i\in[N_p]$.}
\STATE {Calculate residuals $\mathcal{R}(u_{NN};\varphi_i)$ with \eqref{eq:integral_approx} and evaluate the loss $\tilde{\mathcal{L}}(\theta)$ with $topK$ residuals.}
\STATE {Update network parameters with SGD: $\theta \leftarrow \theta - \tau_\theta \nabla_\theta \tilde{\mathcal{L}}(\theta)$.}
\ENDWHILE
\STATE \textbf{Output:} {The trained network $u_{NN}(\bm{x};\theta)$.}
\end{algorithmic}
\end{algorithm}
\section{Experiments}
\label{sec:experiments}
In this section, we present a series of numerical experiments to demonstrate 
the ParticleWNN method's effectiveness in overcoming the common challenge 
of requiring a large number of integration points in weak-form methods. 
We compare our proposed method with two widely recognized weak-form DNN-based approaches: 
the DeepRitz method and the VPINN method. 
Furthermore, we include a comparison with the strong-form method, PINN, to highlight 
ParticleWNN's relative advantages in solving problems with intricate solutions.
Since our paper primarily emphasizes methodological differences in solving PDEs and related problems, 
we limit our comparisons to the classic PINN method without special training techniques
\footnote{It's worth noting that most of these techniques are also applicable to the ParticleWNN method.}. 
In the following, we refer to the PINN method without special training techniques as 
the vanilla PINN method.

\paragraph{Experimental setups.} Without specific clarifications, we choose the trial function $u_{NN}$ as a ResNet with $4$ 
hidden layers and $50$ nodes in each layer, and provide the activations in specific problems.
During the training process, we randomly sample $N_p$ particles in the domain and $N_{bd} = 2d\tilde{N}_{bd}$ 
points ($\tilde{N}_{bd}$ for each of $2d$ sides) on the boundary.
we sample the radius $R_i$ from $[R_{min}, R_{max}]$ for each region, 
where $R_{max}$ decreases linearly with the number of iterations.
We set $R_{min}=10^{-6}$, $R_{max}=10^{-4}$, 
$\lambda_\mathcal{R}=1$, $\lambda_\mathcal{B}=5$, $maxIter=20000$, and 
generate $K_{int}$ integration points in $B(\bm{0},1)$ to evaluate integrals.
We adopt the Adam optimizer with $\tau_\theta=0.001$ and apply the StepLR scheduler with a step size of 
1 epoch and a $\gamma=1-2./maxIter$.
We use the Relative error $\|u_{NN}-u\|_2/\|u\|_2$ and the maximum absolute error (MAE) $\max |u_{NN}-u|$ 
as evaluation metrics and run each example with 5 random seeds to obtain 
the mean and the standard deviation.
For the 1D poisson problem \eqref{eq:poisson_1d}, all related experiments were executed using 
the Google Colab \cite{bisong2019google} with a Tesla T4 GPU.
For other problems, all experiments were executed using 
the Kaggle Notebook \cite{banachewicz2022kaggle} with a Tesla P100 GPU.
\subsection{The 1D Poisson's Equation}
\label{sec:poisson1d}
We first consider the following one-dimensional Poisson's equation:
\begin{equation}\label{eq:poisson_1d}
    \begin{cases}
    -\frac{\partial^2 u}{\partial x^2}(x) = f(x),\quad x\in \Omega=(-1,1), \\
    u(x) = g(x),\quad x\in \partial\Omega.
    \end{cases}
\end{equation}
We construct a solution $u(x)=x\cos(\omega x)$ for \eqref{eq:poisson_1d} and use it to calculate the 
forcing term $f(x)$ and boundary condition $g(x)$. 
Previous work \cite{basir2022critical} has shown that the vanilla PINN struggles to solve this problem 
with a Fully Connected Feedforward Neural Network using Tanh activation. 
Here, we opt for a composite function of Tanh and Sine as the activation in the network models.
We set $N_p=200$, $topK=150$, $K_{int}=50$, and $\tilde{N}_{bd}=1$ for the proposed method. 
For comparison, we generate $10000$ integration points in $\Omega$ 
for the DeepRitz method and $10000$ ($topK=7500$) collocation points for the vanilla PINN.
We set the number of integration points to $N_{int}=200$ and 
the number of test functions to $N_{test}=50$ for the VPINN method.
Other parameters maintain consistent settings.

\begin{figure}[!htbp]
    \centering  
    \subfigure[Relative error]{\label{fig:poisson1d_l2}
        \includegraphics[width=0.22\textwidth]{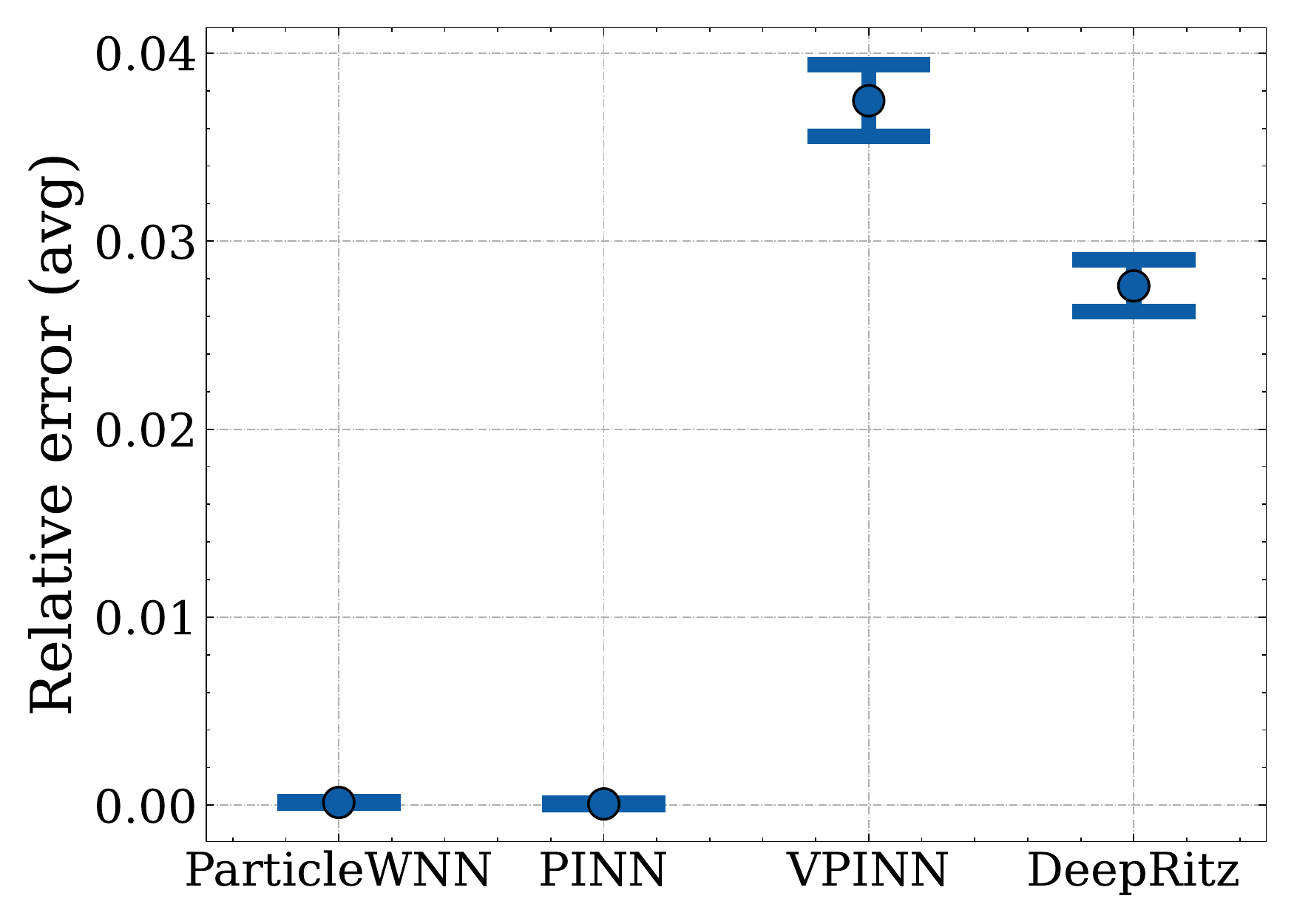}}
    \subfigure[MAE]{\label{fig:poisson1d_mae}
        \includegraphics[width=0.22\textwidth]{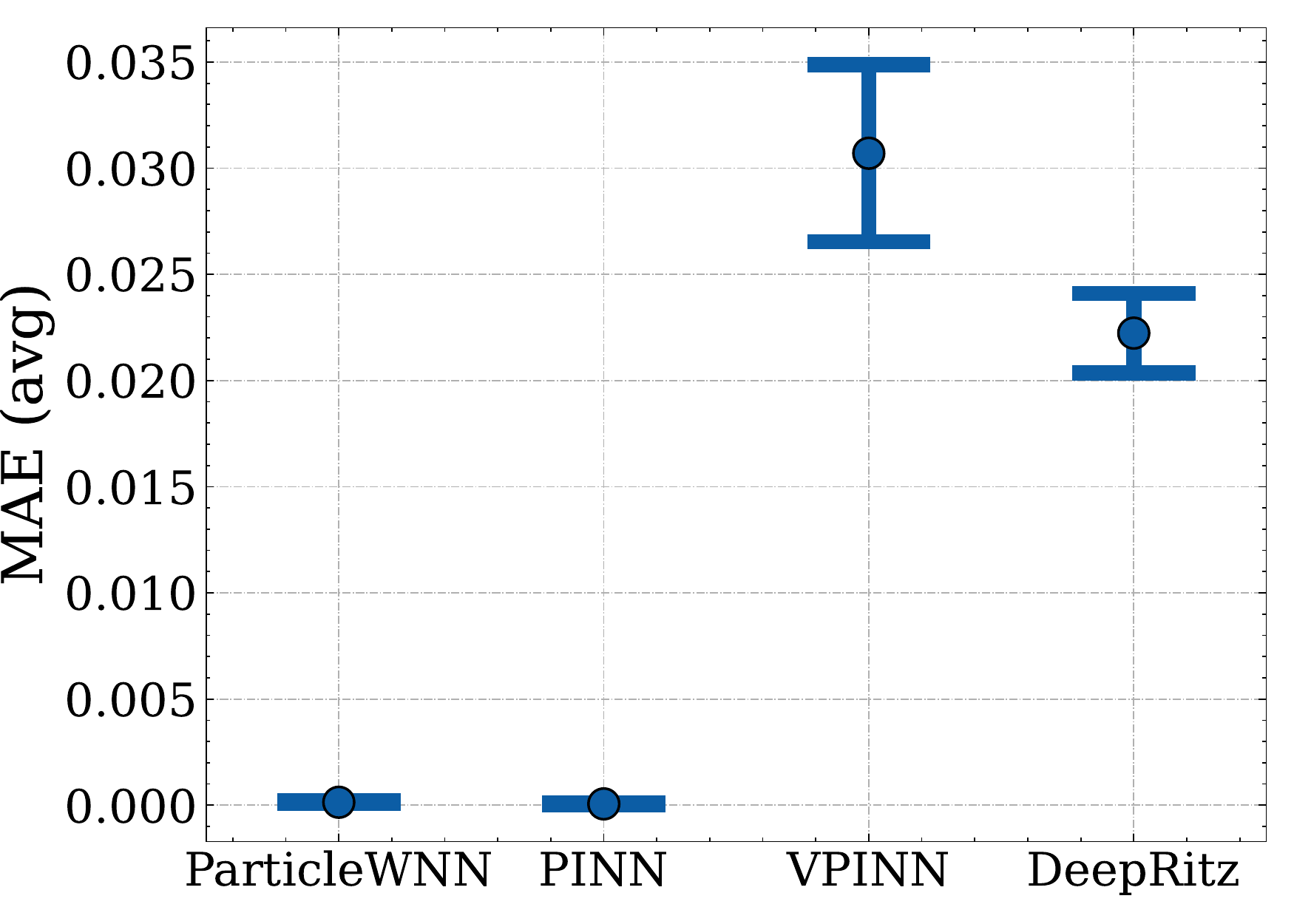}}
    \subfigure[Relative error vs. time]{\label{fig:poisson1d_time}
        \includegraphics[width=0.255\textwidth]{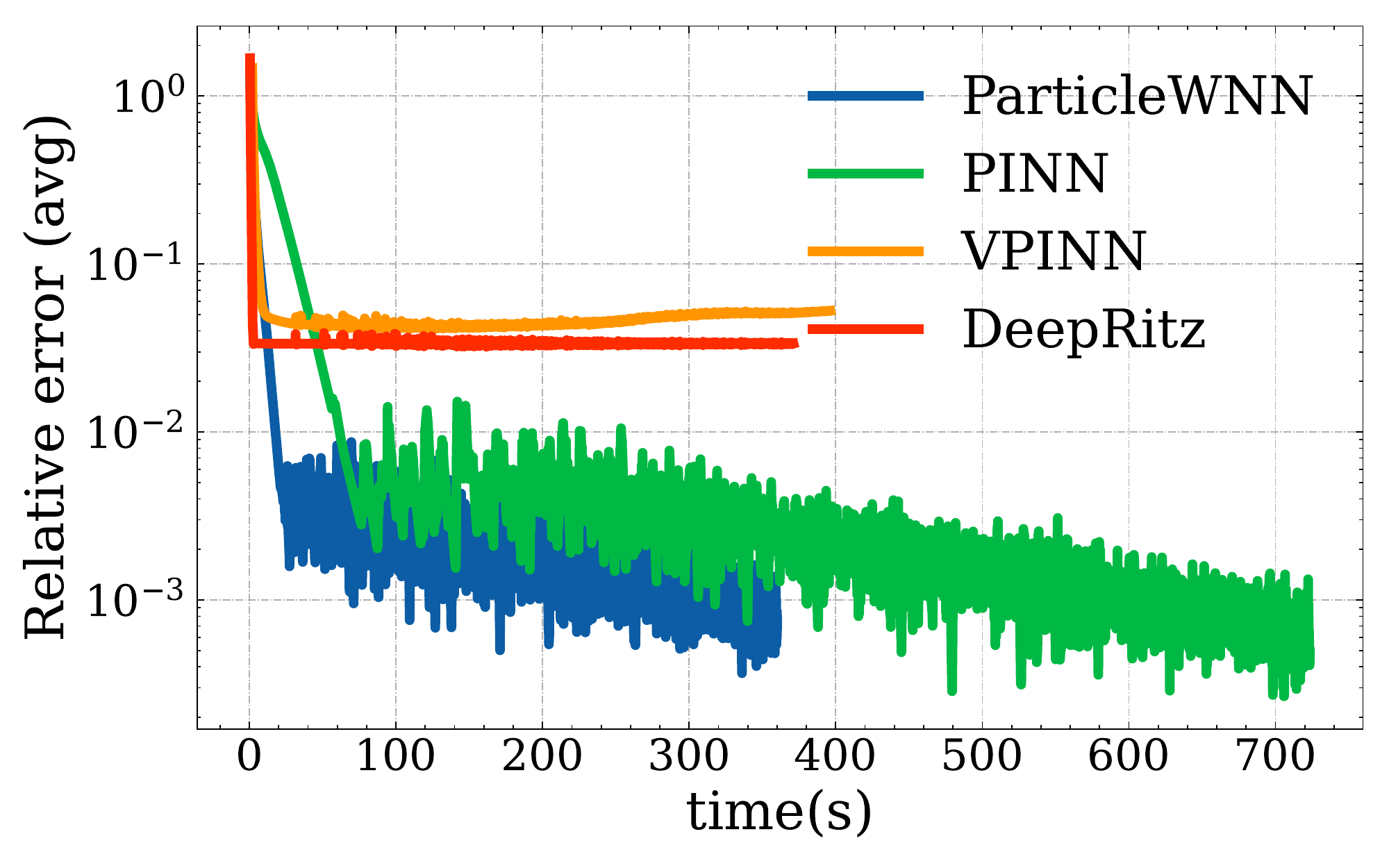}}
    \subfigure[Point-wise error]{\label{fig:poisson1d_abs}
        \includegraphics[width=0.255\textwidth]{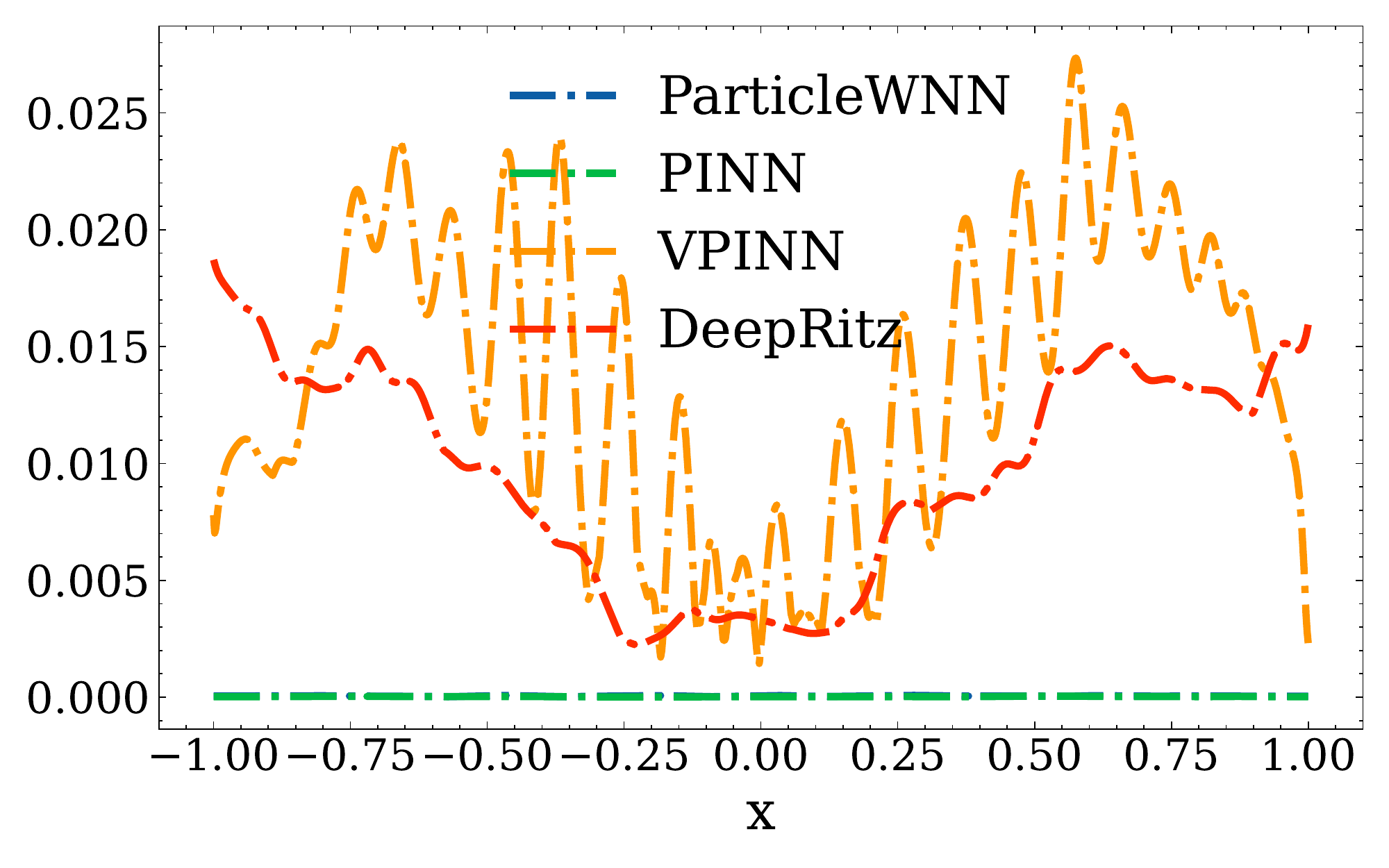}}
    \vspace{-0.25cm}
    \caption{The performance of different methods in solving problem \eqref{eq:poisson_1d} 
    with frequency $\omega=2\pi$. 
    (a) Average Relative errors; (b) Average MAEs; 
    (c) Average Relative errors vs. Average computation times;
    (d) Average point-wise errors. } 
    \label{fig:poisson_1d_low}
\end{figure}

\begin{figure}[!htbp]
    \centering  
    \subfigure[Relative error]{\label{fig:poisson1d_l2_high}
        \includegraphics[width=0.22\textwidth]{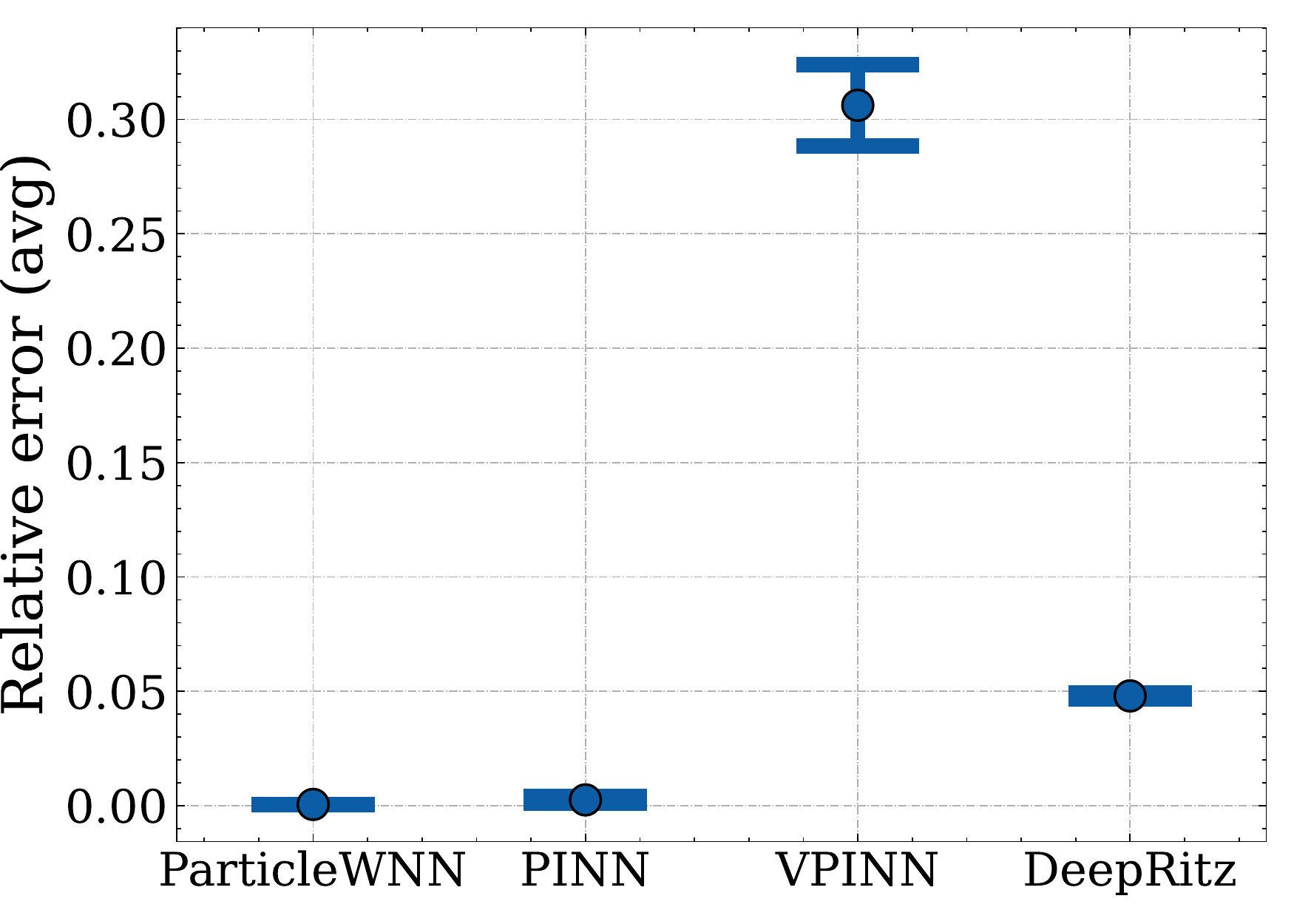}}
    \subfigure[MAE]{\label{fig:poisson1d_mae_high}
        \includegraphics[width=0.22\textwidth]{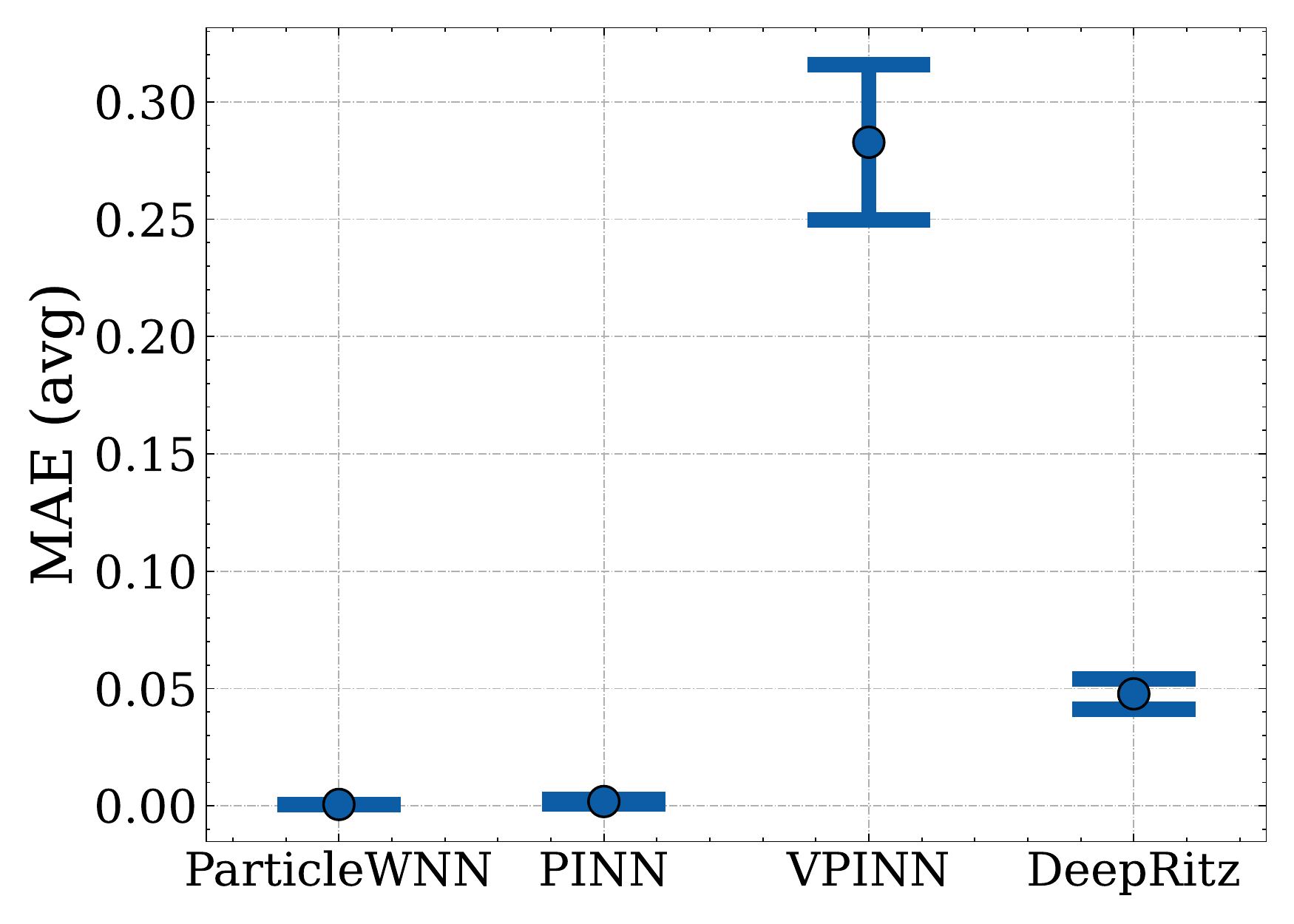}}
    \subfigure[Relative error vs. time]{\label{fig:poisson1d_time_high}
        \includegraphics[width=0.255\textwidth]{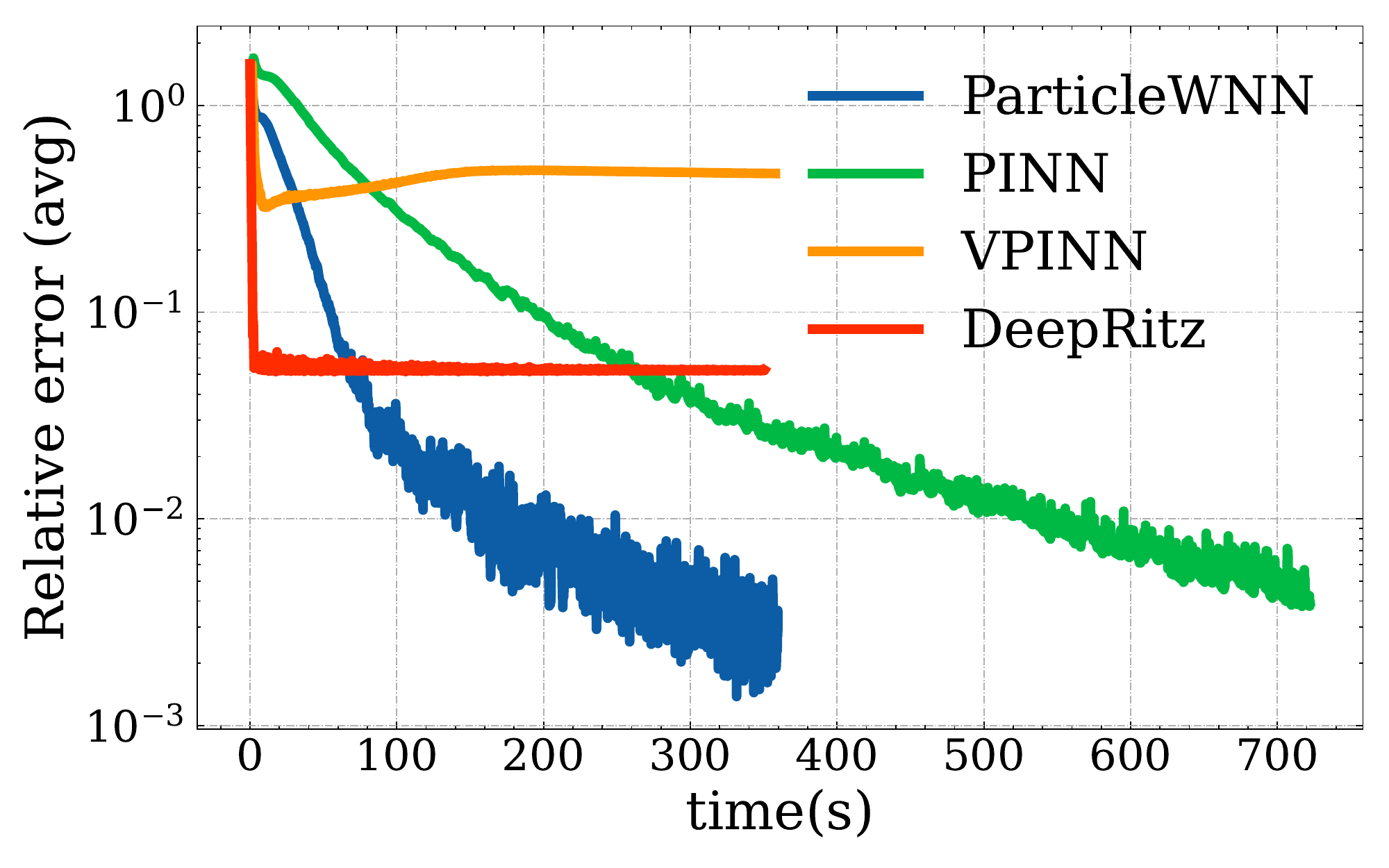}}
    \subfigure[Point-wise error]{\label{fig:poisson1d_abs_high}
        \includegraphics[width=0.255\textwidth]{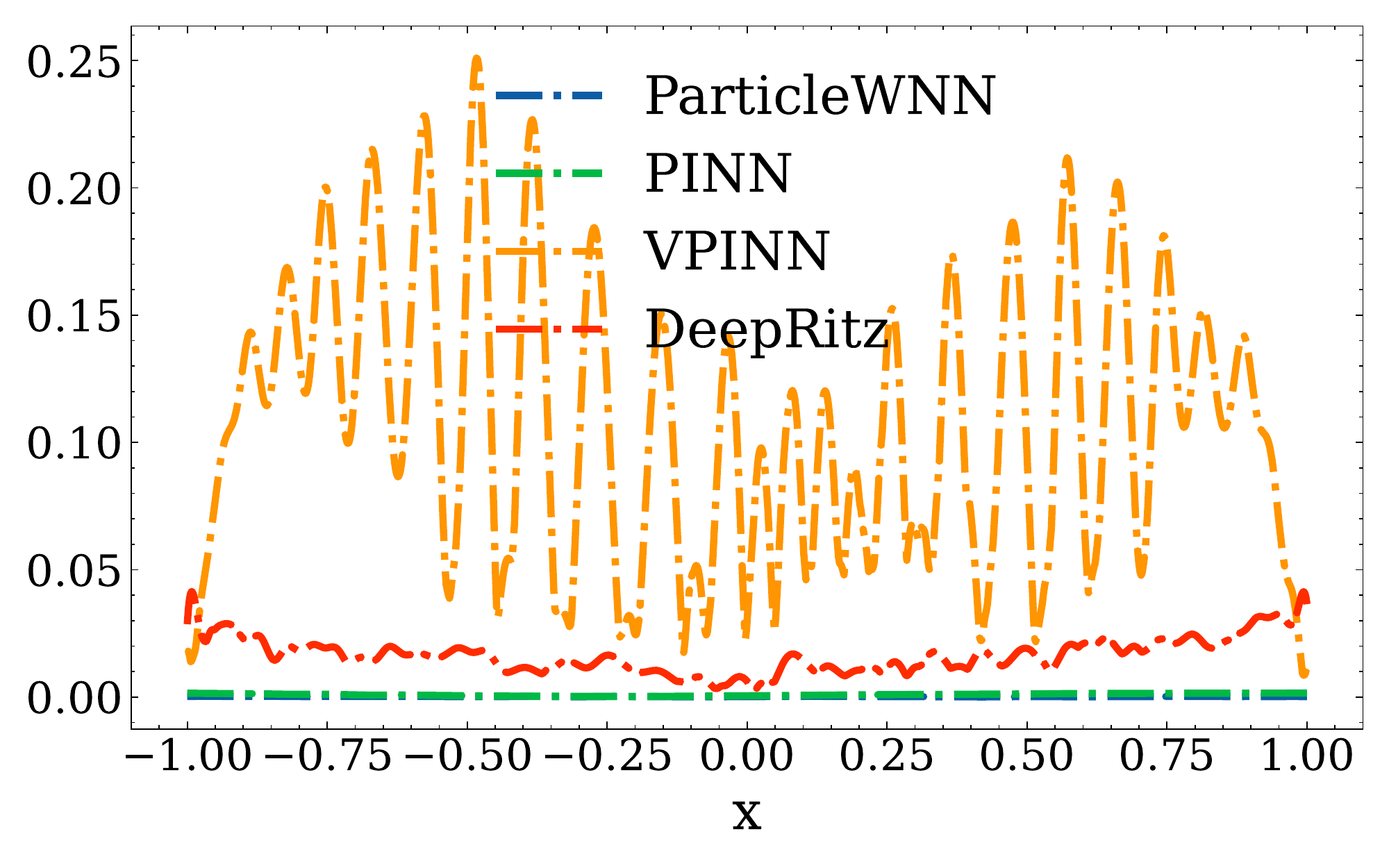}}
    \vspace{-0.25cm}
    \caption{The performance of different methods in solving the Poisson problem \eqref{eq:poisson_1d} 
    with frequency $\omega=15\pi$.  
    (a) Average Relative errors; (b) Average MAEs; 
    (c) Average Relative errors vs. Average computation times.
    (d) Average point-wise errors.} 
    \label{fig:poisson_1d_high}
\end{figure}

We explore two different frequency scenarios: a low-frequency case 
with $\omega=2\pi$ and a high-frequency case with $\omega=15\pi$.
The results for these cases are summarized in Table \ref{tab:poisson1d}, and 
visualized in Figures \ref{fig:poisson_1d_low} and \ref{fig:poisson_1d_high} for 
the low and high-frequency cases, respectively.
In the low-frequency scenario, as illustrated in Figure \ref{fig:poisson_1d_low}, 
both the DeepRitz and VPINN methods show faster convergence than ParticleWNN. 
However, their accuracy falls significantly short of ParticleWNN due to the limitation 
imposed by integration points. 
Although vanilla PINN achieves higher accuracy than ParticleWNN, it does so at 
the cost of slower convergence and increased computation time, as evident from Table \ref{tab:poisson1d}.
In contrast, as seen in Figure \ref{fig:poisson_1d_high} for the high-frequency case, 
ParticleWNN outperforms vanilla PINN, primarily because the error introduced by 
integral approximation in ParticleWNN becomes negligible compared to the 
approximation error of the neural network model in high-frequency scenarios. 
This underscores ParticleWNN's advantage in solving problems with complex solutions.
Furthermore, the limitation of integration points becomes more pronounced for 
the DeepRitz and VPINN methods in the high-frequency case. 
The VPINN method, in particular, struggles to produce acceptable solutions. 
Figure \ref{fig:poisson1d_time_high} shows that the relative error of the VPINN method 
initially decreases and then increases during the training process, indicative of 
overfitting due to insufficient integration points.
To validate the impact of increased integration points on the VPINN method, 
we doubled the integration points while keeping other settings constant. 
With this adjustment, the VPINN method achieved an average relative error of $0.1004 \pm 0.0178$, 
an average MAE of $0.1296 \pm 0.0266$, and an average computation time of $588.70 \pm 5.00$.
The results demonstrate an obvious improvement, but it still falls short of the performance 
achieved by the ParticleWNN method.
\begin{table}[!ht]\small
\centering
\caption{Experiment results for the Poisson problem \eqref{eq:poisson_1d}.}
\begin{tabular}{c|c|c|c|c|cc} \bottomrule
    \multicolumn{2}{c|}{} & ParticleWNN & vanilla PINN  & VPINN & DeepRitz \\ \hline
    \multirow{3}{*}{$\omega=2\pi$}& {Relative error} & $1.46e^{-4}\pm 3.86e^{-5}$ & $6.71e^{-5}\pm 3.54e^{-5}$ & $3.75e^{-2}\pm 1.91e^{-3}$ & $2.76e^{-2}\pm 1.38e^{-3}$  \\
    {} & {MAE} & $1.39e^{-4}\pm 5.05e^{-5}$ & $6.31e^{-5}\pm 3.43e^{-5}$ & $3.07e^{-2}\pm 4.17e^{-3}$ & $2.22e^{-2}\pm 1.86e^{-3}$ \\ 
    {} & {Time(s)} & $360.05\pm 1.92$ & $723.51\pm 2.89$ &$396.39\pm 9.13$ & $371.18\pm 3.99$  \\\hline
    \multirow{3}{*}{$\omega=15\pi$} & {Relative error} & $5.86e^{-4}\pm 8.28e^{-5}$ & $2.57e^{-3}\pm 1.59e^{-3}$ & $3.06e^{-1}\pm 1.77e^{-2}$ & $4.80e^{-2}\pm 1.44e^{-3}$ \\
    {} & {MAE} & $6.04e^{-4}\pm 7.43e^{-5}$ & $1.85e^{-3}\pm 1.01e^{-3}$ & $2.83e^{-1}\pm 3.30e^{-2}$ & $4.77e^{-2}\pm 6.48e^{-3}$\\ 
    {} & {Time(s)} & $359.63\pm 1.86$ & $722.16\pm 11.67$ &$357.92\pm 1.78$ & $350.57\pm 0.95$\\\toprule
\end{tabular}
\label{tab:poisson1d}
\end{table}
\subsection{The Allen-Cahn equation}
\label{sec:Allen-cahn}
\begin{figure}[!htbp]
    \centering  
    \hspace{-1cm}
    \subfigure[The exact $u$]{\label{fig:allen_cahn_abs}
        \includegraphics[width=0.35\textwidth]{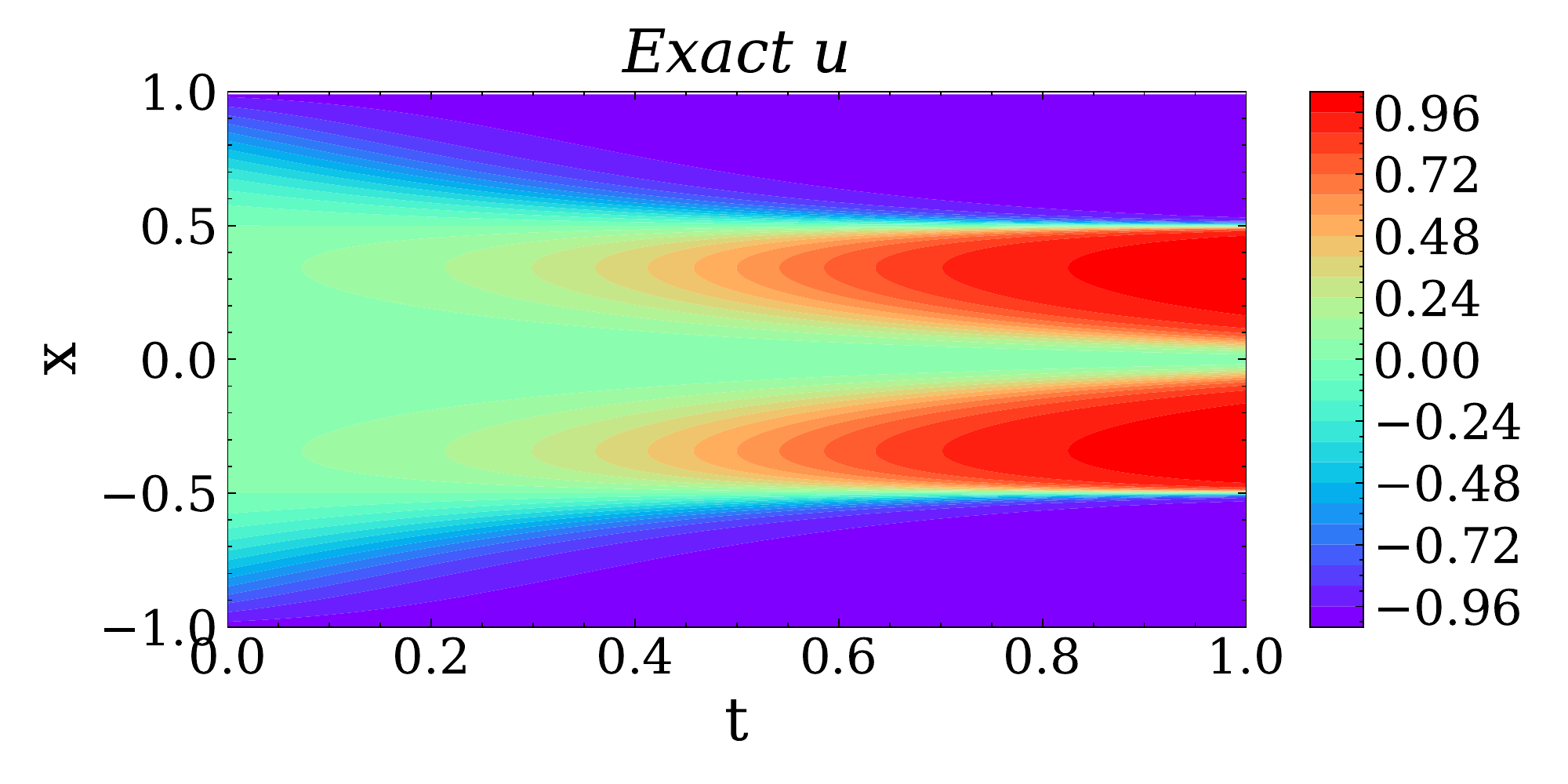}}
    \subfigure[Relative error]{\label{fig:allen_cahn_l2_errbar}
        \includegraphics[width=0.25\textwidth]{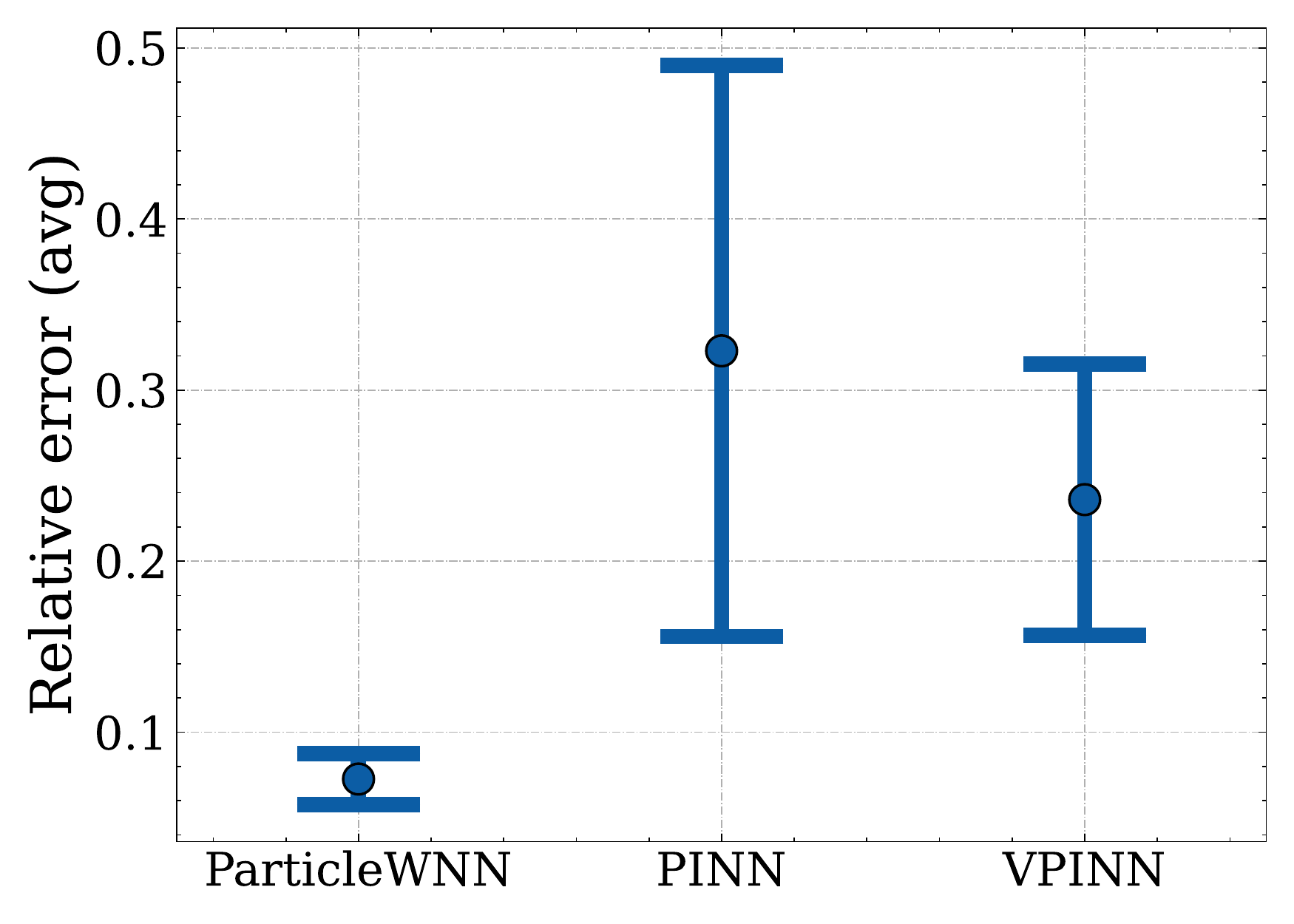}}
    \subfigure[Relative error vs. time]{\label{fig:allen_cahn_l2}
        \includegraphics[width=0.25\textwidth]{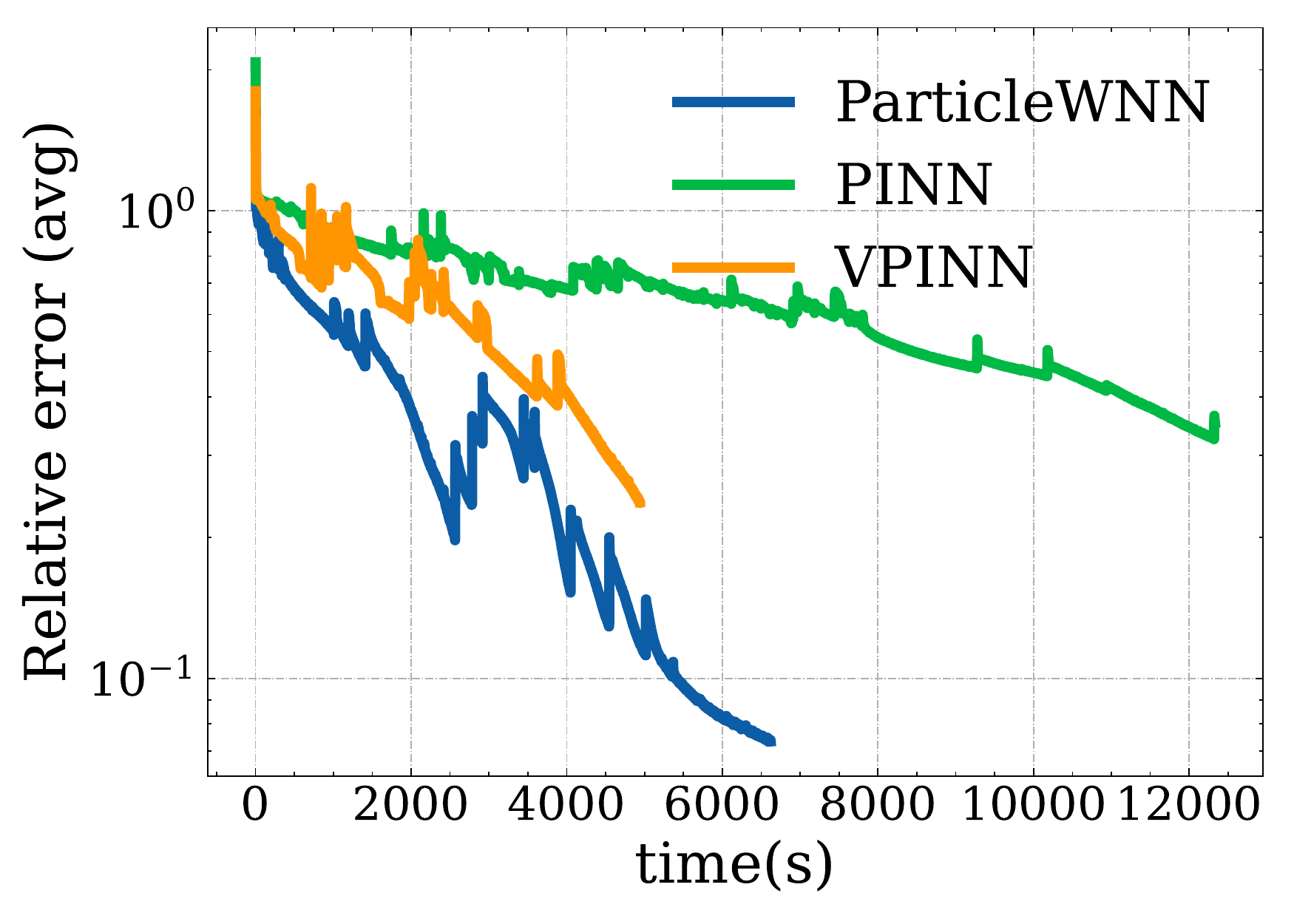}}
    \subfigure[$u_{NN}$ (ParticleWNN)]{\label{fig:allen_cahn_particlewnn}
    \includegraphics[width=0.32\textwidth]{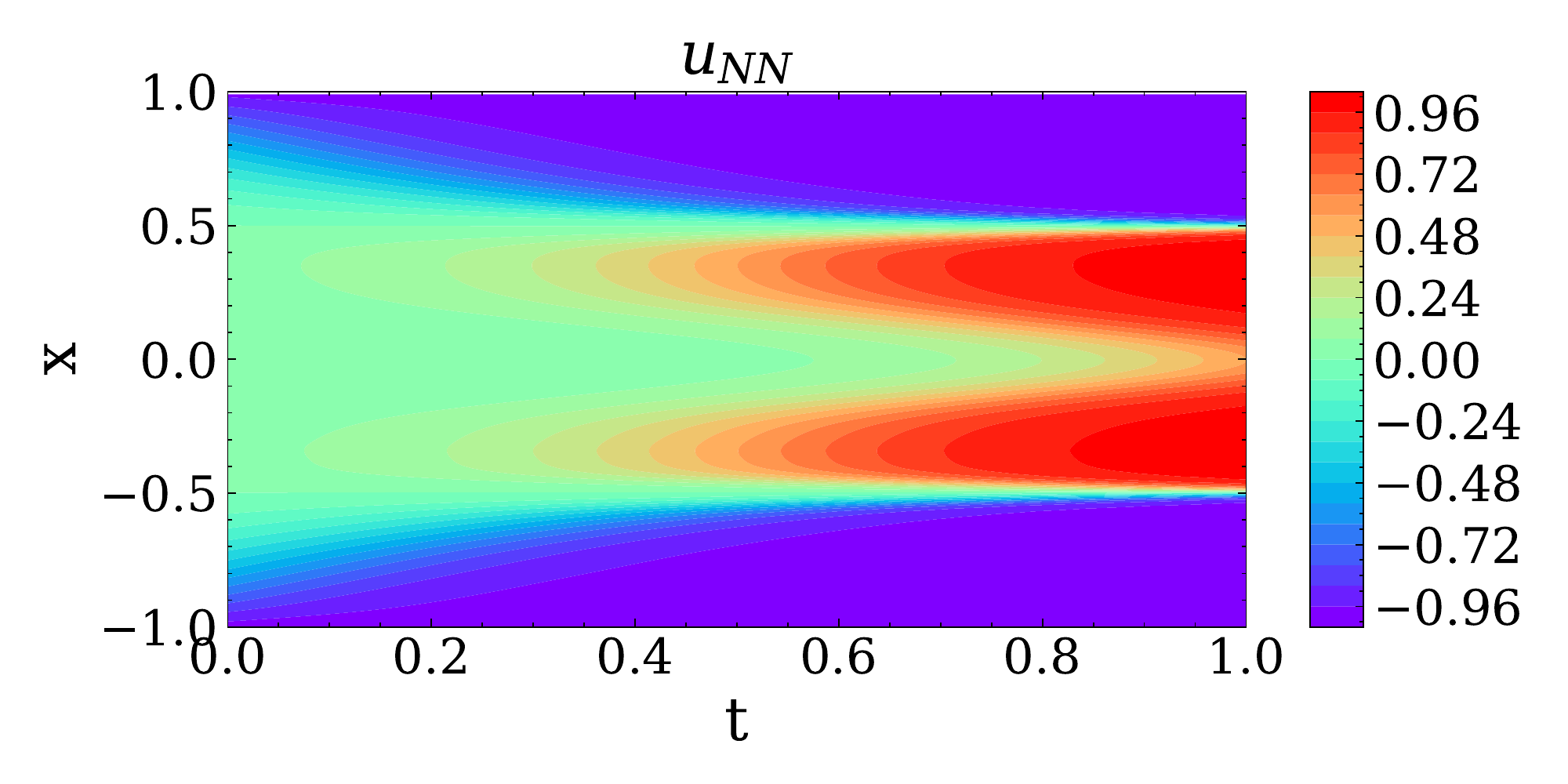}}
    \subfigure[$u_{NN}$ (vanilla PINN)]{\label{fig:allen_cahn_PINN}
        \includegraphics[width=0.32\textwidth]{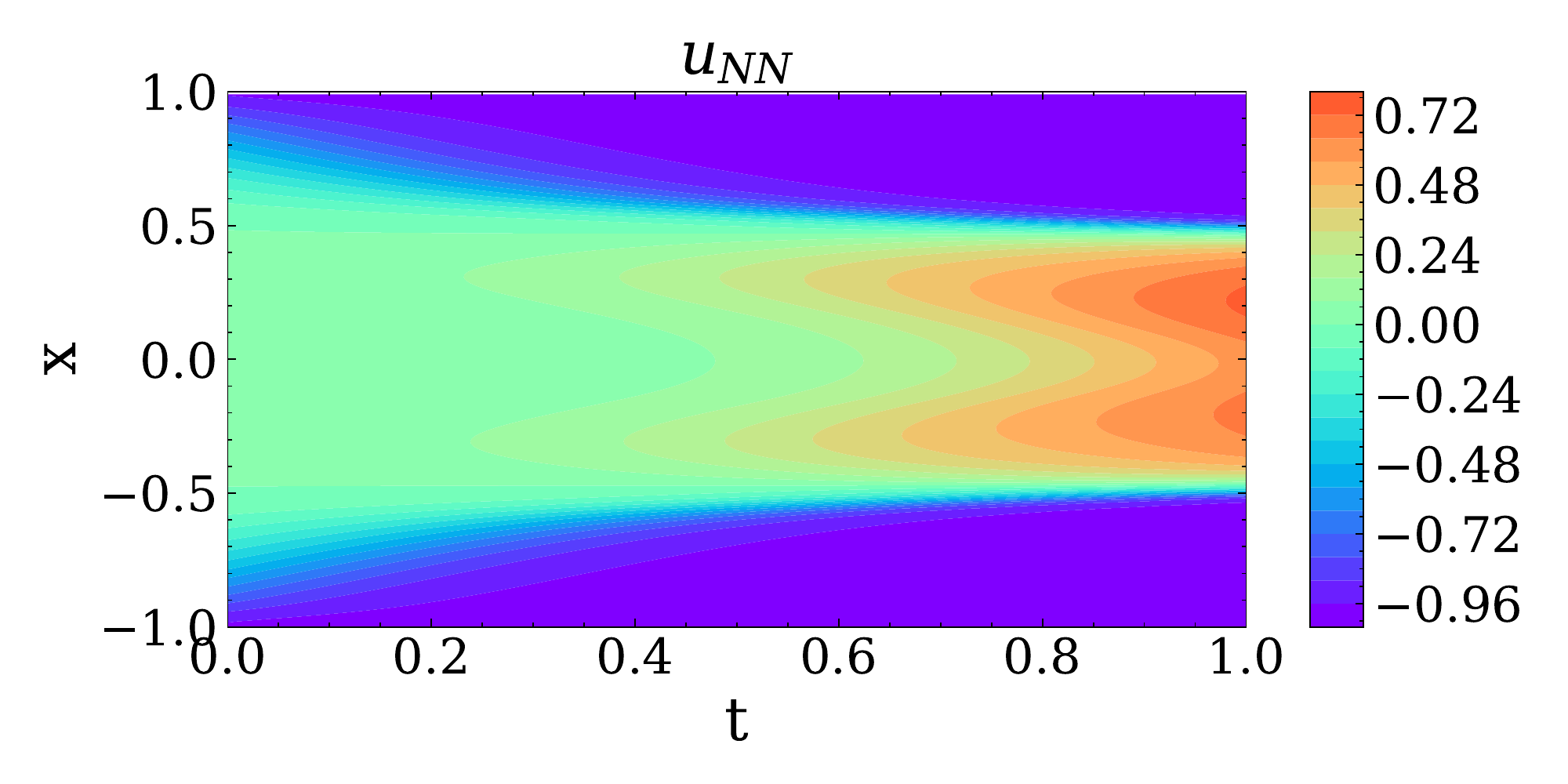}}
    \subfigure[$u_{NN}$ (VPINN)]{\label{fig:allen_cahn_VPINN}
        \includegraphics[width=0.32\textwidth]{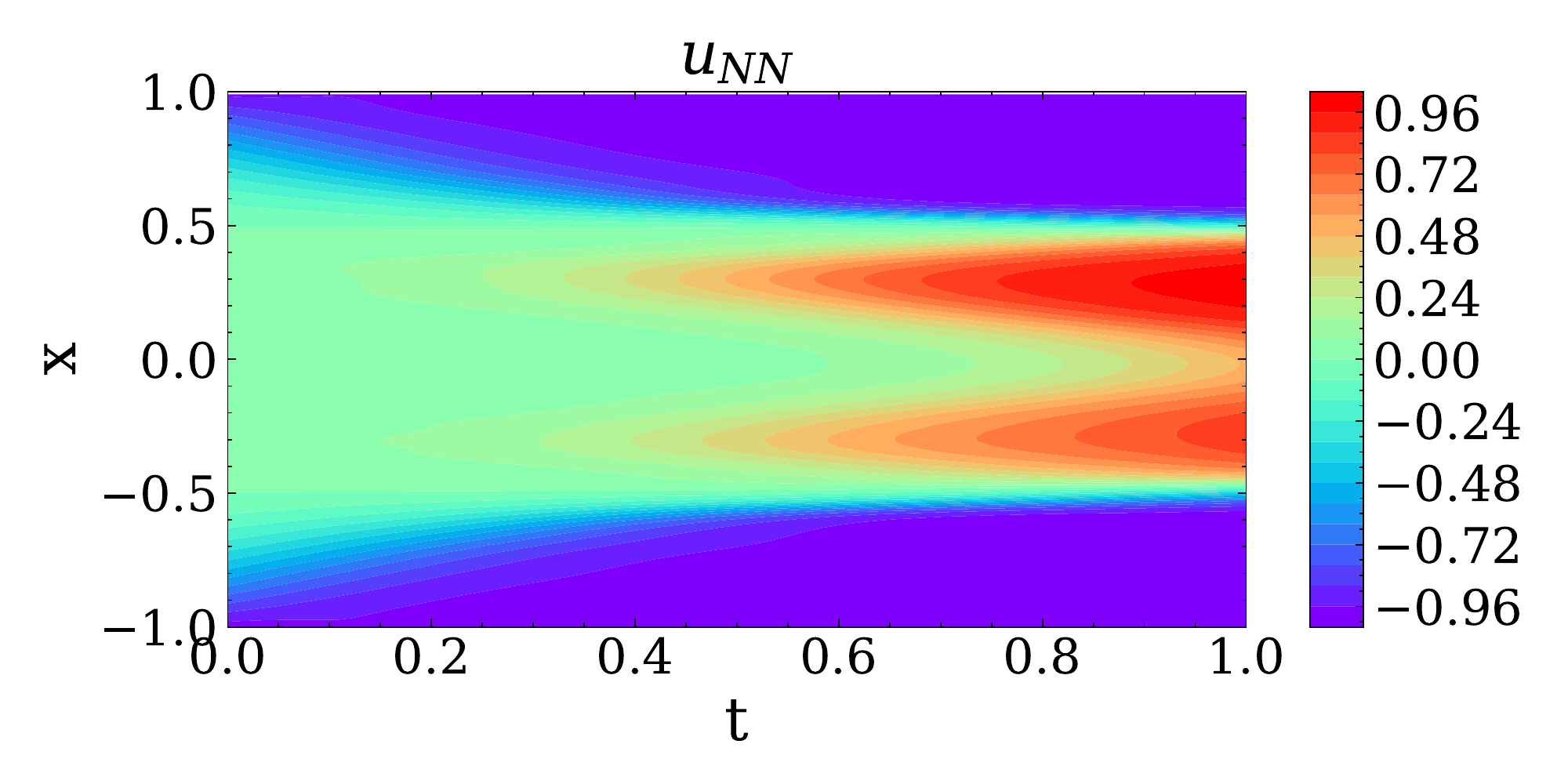}}
    \subfigure[Point-wise error (ParticleWNN)]{\label{fig:allen_cahn_point_wise_particlewnn}
        \includegraphics[width=0.32\textwidth]{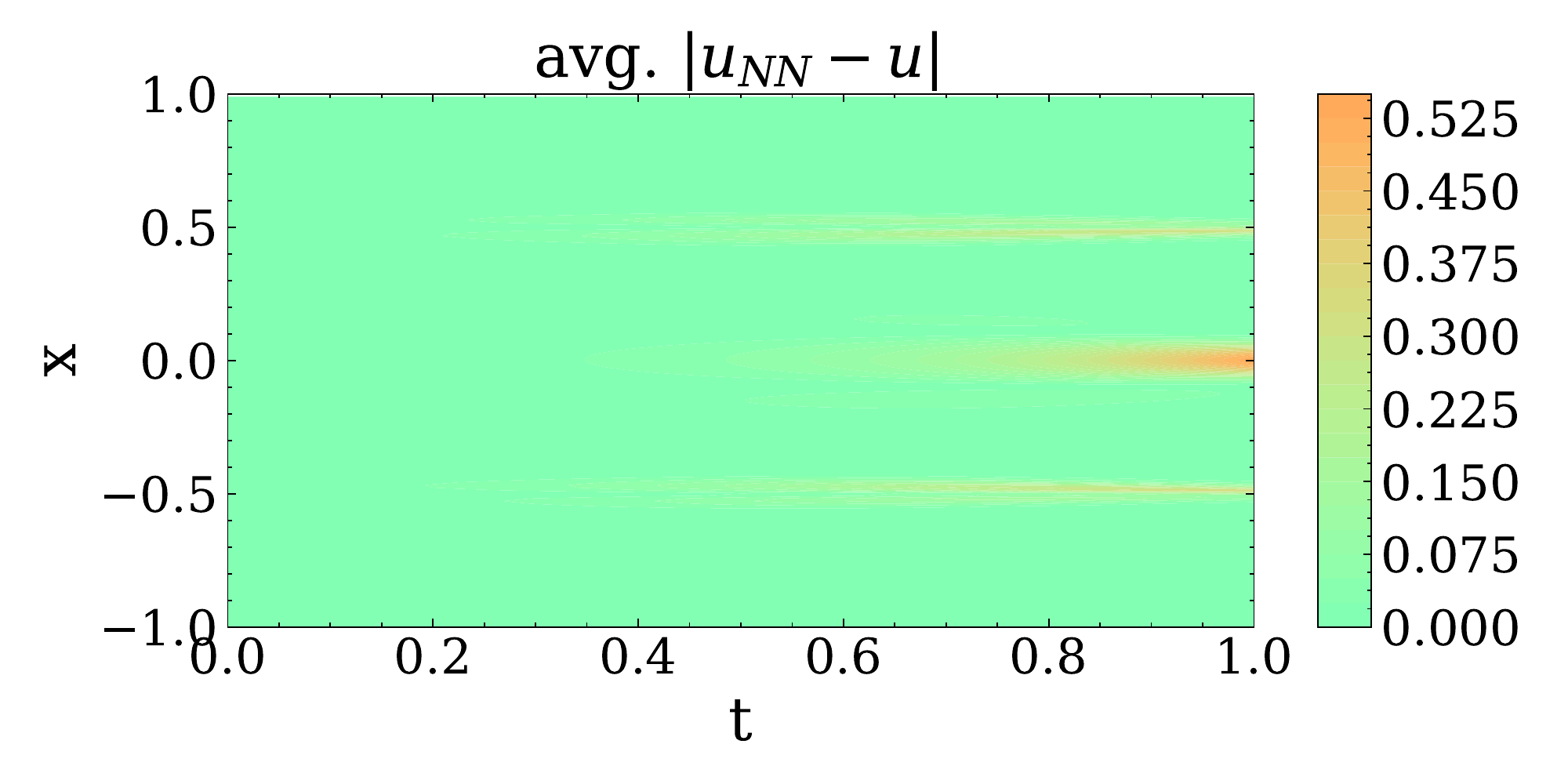}}
    \subfigure[Point-wise error (vanilla PINN)]{\label{fig:allen_cahn_point_wise_PINN}
        \includegraphics[width=0.32\textwidth]{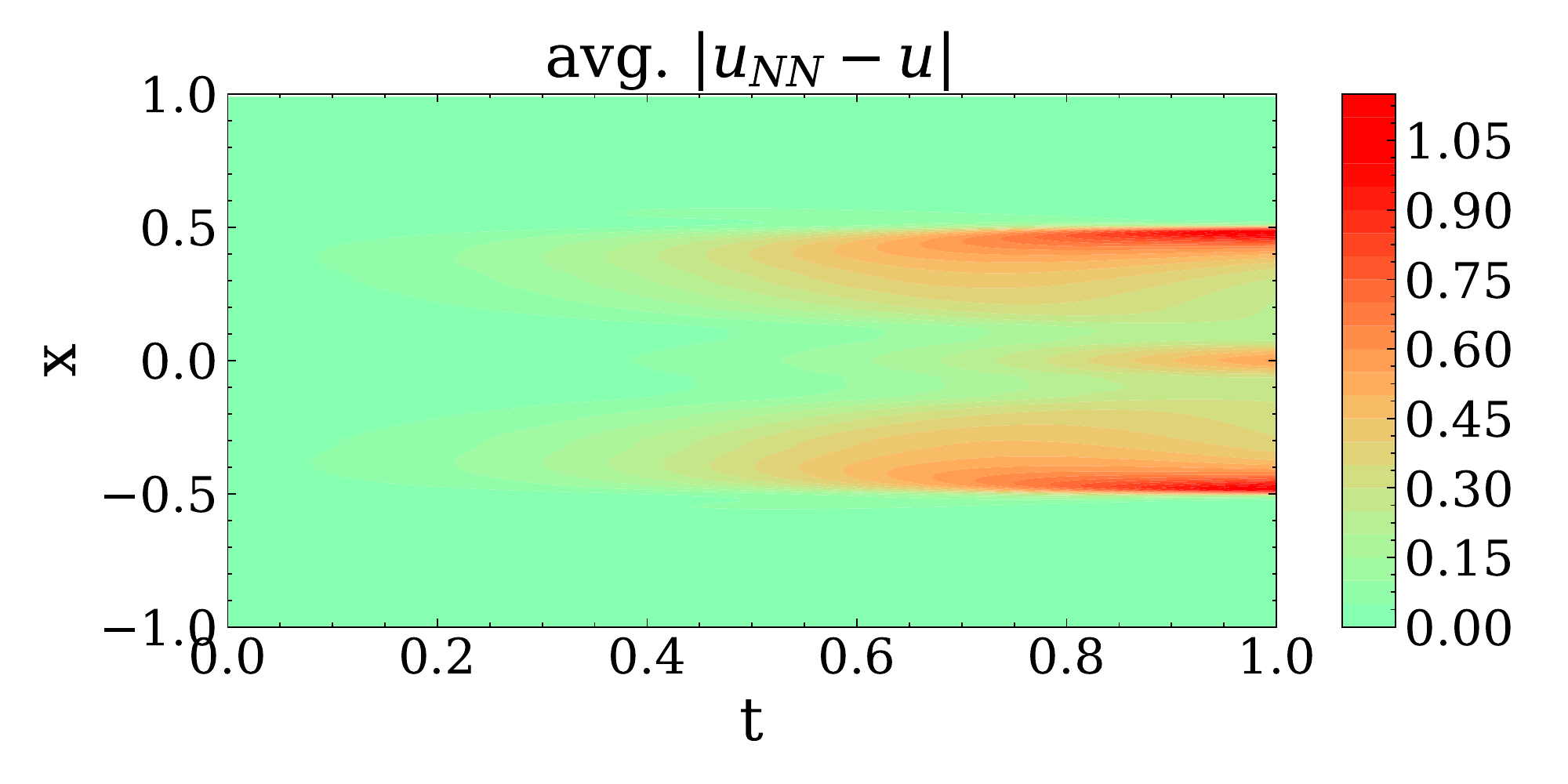}}
    \subfigure[Point-wise error (VPINN)]{\label{fig:allen_cahn_point_wise_VPINN}
        \includegraphics[width=0.32\textwidth]{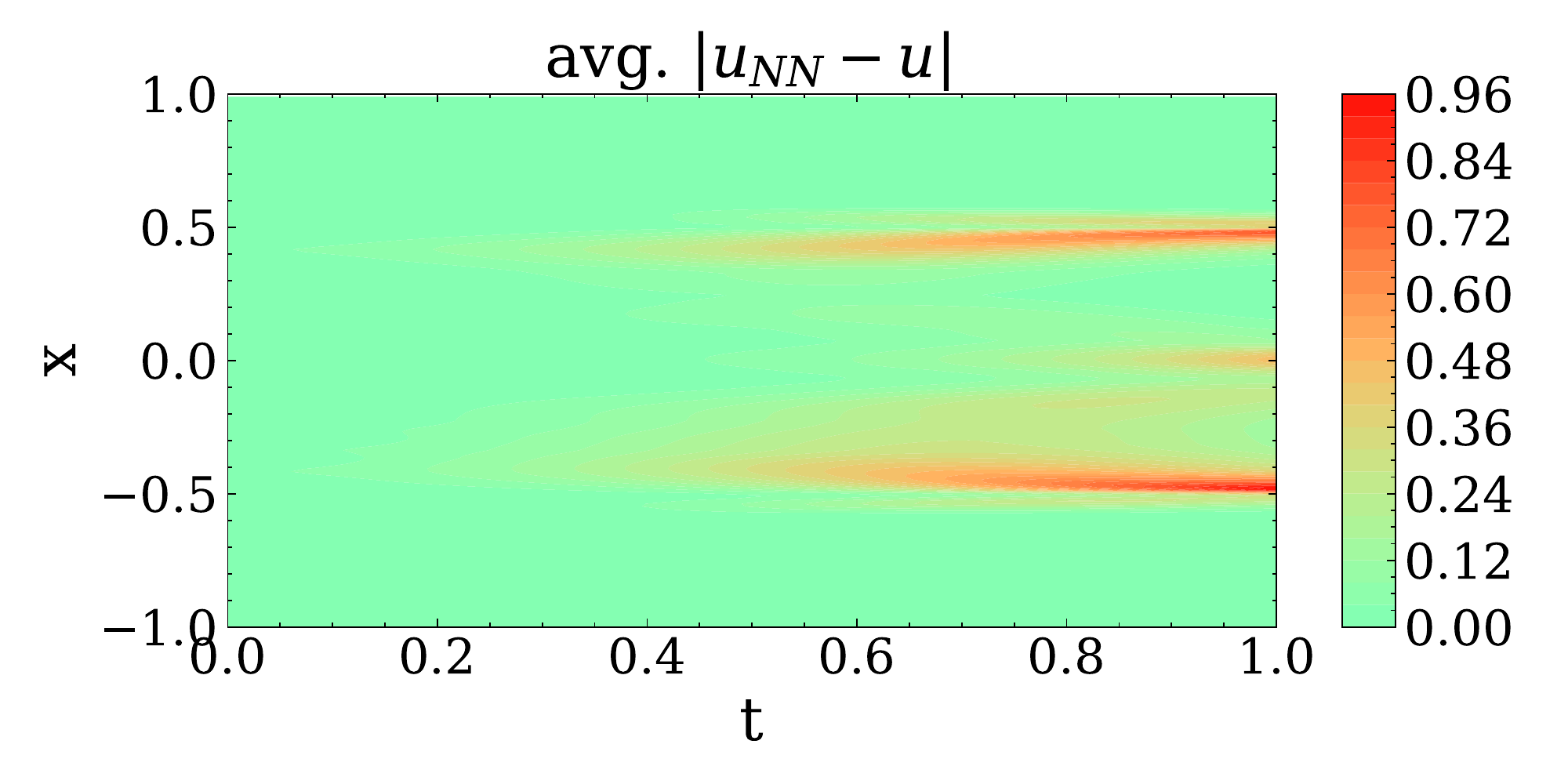}}
    \vspace{-0.25cm}
    \caption{The performance of different methods in solving the Allen-Cahn problem \eqref{eq:allen-cahn}. 
    (a) The exact $u$; (b) Average Relative errors; 
    (c) Average Relative errors vs. Average computation times. 
    (d), (e), (f) The average $u_{NN}$ obtained by different methods;
    (g), (h), (i) The average point-wise errors obtained by different methods.} 
    \label{fig:allen-cahn}
\end{figure}

We then consider the following nonlinear Allen-Cahn equation
\begin{equation}\label{eq:allen-cahn}
    \begin{cases}
        u_t - \lambda u_{xx} + 5u^3 - 5u =0,\quad t\in(0,1],\ x\in[-1,1], \\
        u(x,0) = x^2\cos(\pi x), \\
        u(t,-1) = u(t,1),\ u_x(t,-1) = u_x(t,1),
    \end{cases}
\end{equation}
where $\lambda=0.0001$.
The equation is difficult to solve due to its sharp solution (see Figure \ref{fig:allen_cahn_abs}).
To solve this problem with the ParticleWNN, we select the test function 
$\varphi(\bm{x})$ in Algorithm \ref{alg:ParticleWNN} such that it 
depends only on $\bm{x}$ and multiply 
it with both sides of Equation \eqref{eq:allen-cahn}.
Then, we get the weak formulation through variation:
\begin{equation}\label{eq:weak_time}
\int_{\Omega}(u_t+5u^3-5u) \varphi\ d\bm{x}  + \lambda\int_{\Omega} \nabla_{\bm{x}} u \cdot\nabla_{\bm{x}} \varphi\ d\bm{x} = 0,\quad t\in(0,1]. 
\end{equation}
We handle the initial value conditions similarly to the boundary conditions.
We sample $N_{init}=200$ points in $\Omega$ to evaluate the mismatch of initial conditions 
and set the corresponding loss weight as $\lambda_\mathcal{I}=50$.
To evaluate the weak-form residuals, we sample $N_t=100$ time points in $(0,1]$.
At each time point, we randomly sample $N_p=50$ particles in $\Omega$ and set $K_{int}=25$ to 
calculate the integrals in \eqref{eq:weak_time}. 
For other parameters, we set $N_{bd}=200$, $topK=4000$, $\lambda_{\mathcal{R}}=100$, $maxIter=50000$, 
and increase the number of neurons per layer in the ResNet to $100$ and use the composite function 
of Tanh and Sine as the activation.
For comparison, we generate $125,000$ ($topK=100,000$) collocation points for the vanilla PINN 
to evaluate the strong residuals and keep other settings consistent with ParticleWNN.
For the VPINN method, we employ $N_t=100$ time points, along with $N_{test}=50$ test functions 
and $N_{int}=25$ integration points. 

The experimental results are summarized in Table \ref{tab:Allen-cahn} and 
depicted in Figure \ref{fig:allen-cahn}.
Figure \ref{fig:allen_cahn_l2} indicates that ParticleWNN achieves higher accuracy and 
faster convergence than both the vanilla PINN and the VPINN method.
The Table \ref{tab:Allen-cahn} also shows that the ParticleWNN takes almost half 
the computation time of the vanilla PINN.
\begin{table}[!ht]\small
\centering
\caption{Experiment results for the Allen-Cahn problem \eqref{eq:allen-cahn}.}
\begin{tabular}{c|c|c|cc} \bottomrule
                    {} & ParticleWNN & vanilla PINN  & VPINN \\ \hline
    {Relative error} & $0.073\pm 0.015$ & $0.323\pm 0.167$ & $0.236\pm 0.079$ \\
    {MAE} & $0.529\pm 0.058$ & $1.243\pm 0.402$ & $1.158\pm 0.368$ \\ 
    {Time(s)} & $6622.33\pm 29.47$ & $12332.18\pm 70.20$ &$4942.12\pm 13.36$\\\toprule
\end{tabular}
\label{tab:Allen-cahn}
\end{table}
\subsection{The 2D imcompressible Navier-Stokes equation}
We proceed to solve a steady 2D incompressible NS equation, 
using the Kovasznay flow as an example.
In non-dimensional form, the mathematical model is given by:
\begin{equation}\label{eq:NS_steady}
    \begin{cases}
        -\nu \Delta\bm{u} + (\bm{u}\cdot\nabla)\bm{u} + \nabla p = 0,\quad \text{in}\ \Omega=[-1,1]\times[-1,1], \\
        \nabla\cdot\bm{u} = 0, \quad \text{in}\ \Omega
    \end{cases}
\end{equation}
where $\nu$ is the viscosity coefficient, and $\bm{u}=(u,v)$ and $p$ 
are the velocity field and the pressure field, respectively. 
The analytical solution for this equation is given by \cite{drazin2006navier}:
\begin{equation}
    u(x,y) = 1-e^{\lambda x}\cos(\omega y),\quad 
    v(x,y) = \frac{\lambda}{\omega}e^{\lambda x}\sin(\omega y),\quad 
    p(x,y) = \frac{1}{2}(1-e^{2\lambda x}),
\end{equation}
where 
\begin{equation*}
    \lambda = \frac{1}{2\nu} - \sqrt{\frac{1}{4\nu^2} + \omega^2}.
\end{equation*}
In our setup, we choose $\nu=0.025$ and $\omega=3\pi$. 
To solve the problem, we approximate the pressure $p$ with one network and 
use another network with two outputs to approximate both $u$ and $v$. 
We adopt the Tanh activation function for the DNN models.
In the ParticleWNN method, we set $N_p=250$ (without $topK$), $K_{int}=124$, 
$\tilde{N}_{bd}=250$.
For a fair comparison, the vanilla PINN method utilizes $31,000$ 
collocation points, while the VPINN method use $N{test}=50$ test functions 
and $N_{int}=625$ integration points. 
Other settings are consistent across all methods.

\begin{figure}[!htbp]
    \centering  
    \hspace{-0.75cm}
    \subfigure[Relative error of $u$ vs. time]{\label{fig:NS_l2_u}
        \includegraphics[width=0.3\textwidth]{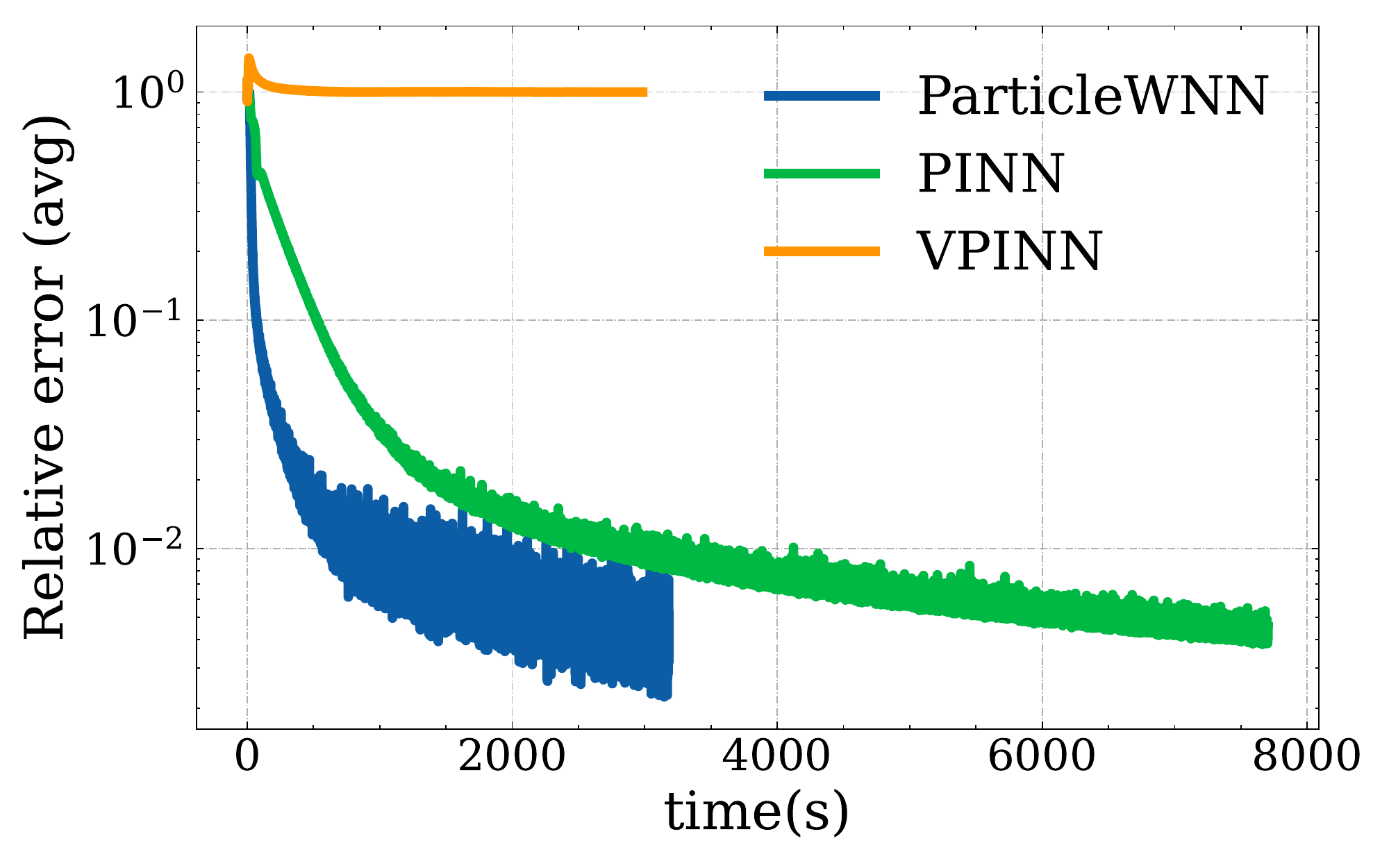}}
    \subfigure[Relative error of $v$ vs. time]{\label{fig:NS_l2_v}
        \includegraphics[width=0.3\textwidth]{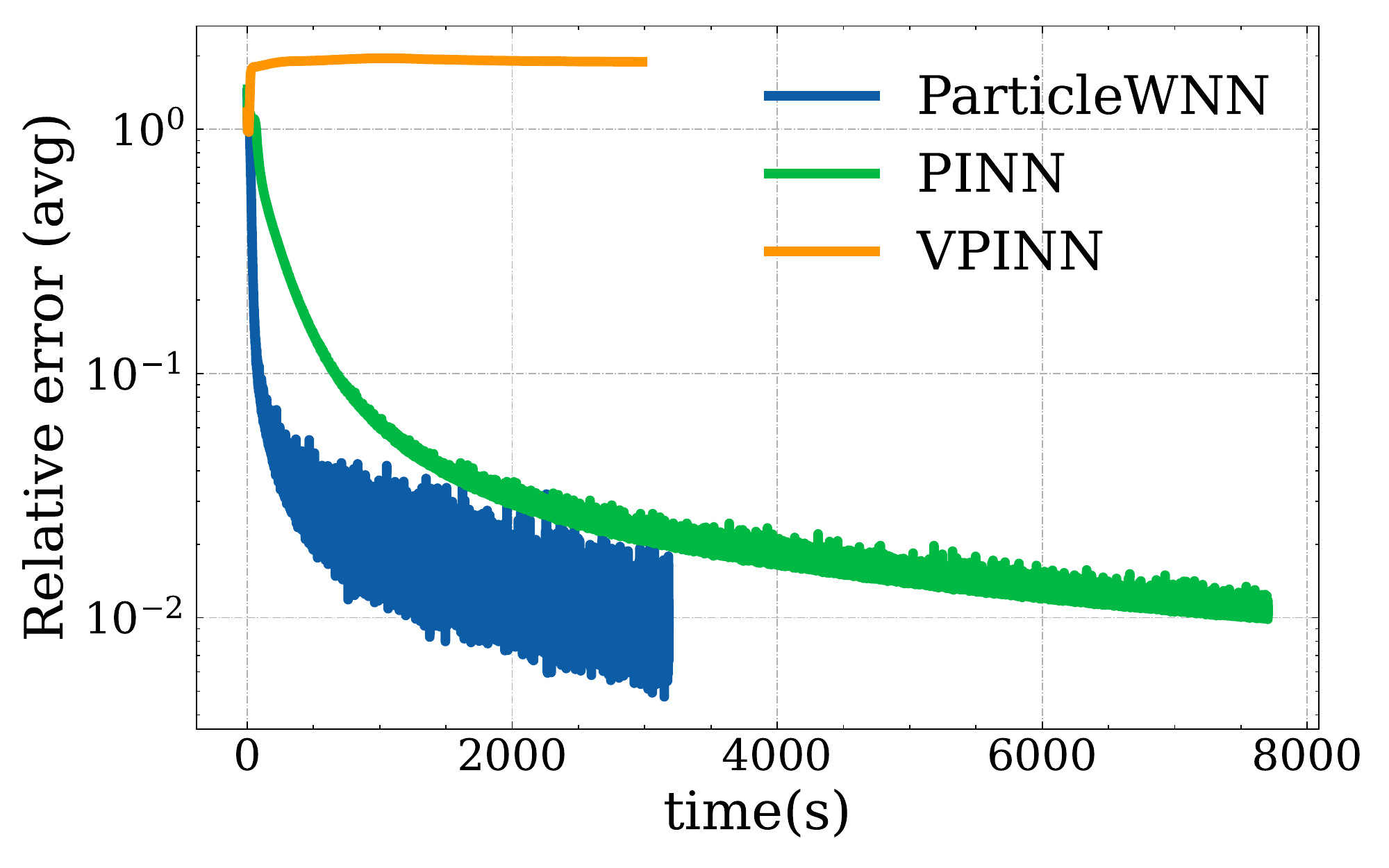}}
    \subfigure[Relative error of $p$ vs. time]{\label{fig:NS_l2_p}
        \includegraphics[width=0.3\textwidth]{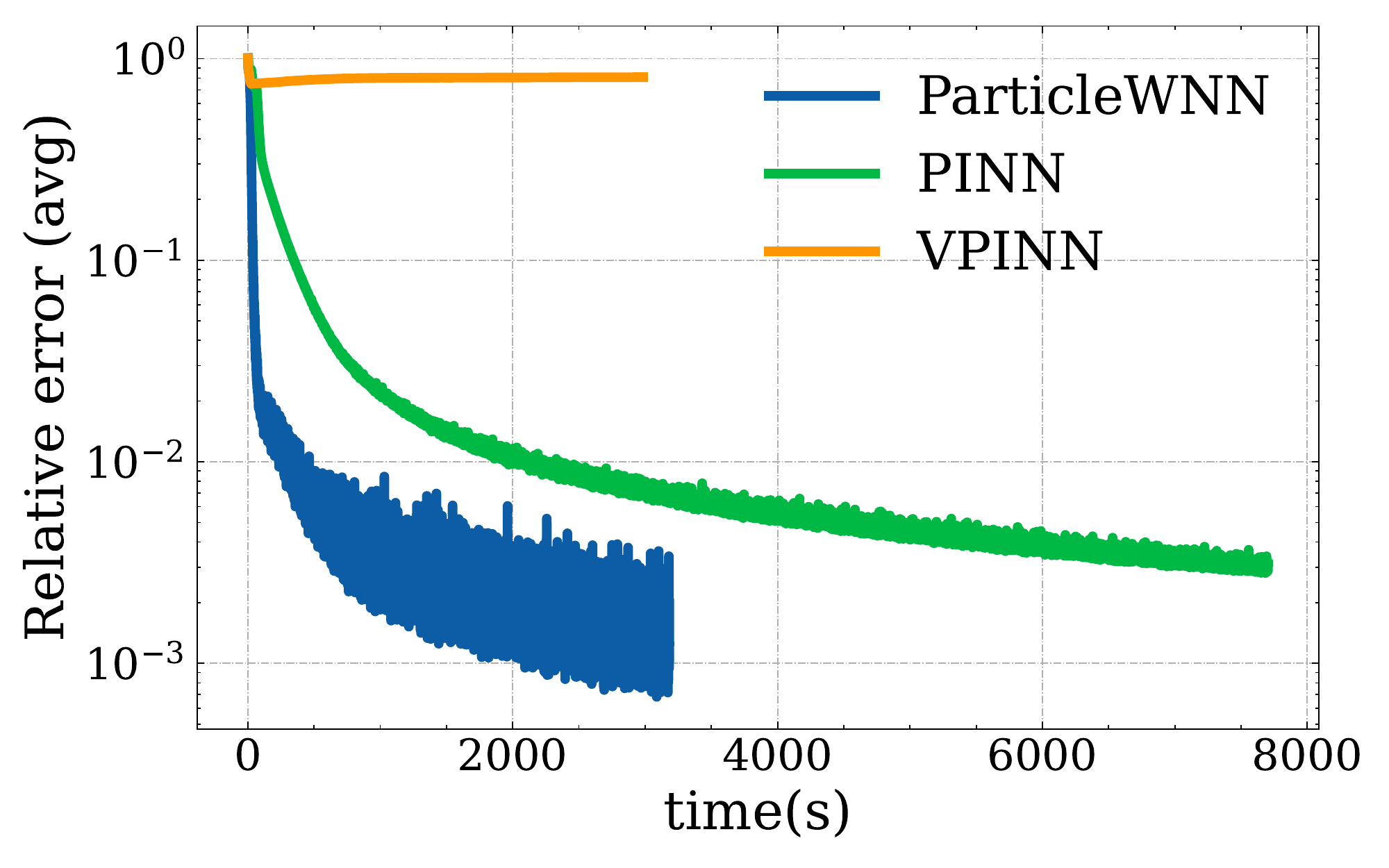}}
    \subfigure[$u_{NN}$ (ParticleWNN)]{\label{fig:NS_u_particlewnn}
        \includegraphics[width=0.32\textwidth]{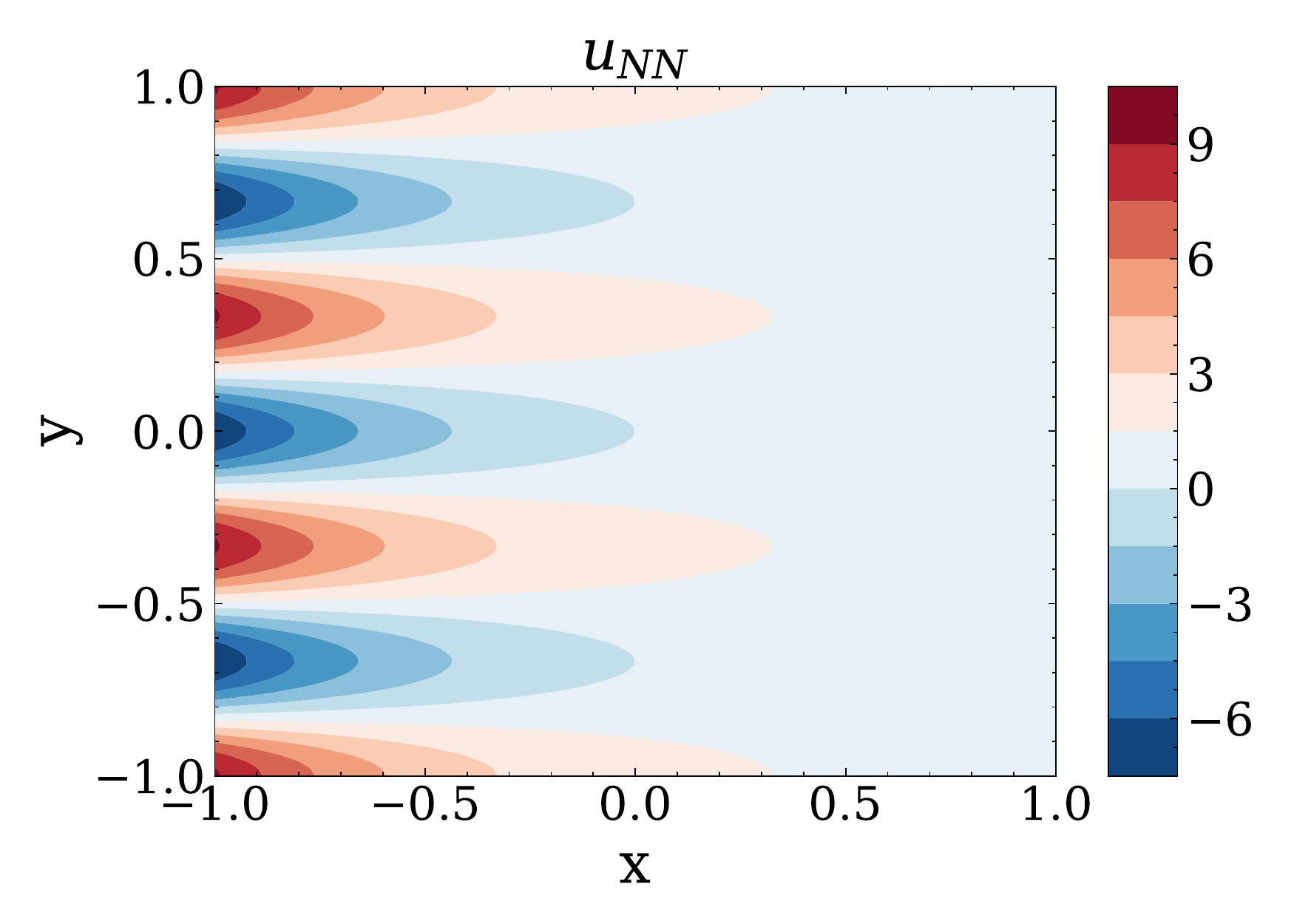}}
    \subfigure[$v_{NN}$ (ParticleWNN)]{\label{fig:NS_v_particlewnn}
        \includegraphics[width=0.32\textwidth]{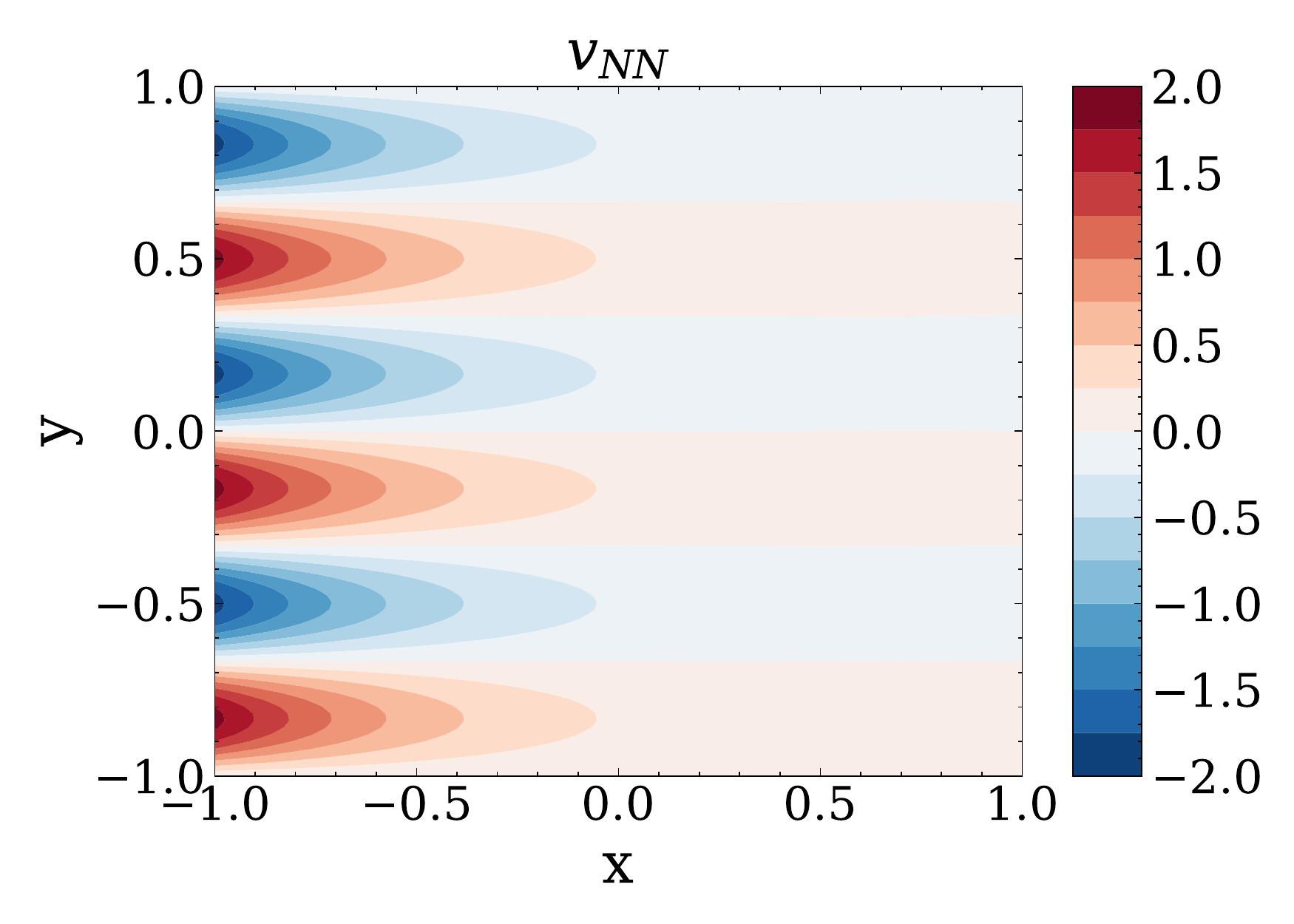}}
    \subfigure[$p_{NN}$ (ParticleWNN)]{\label{fig:NS_p_particlewnn}
        \includegraphics[width=0.32\textwidth]{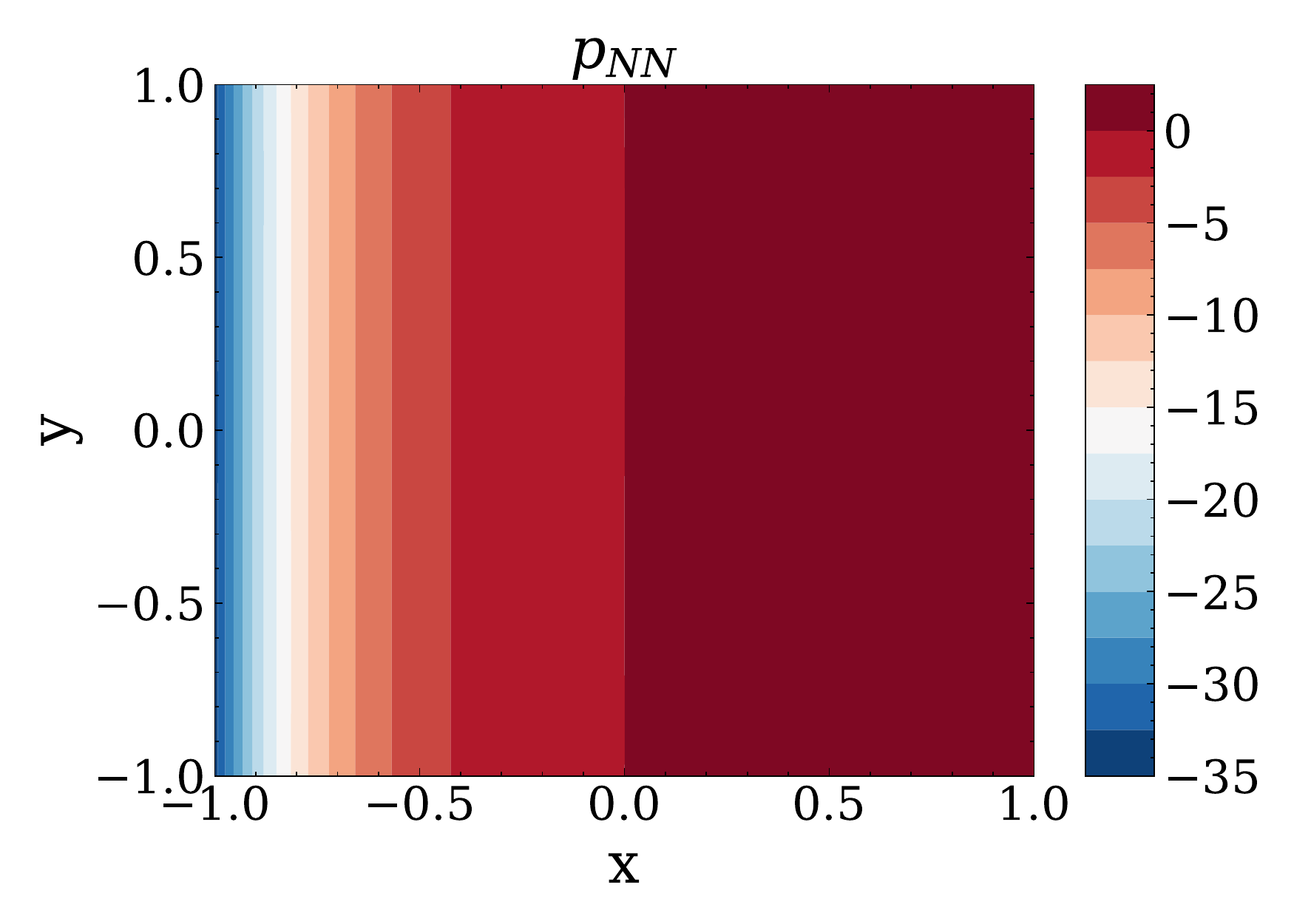}}
    \subfigure[$u_{NN}$ (vanilla PINN)]{\label{fig:NS_u_PINN}
        \includegraphics[width=0.32\textwidth]{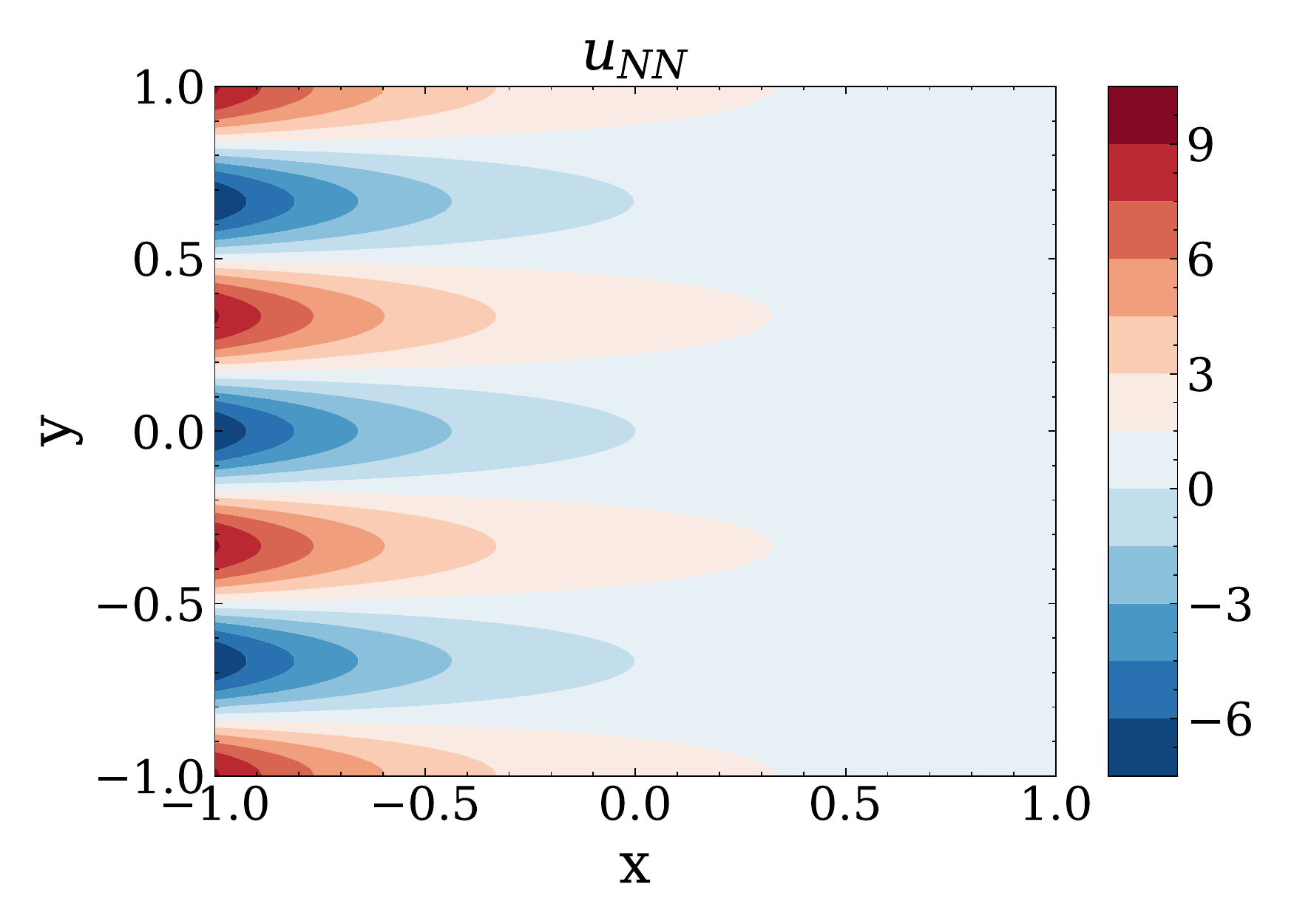}}
    \subfigure[$v_{NN}$ (vanilla PINN)]{\label{fig:NS_v_PINN}
        \includegraphics[width=0.32\textwidth]{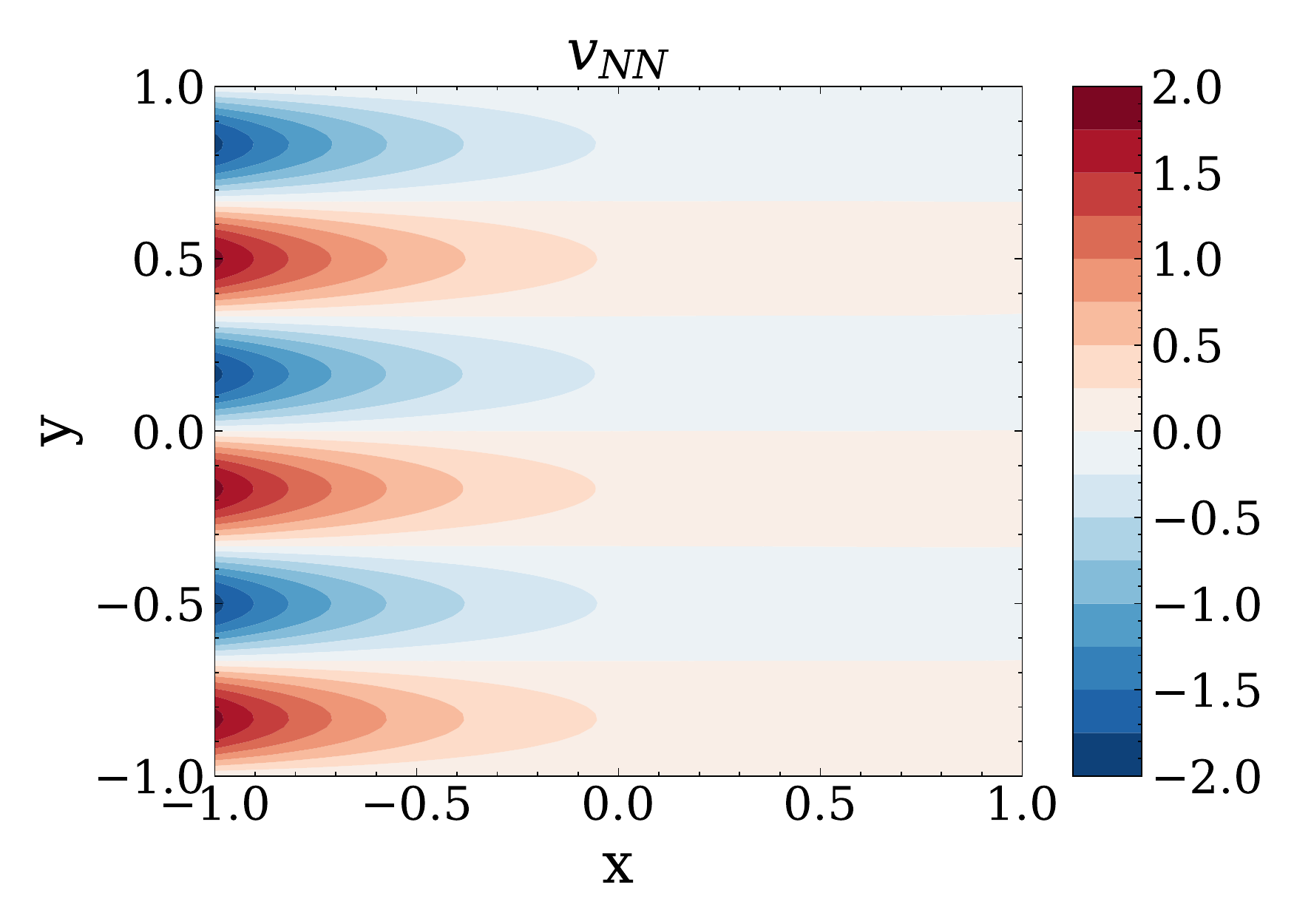}}
    \subfigure[$p_{NN}$ (vanilla PINN)]{\label{fig:NS_p_PINN}
        \includegraphics[width=0.32\textwidth]{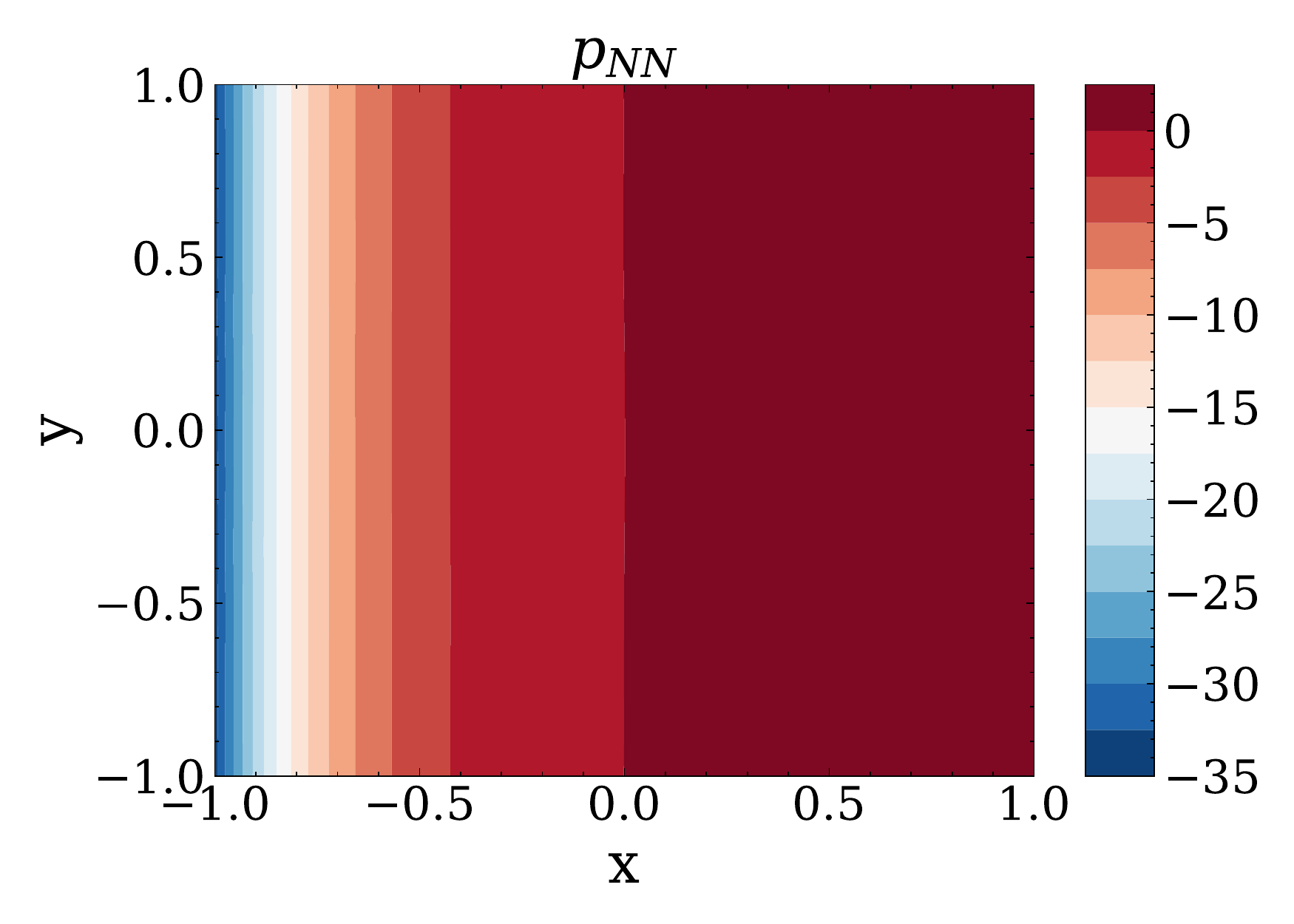}}
    \subfigure[$u_{NN}$ (VPINN)]{\label{fig:NS_u_VPINN}
        \includegraphics[width=0.32\textwidth]{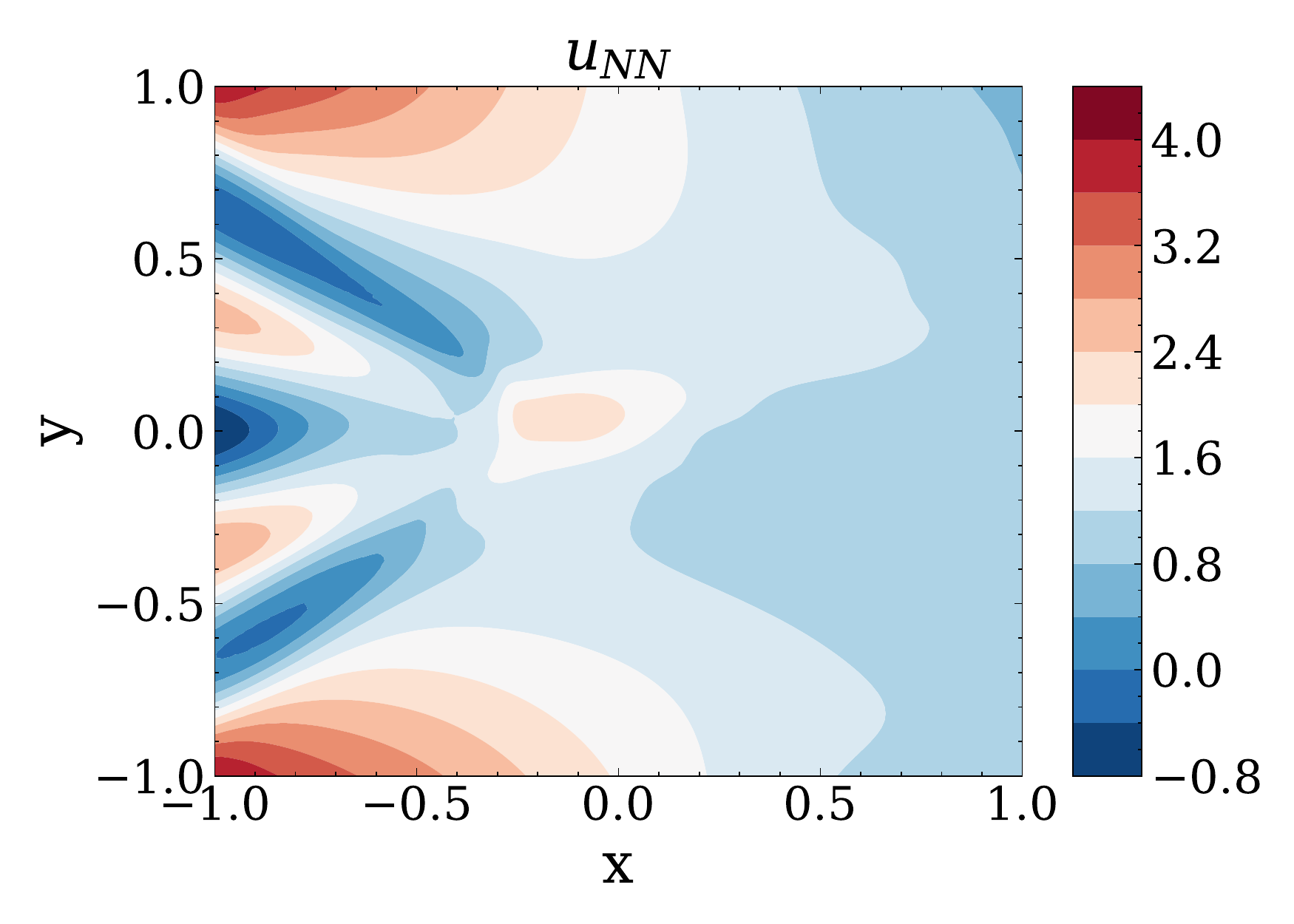}}
    \subfigure[$v_{NN}$ (VPINN)]{\label{fig:NS_v_VPINN}
        \includegraphics[width=0.32\textwidth]{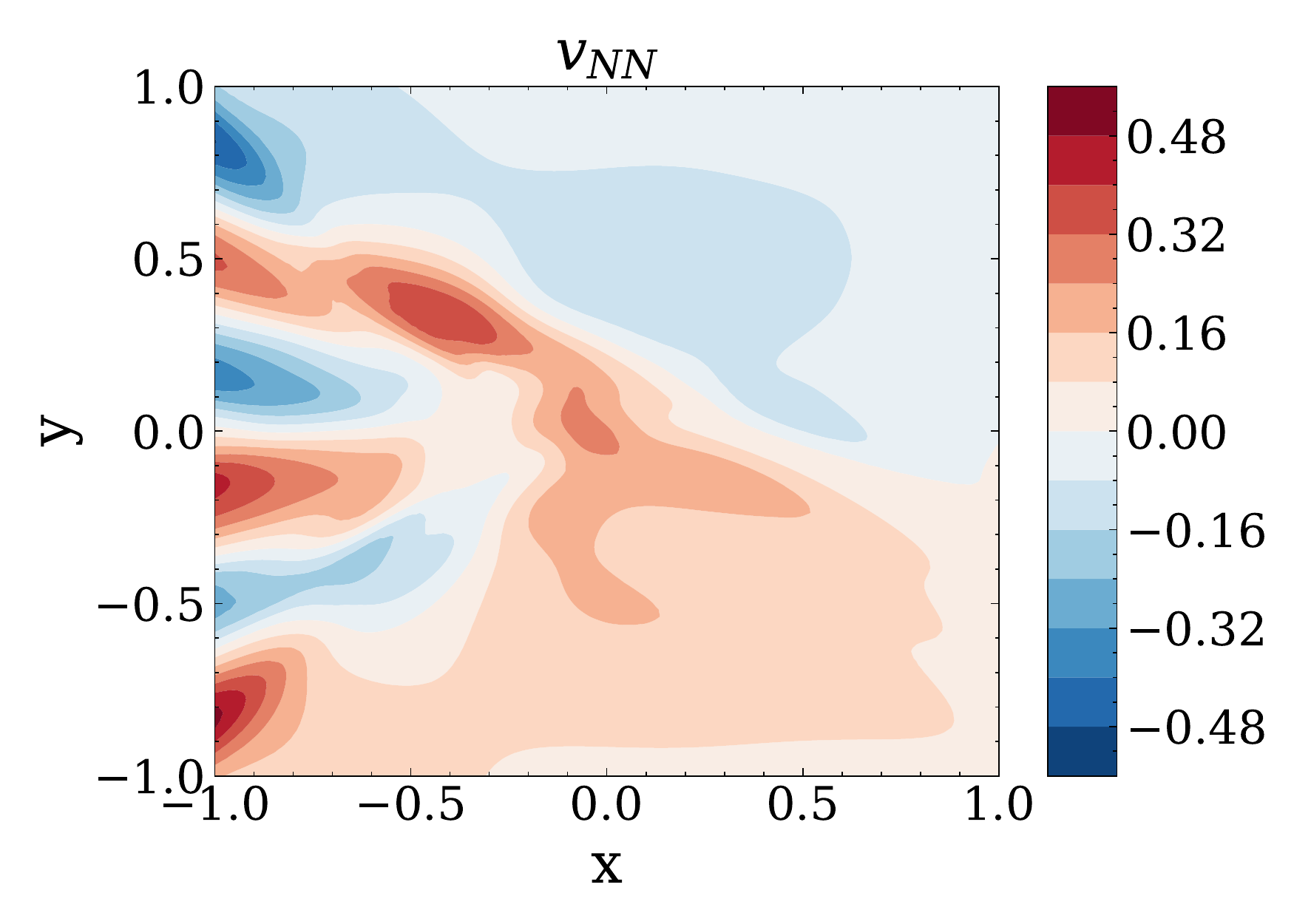}}
    \subfigure[$p_{NN}$ (VPINN)]{\label{fig:NS_p_VPINN}
        \includegraphics[width=0.32\textwidth]{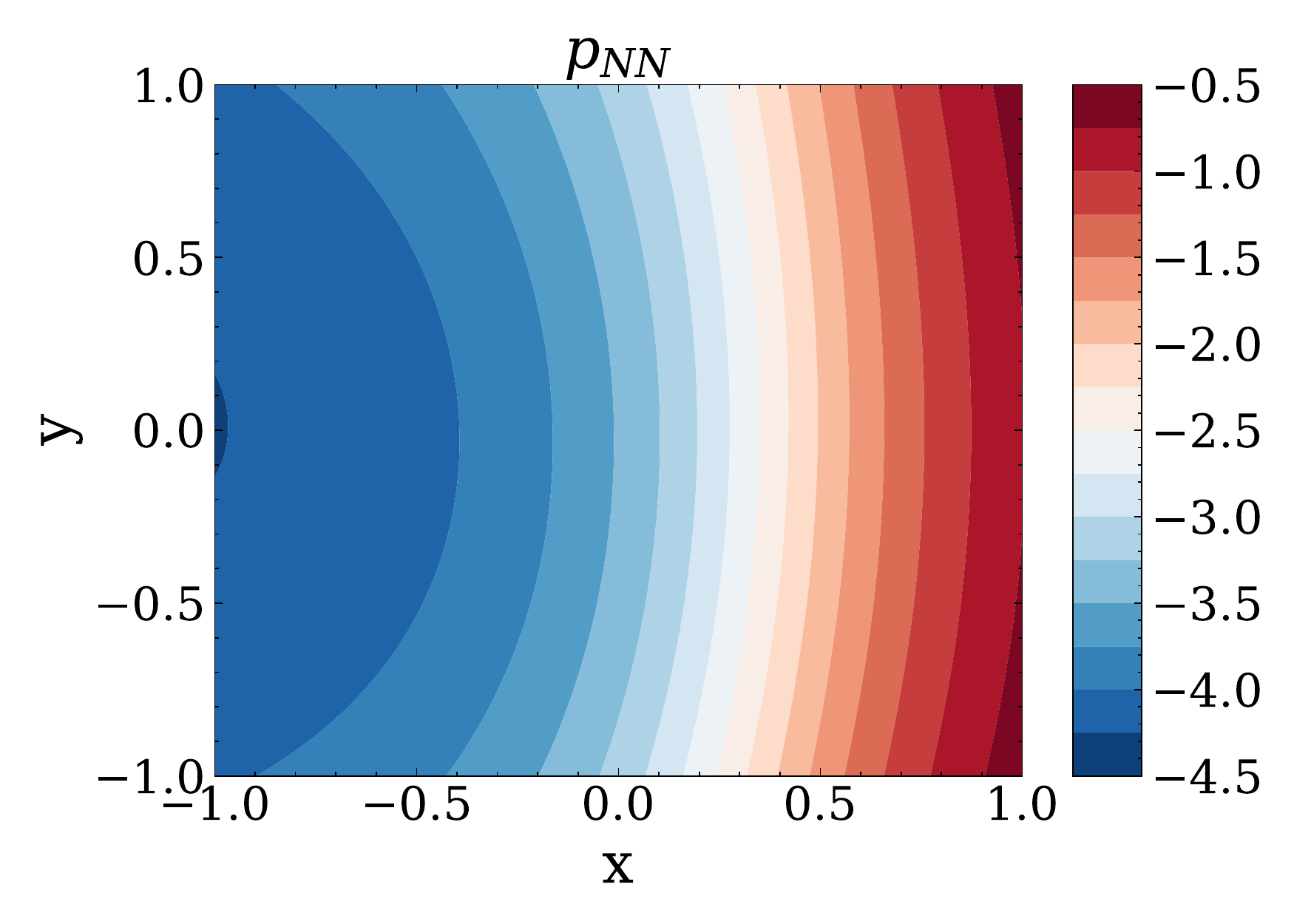}}
    \vspace{-0.25cm}
    \caption{The performance of different methods in solving the steady NS problem \eqref{eq:NS_steady}. 
    (a) The average Relative errors of $u$ vs. the average times;
    (b) The average Relative errors of $v$ vs. the average times;
    (c) The average Relative errors of $p$ vs. the average times;
    (d), (e), (f) The average predicted $u_{NN}$, $v_{NN}$, and $p_{NN}$ obtained by the ParticleWNN, respectively;
    (g), (h), (i) The average predicted $u_{NN}$, $v_{NN}$, and $p_{NN}$ obtained by the vanilla PINN, respectively;
    (j), (k), (l) The average predicted $u_{NN}$, $v_{NN}$, and $p_{NN}$ obtained by the VPINN, respectively.} 
    \label{fig:NS_steady}
\end{figure}

After $maxIter=100,000$ iterations, the results are summarized in Table \ref{tab:NS_3pi}. 
Figure \ref{fig:NS_steady} displays the numerical predictions of $u$, $v$, and $p$ obtained 
by each method, along with corresponding relative error plots over time.
From Figure \ref{fig:NS_steady}, it is evident that the VPINN method struggles to achieve 
an accurate solution due to the limitations imposed by insufficient integration points.
In contrast, both the vanilla PINN and ParticleWNN methods deliver accurate solutions.
The ParticleWNN method also demonstrates faster convergence than the vanilla PINN, 
with the added benefit of consuming less than half the computation time 
(details in Table \ref{tab:NS_3pi}). 
This advantage arises from ParticleWNN's avoidance of higher-order derivative computations 
for $u$, $v$, and $p$.
\begin{table}[tbp]
    \centering
    \caption{Experiment results for the steady NS problem \eqref{eq:NS_steady}.}
    \begin{tabular}{c|ccc}\bottomrule
                       & ParticleWNN & vanilla PINN   & VPINN\\ \hline
    Relative error for $u$ & $1.34e^{-3}\pm 7.41e^{-5}$ & $3.58e^{-3}\pm 6.78e^{-4}$ & $8.99e^{-1}\pm 1.69e^{-3}$\\ 
    MAE for $u$            & $1.44e^{-2}\pm 2.04e^{3}$ & $3.50e^{-2}\pm 5.00e^{-3}$ & $9.02\pm 0.33$ \\\hline 
    Relative error for $v$ & $3.41e^{-3}\pm 1.94e^{-4}$ & $9.67e^{-3}\pm 1.16e^{-3}$ & $0.95\pm 0.03$ \\ 
    MAE for $v$            & $7.01e^{-3}\pm 1.25e^{-3}$ & $3.09e^{-2}\pm 5.74e^{-3}$ & $1.99\pm 0.06$\\\hline 
    Relative error for $p$ & $4.52e^{-4}\pm 3.60e^{-5}$ & $2.68e^{-3}\pm 4.86e^{-4}$ & $7.38e^{-1}\pm 1.10e^{-2}$ \\ 
    MAE for $p$            & $3.25e^{-2}\pm 9.22e^{-3}$ & $2.73e^{-1}\pm 5.94e^{-2}$ & $29.61\pm 2.46$ \\\hline
    Time (s)           & $3182.74\pm 2.80$ & $7703.80\pm 29.23$ & $2983.63\pm 3.27$ \\ \toprule
    \end{tabular}
    \label{tab:NS_3pi}
\end{table}
\subsection{Inverse Problems}
\begin{figure}[!htbp]
    \centering  
    \subfigure[The exact $a$]{\label{fig:inverse_coef_a}
        \includegraphics[width=0.235\textwidth]{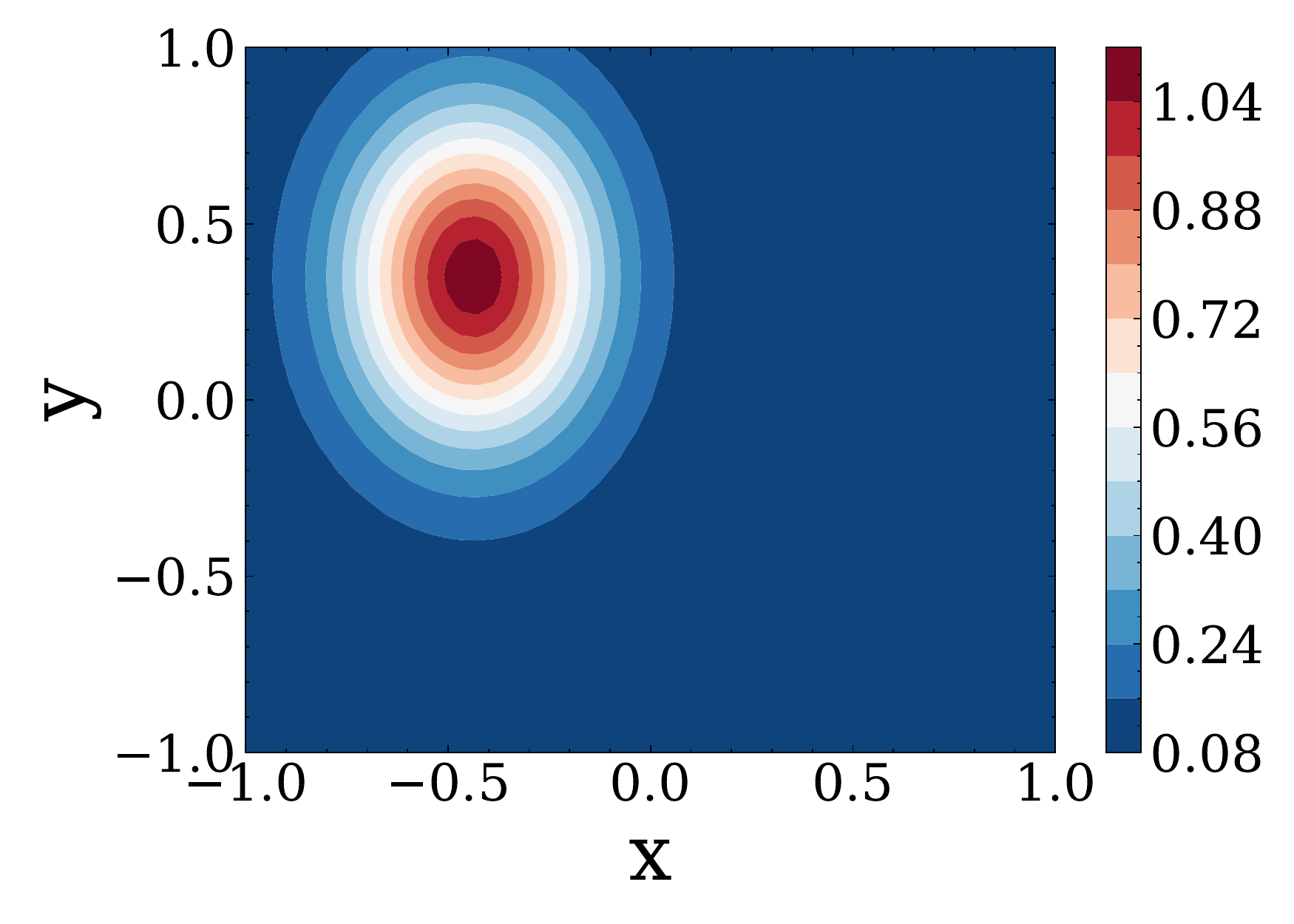}}
    \subfigure[The exact $u$ and sensors]{\label{fig:inverse_coef_b}
        \includegraphics[width=0.235\textwidth]{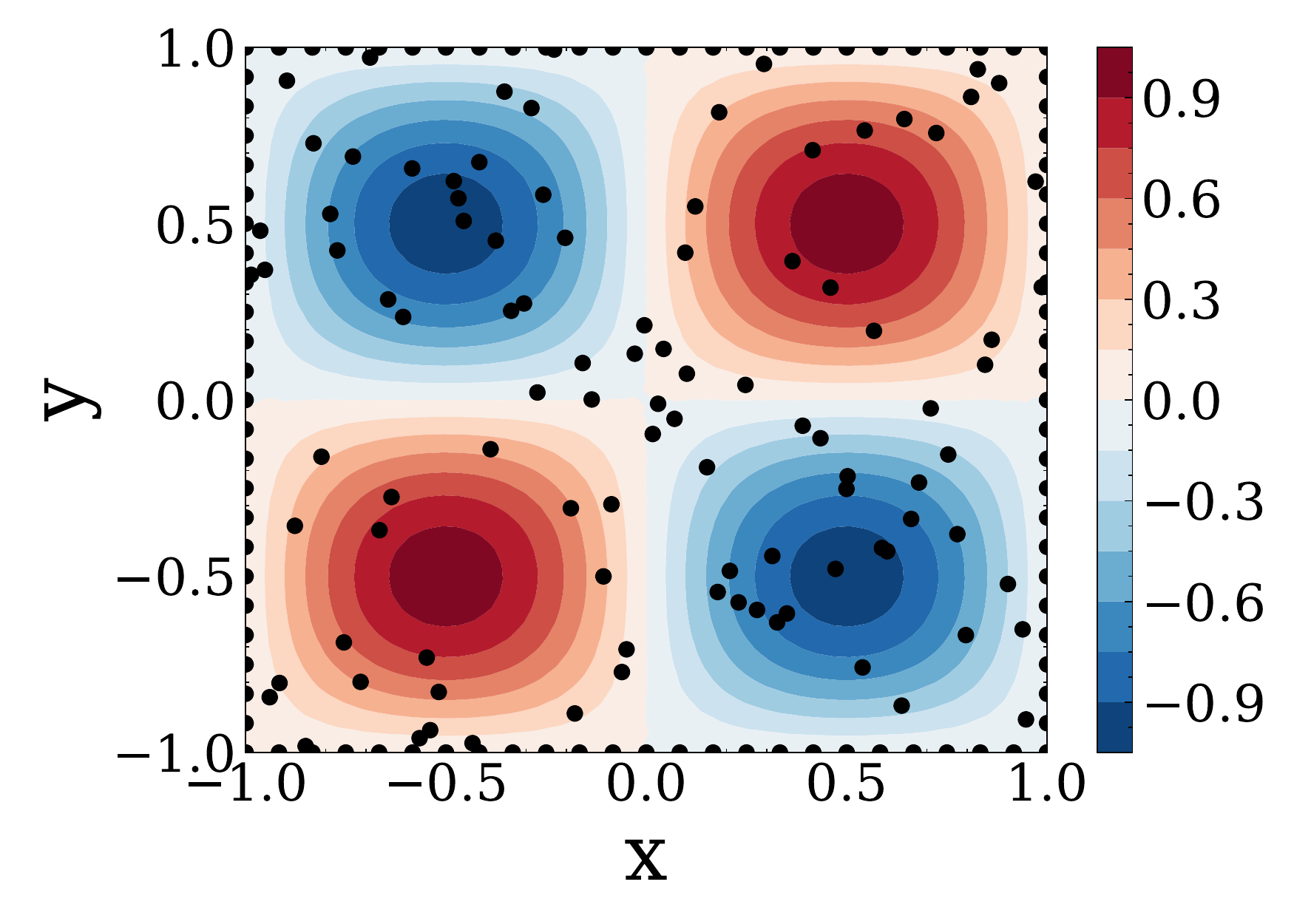}}
    \subfigure[The loss vs. time (Noise level 0.01)]{\label{fig:inverse_loss_001}
        \includegraphics[width=0.235\textwidth]{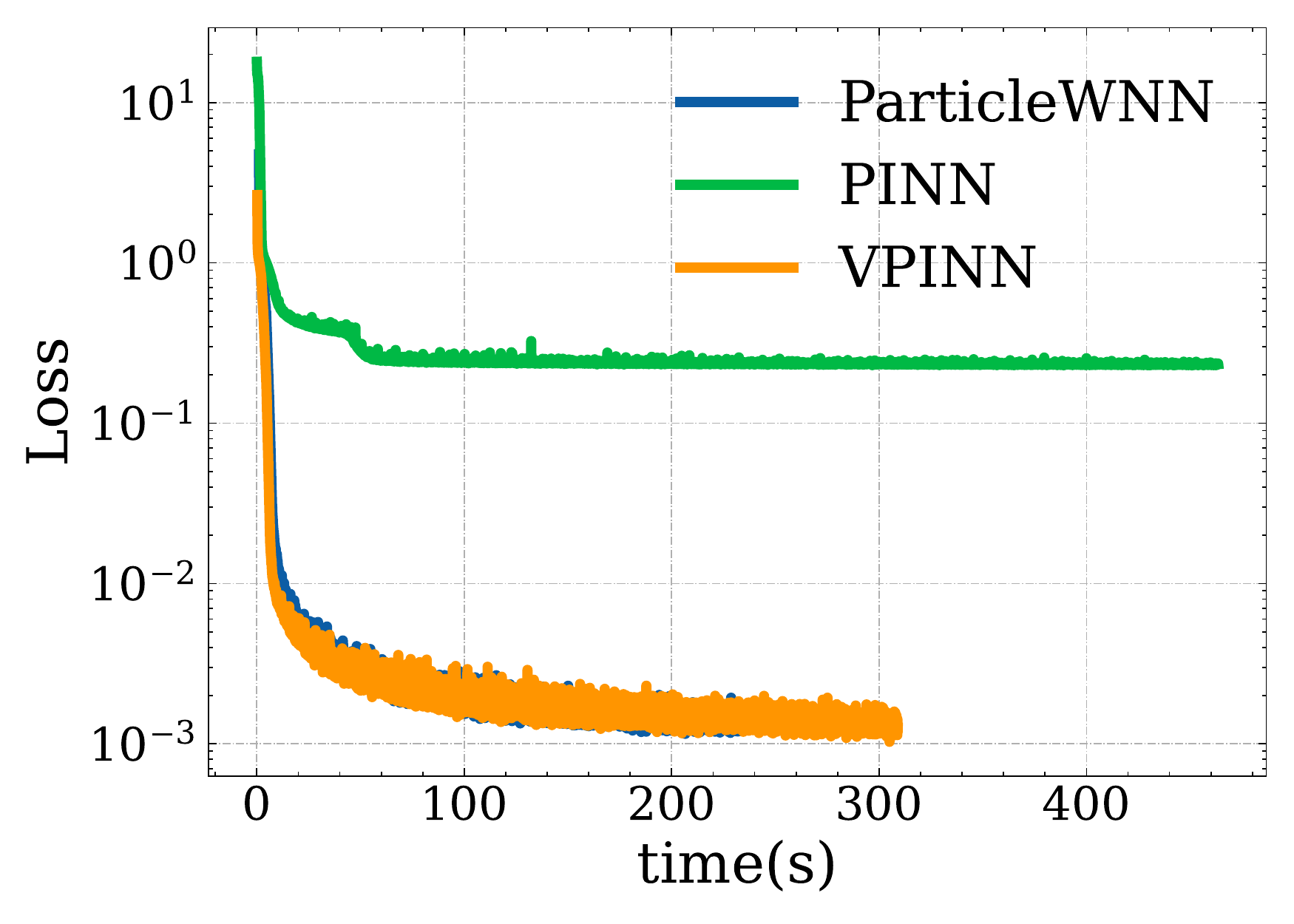}}
    \subfigure[The loss vs. time (Noise level 0.1)]{\label{fig:inverse_loss_01}
        \includegraphics[width=0.235\textwidth]{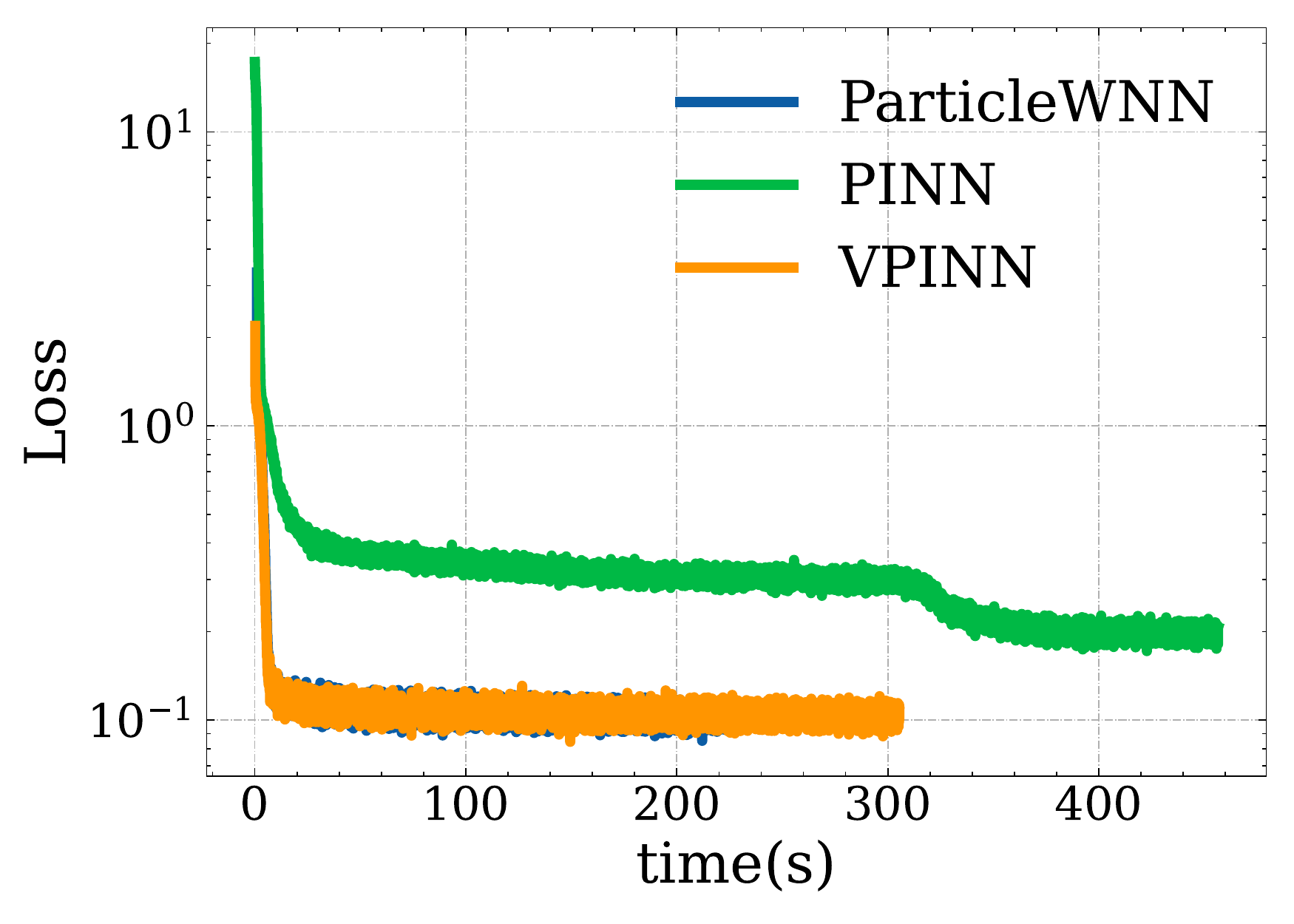}}
    \subfigure[Relative error of $u$ (Noise level 0.01)]{\label{fig:inverse_noise001_l2_u}
        \includegraphics[width=0.235\textwidth]{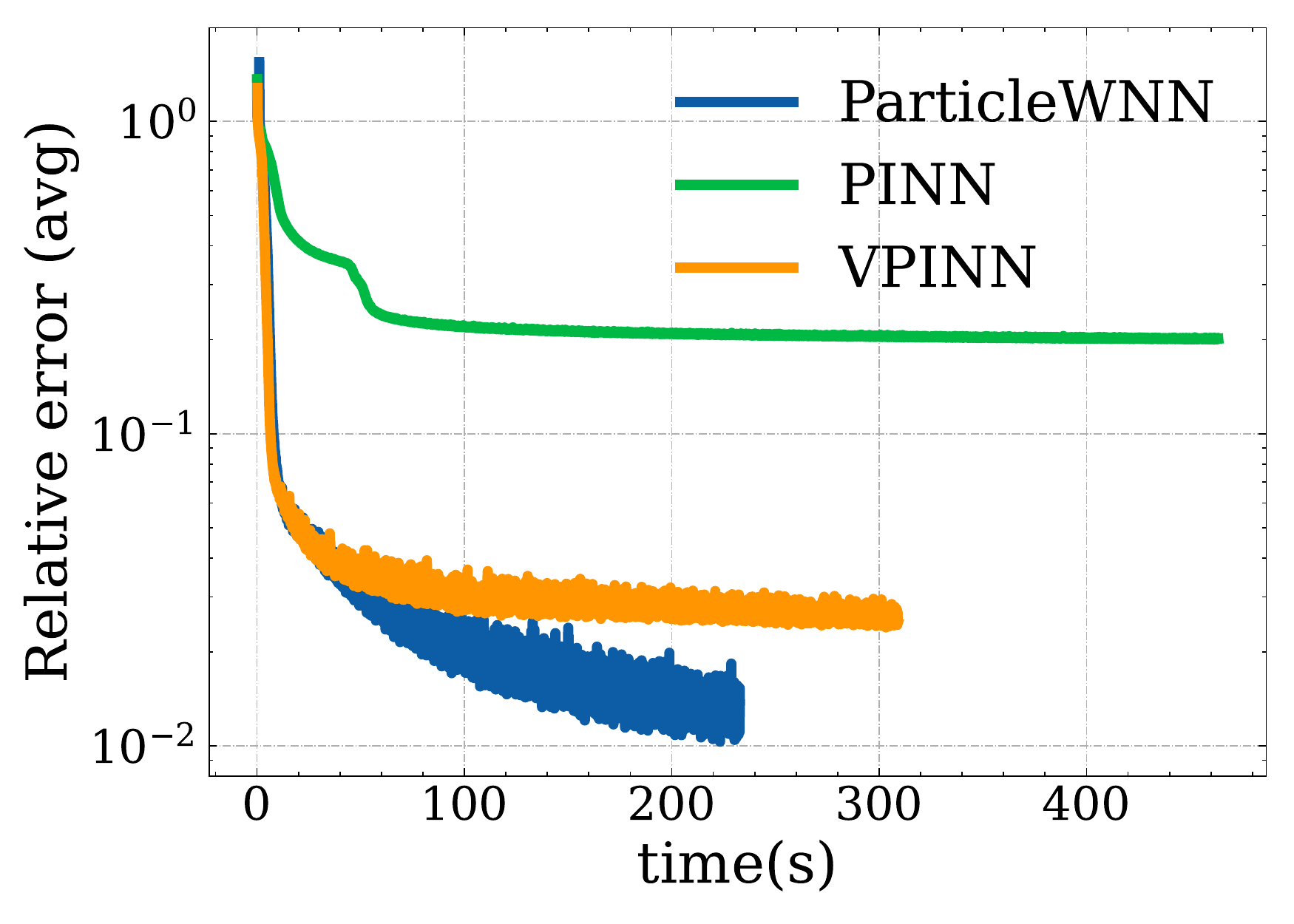}}
    \subfigure[Relative error of $a$ (Noise level 0.01)]{\label{fig:inverse_noise001_l2_a}
        \includegraphics[width=0.235\textwidth]{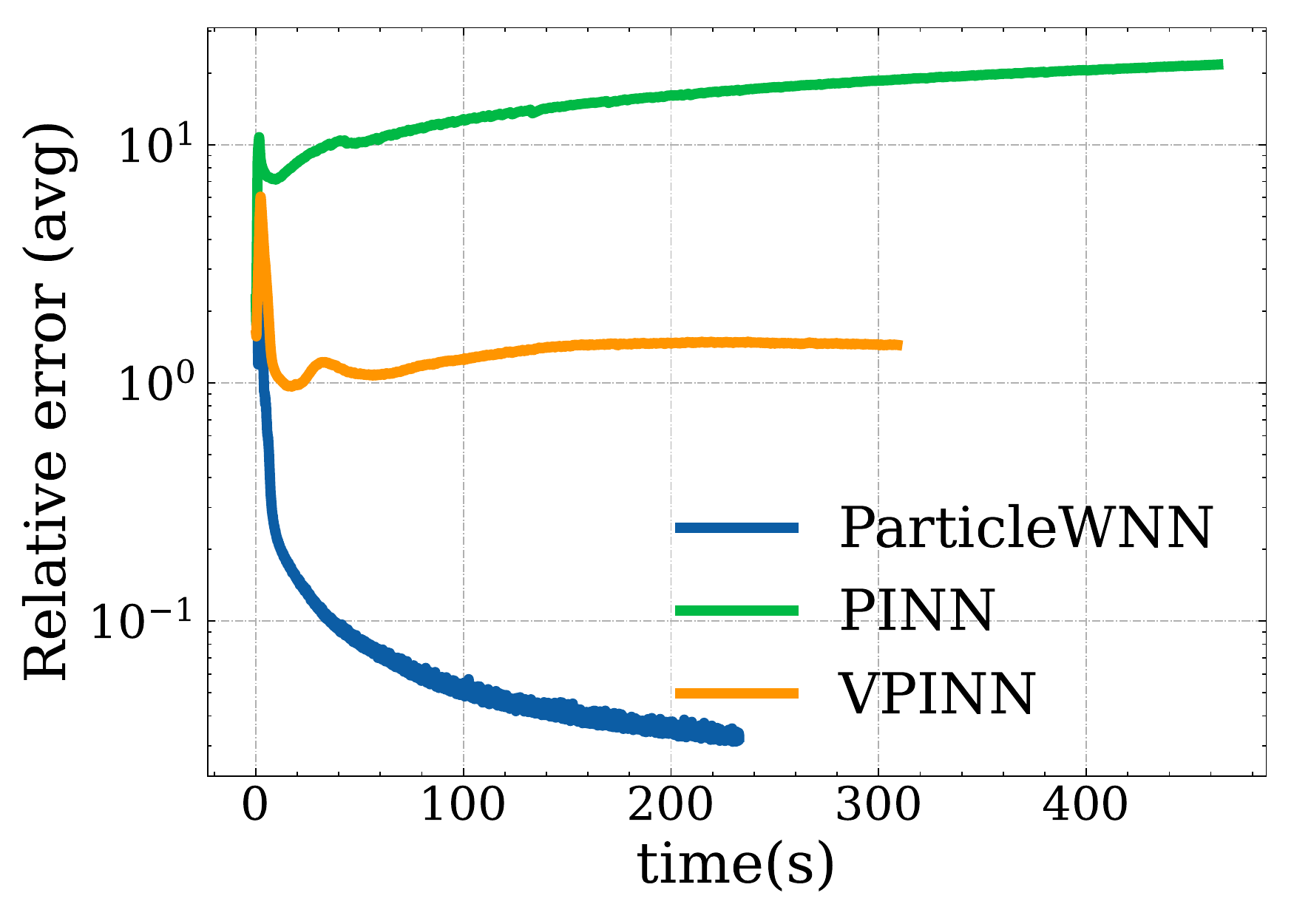}}
    \subfigure[Relative error of $u$ (Noise level 0.1)]{\label{fig:inverse_noise01_l2_u}
        \includegraphics[width=0.235\textwidth]{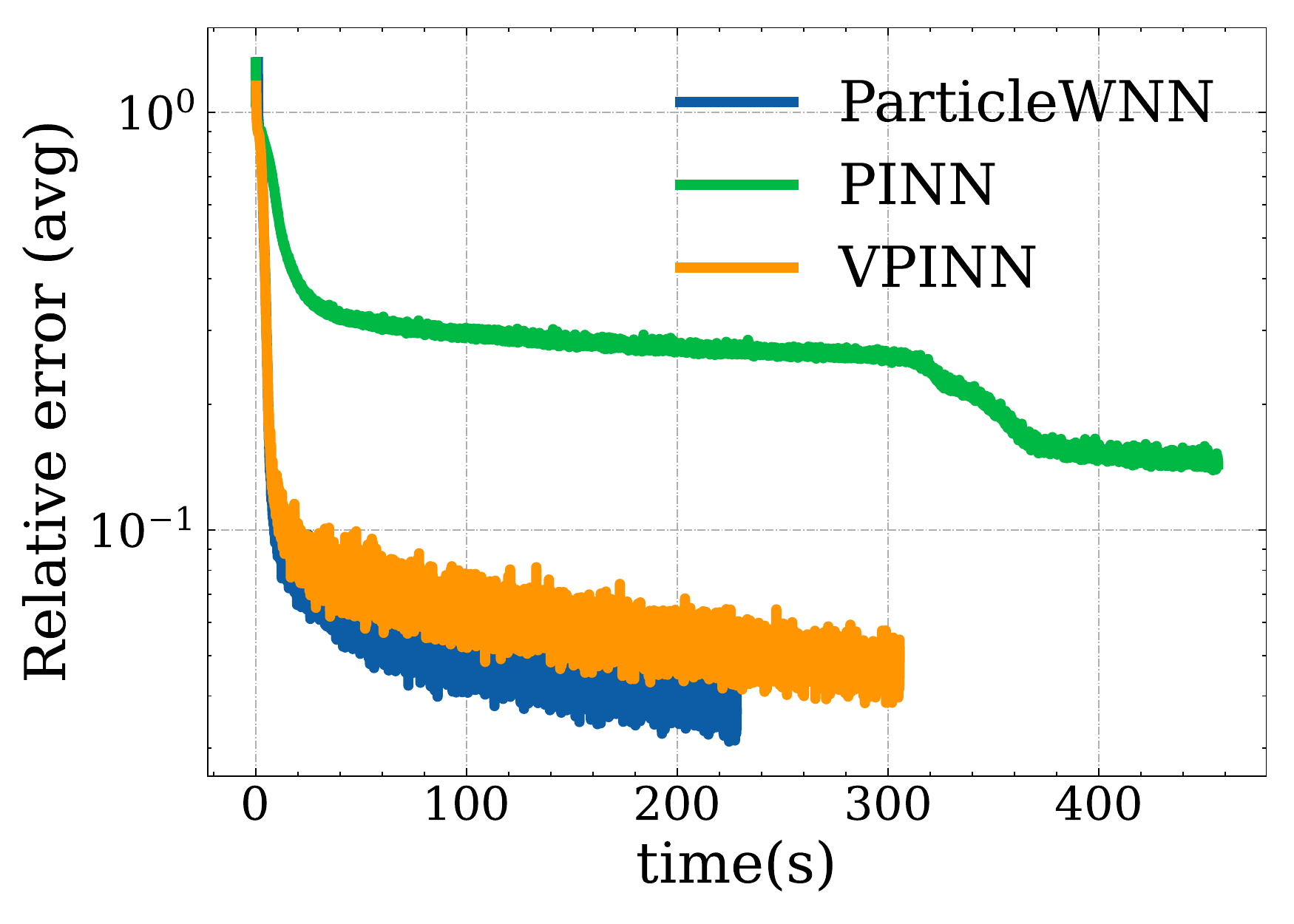}}
    \subfigure[Relative error of $a$ (Noise level 0.1)]{\label{fig:inverse_noise01_l2_a}
        \includegraphics[width=0.235\textwidth]{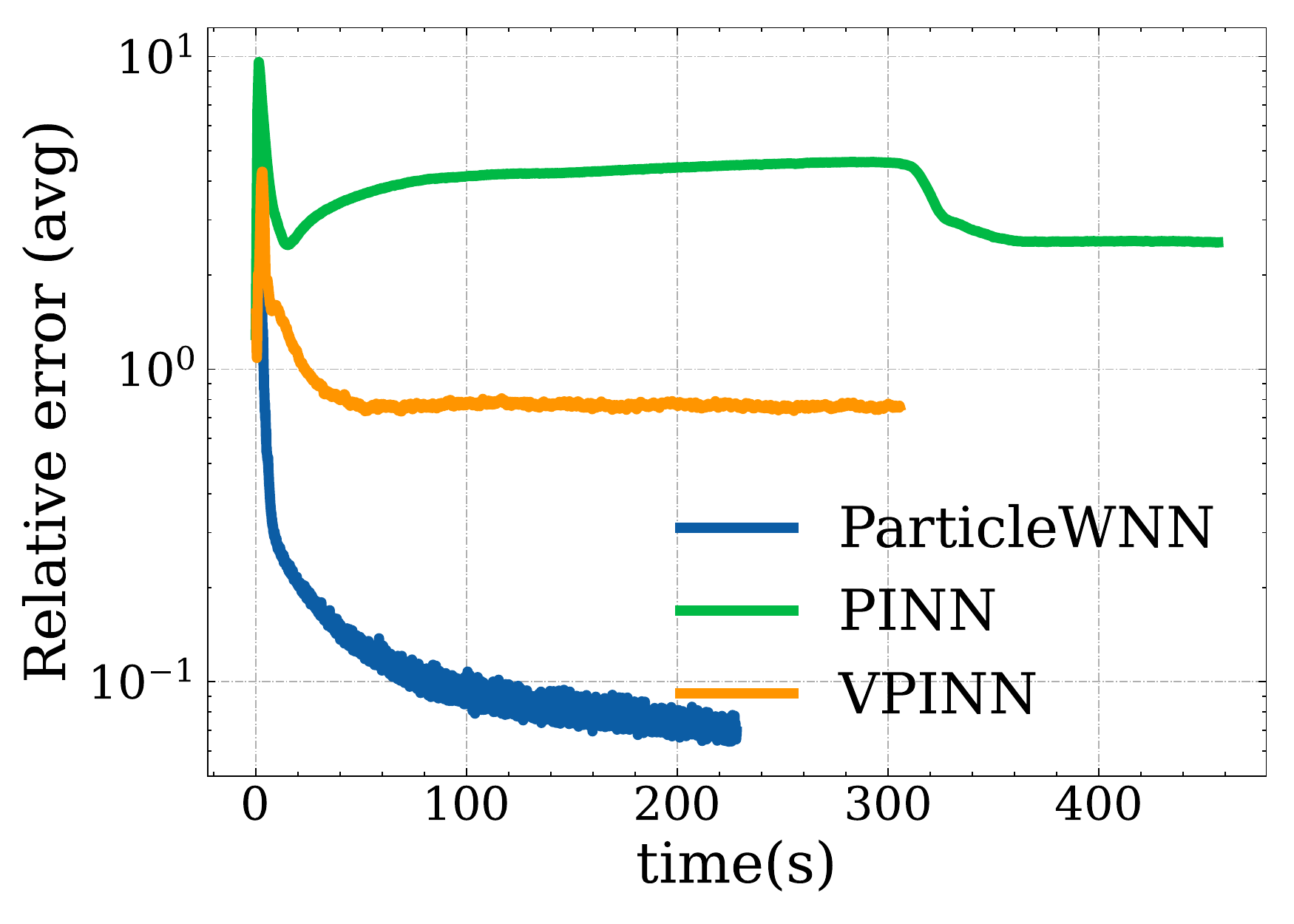}}
    \caption{The performance of different methods in solving the inverse problem 
    \eqref{eq:poisson2d_inverse} with different noise levels. 
    (a) The exact $a$; (b) The exact $u$ and sensors (black dots); 
    (c), (d) The average Loss vs. the average computation time at noise level 0.01 and 0.1, respectively;
    (e), (f) The average Relative error vs. average computation time at noise level 0.01 for $u$ and $a$, respectively;
    (g), (h) The average Relative error vs. average computation time at noise level 0.1 for $u$ and $a$, respectively.}
    \label{fig:inverse_coef}
\end{figure}

The particleWNN can also be applied to solve inverse problems.
Typically, we consider the following coefficient identification problem often 
used to verify the effectiveness of DNN-based methods:
\begin{equation}\label{eq:poisson2d_inverse}
        -\nabla \cdot(a(x,y) \nabla u) = f(x,y), \quad (x,y) \in\Omega=[-1,1]\times[-1,1],
\end{equation}
where $u$ is the solution of the equation, $f$ indicates the source term, and
$a\in\{a\in L^{\infty}(\Omega):0<\underline{a}\leq a(x,y)\leq\overline{a}\ a.e.\ in\ \Omega\}$
represents the coefficient.
In the inverse problem, given the source term $f$, the objective is to identify the 
coefficient $a$ with inexact measurements $u^{\delta}$ 
where $\delta$ indicates Gaussian noise.
We set the exact coefficient as
\begin{equation}
a(x,y) = 0.1 + \exp^{-\frac{(x-x_1)^2}{\sigma_1}+\frac{(y-y_1)^2}{\sigma_2}},
\end{equation}
where $\sigma_1,\sigma_2$ are sampled from $[0.01,0.5]$, and $x_1,y_1$ are sampled from $[-0.5,0.5]$, 
as shown in Figure \ref{fig:inverse_coef_a}.
For convenience, we manufacture a solution $u(x,y) = \sin(\pi x)\sin(\pi y)$ and use the exact 
coefficient and solution to compute $f$.
Assuming that measurements $u^{\delta}$ can be obtained from 100 sensors inside the domain and 
$\tilde{N}_{bd}=25$ equally distributed sensors at each boundary for the Dirichlet boundary condition 
(see Figure \ref{fig:inverse_coef_b}). 
To solve the problem, we use two independent DNNs with the same structure to parameterize 
$a$ and $u$, respectively.
We set $N_p=200$, $topK=150$, $K_{int}=60$, and use Tanh as the activation 
for the ParticleWNN method.
For a fair comparison, we use $12,000$ ($topK=9000$) collocation points for the vanilla PINN and keep 
other settings consistent with the ParticleWNN.
For the VPINN method, we choose $N_{test}=50$ test functions, $N_{int}=225$ integration points, and keep 
other settings consistent with ParticleWNN.

\begin{figure}[!htbp]
    \centering  
    \subfigure[$u_{NN}$ obtained by ParticleWNN (Noise level 0.01)]{\label{fig:inverse_u_particlewnn_001}
        \includegraphics[width=0.235\textwidth]{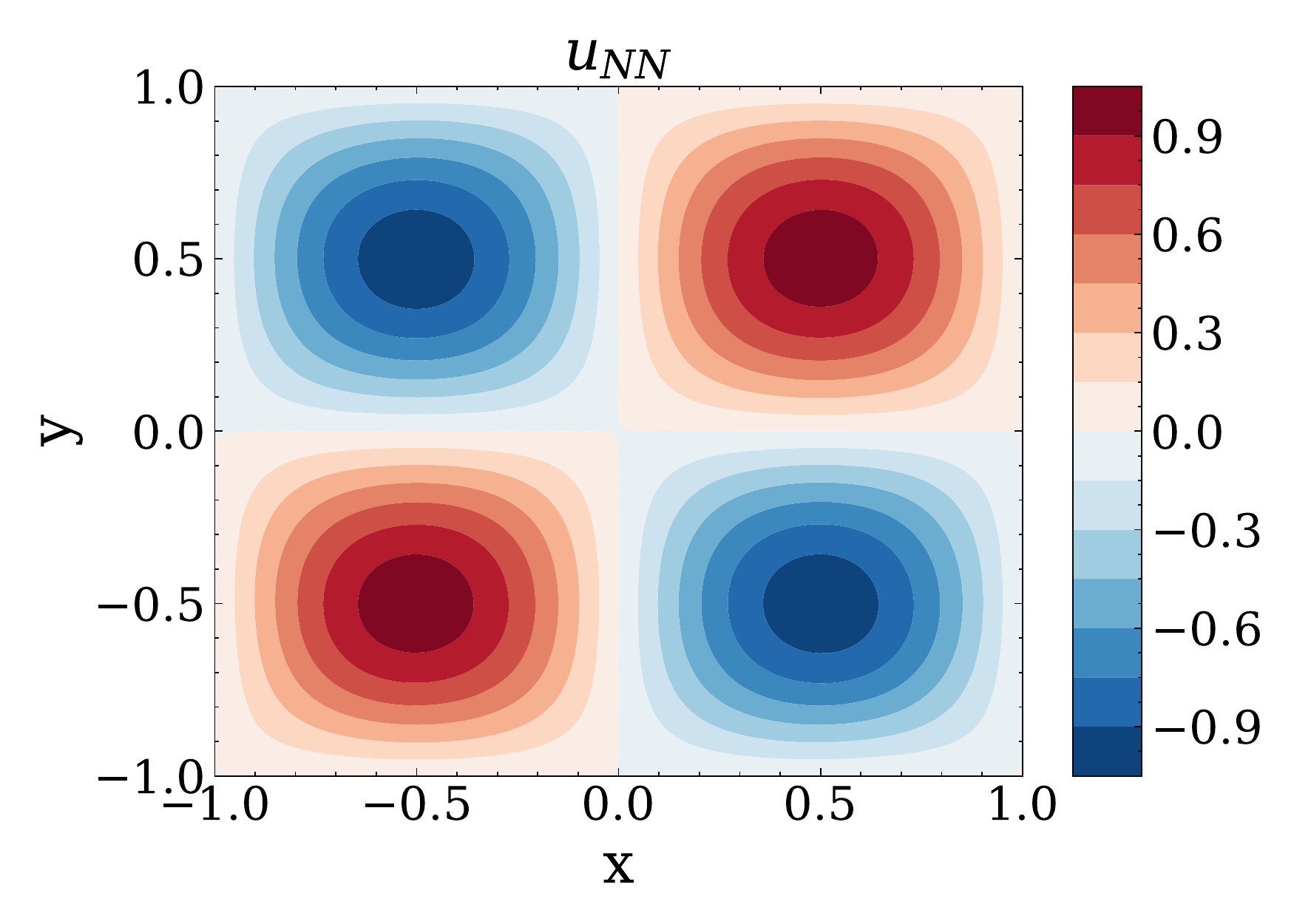}}
    \subfigure[$a_{NN}$ obtained by ParticleWNN (Noise level 0.01)]{\label{fig:inverse_a_particlewnn_001}
        \includegraphics[width=0.235\textwidth]{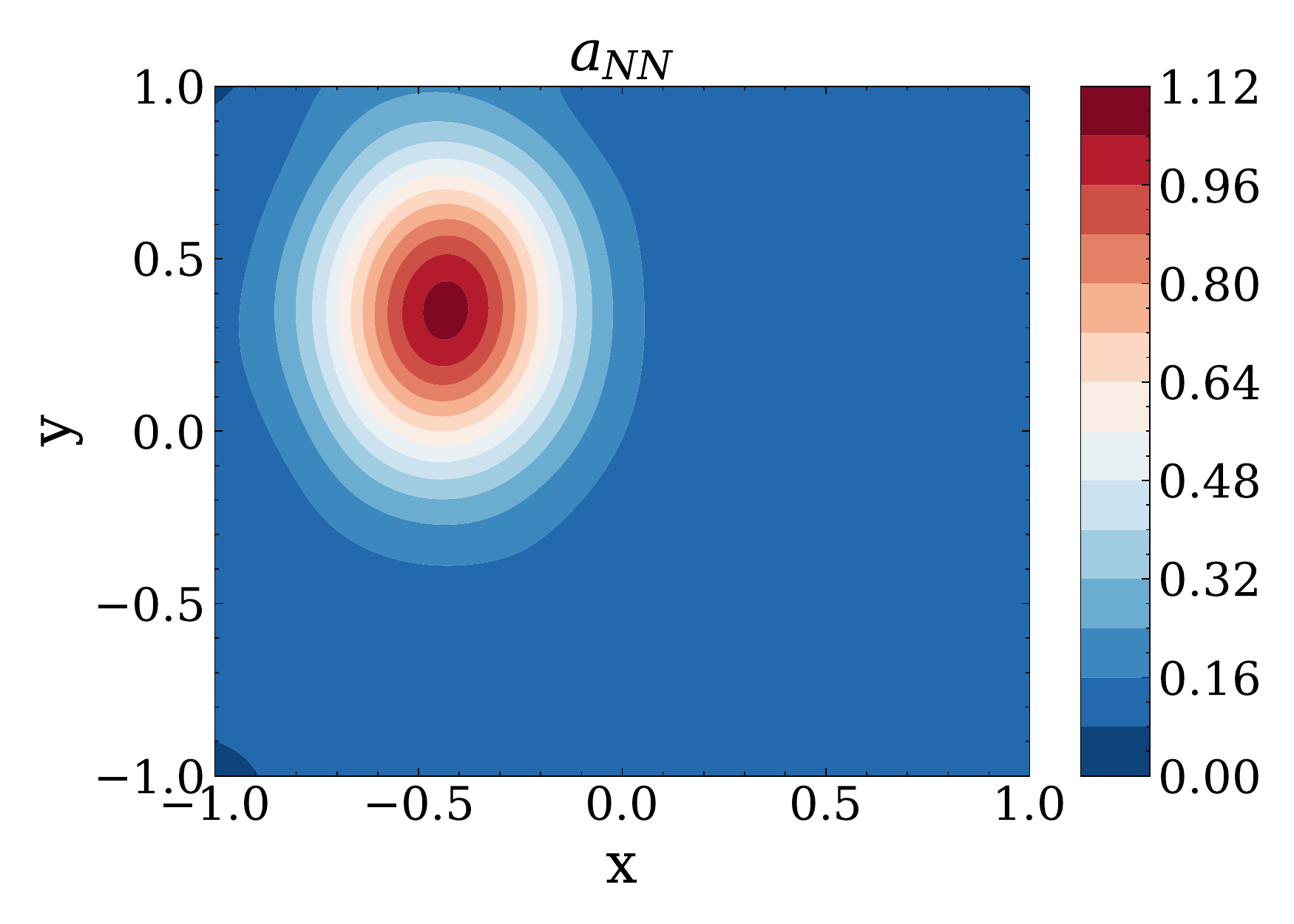}}
    \subfigure[$u_{NN}$ obtained by ParticleWNN (Noise level 0.1)]{\label{fig:inverse_u_particlewnn_01}
        \includegraphics[width=0.235\textwidth]{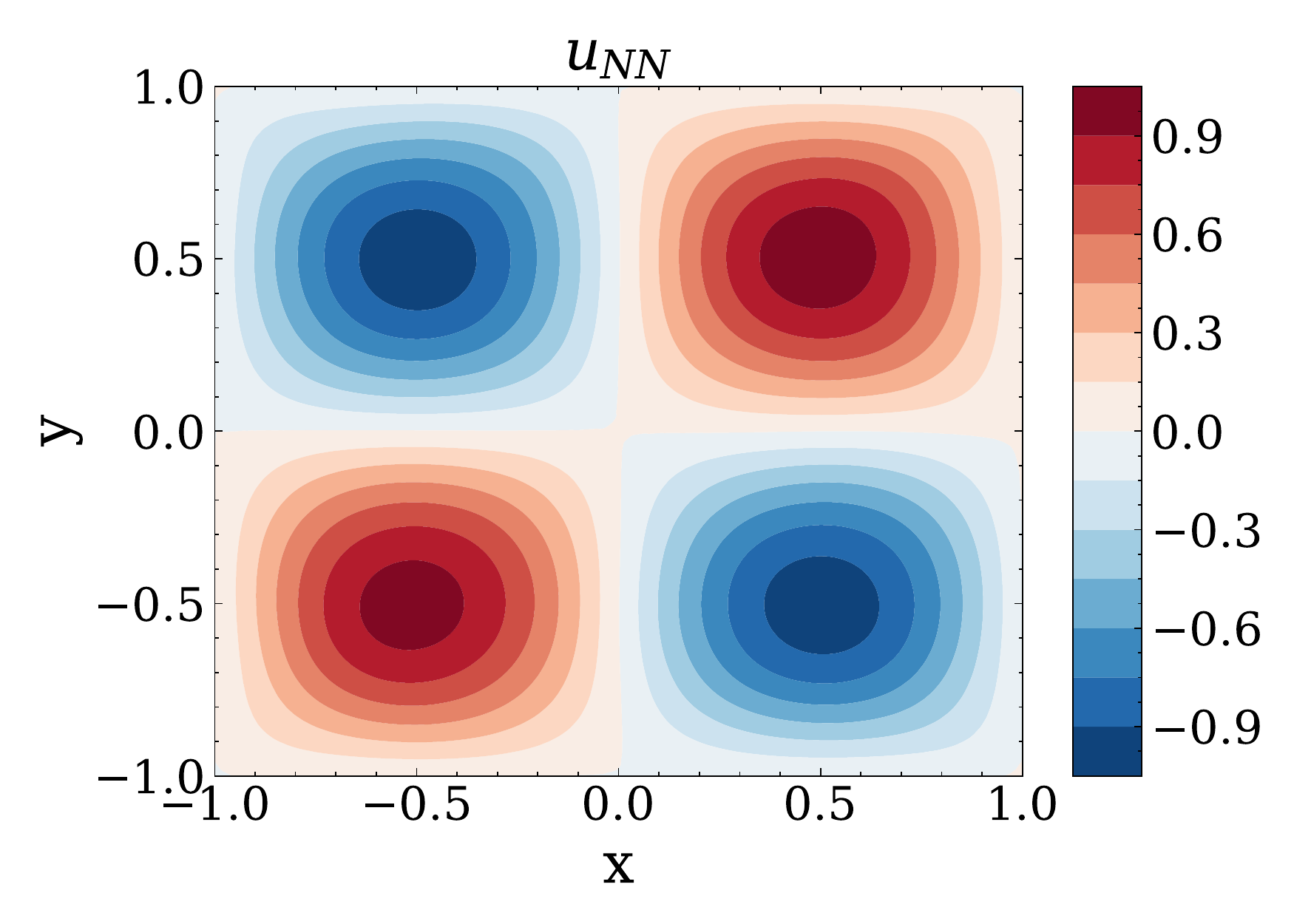}}
    \subfigure[$a_{NN}$ obtained by ParticleWNN (Noise level 0.1)]{\label{fig:inverse_a_particlewnn_01}
        \includegraphics[width=0.235\textwidth]{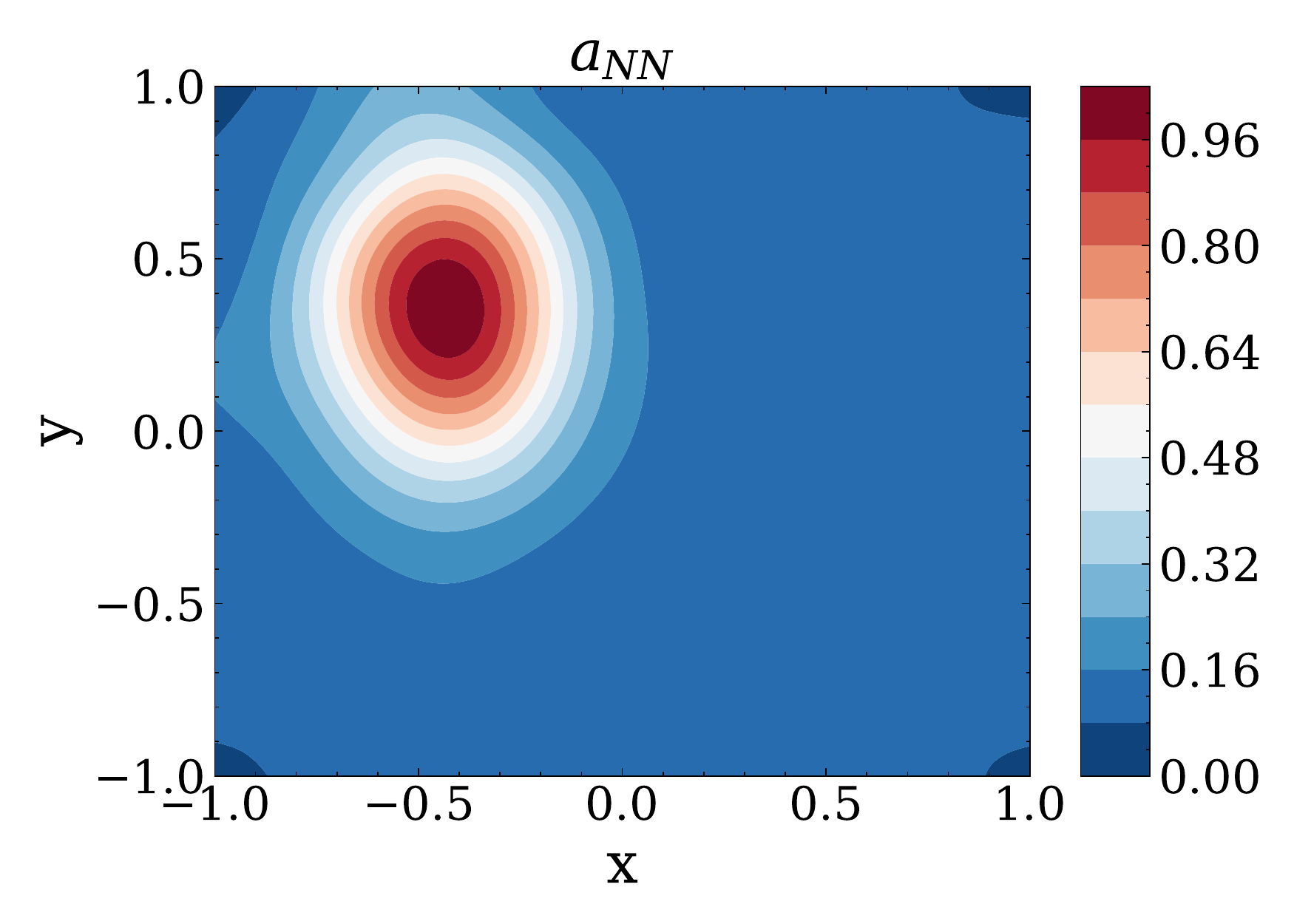}}
    \subfigure[$u_{NN}$ obtained by vanilla PINN (Noise level 0.01)]{\label{fig:inverse_u_pinn_001}
        \includegraphics[width=0.235\textwidth]{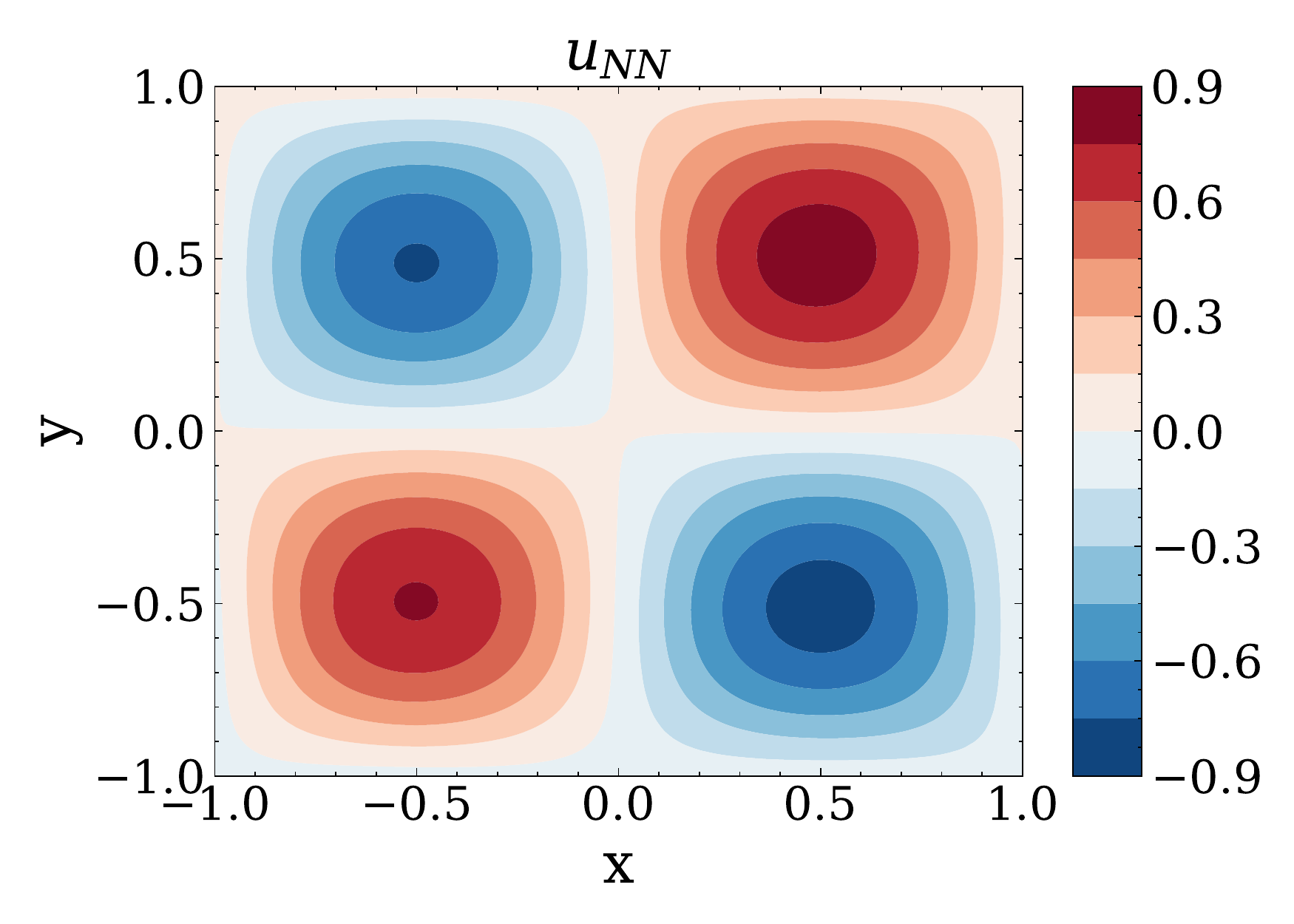}}
    \subfigure[$a_{NN}$ obtained by vanilla PINN (Noise level 0.01)]{\label{fig:inverse_a_pinn_001}
        \includegraphics[width=0.235\textwidth]{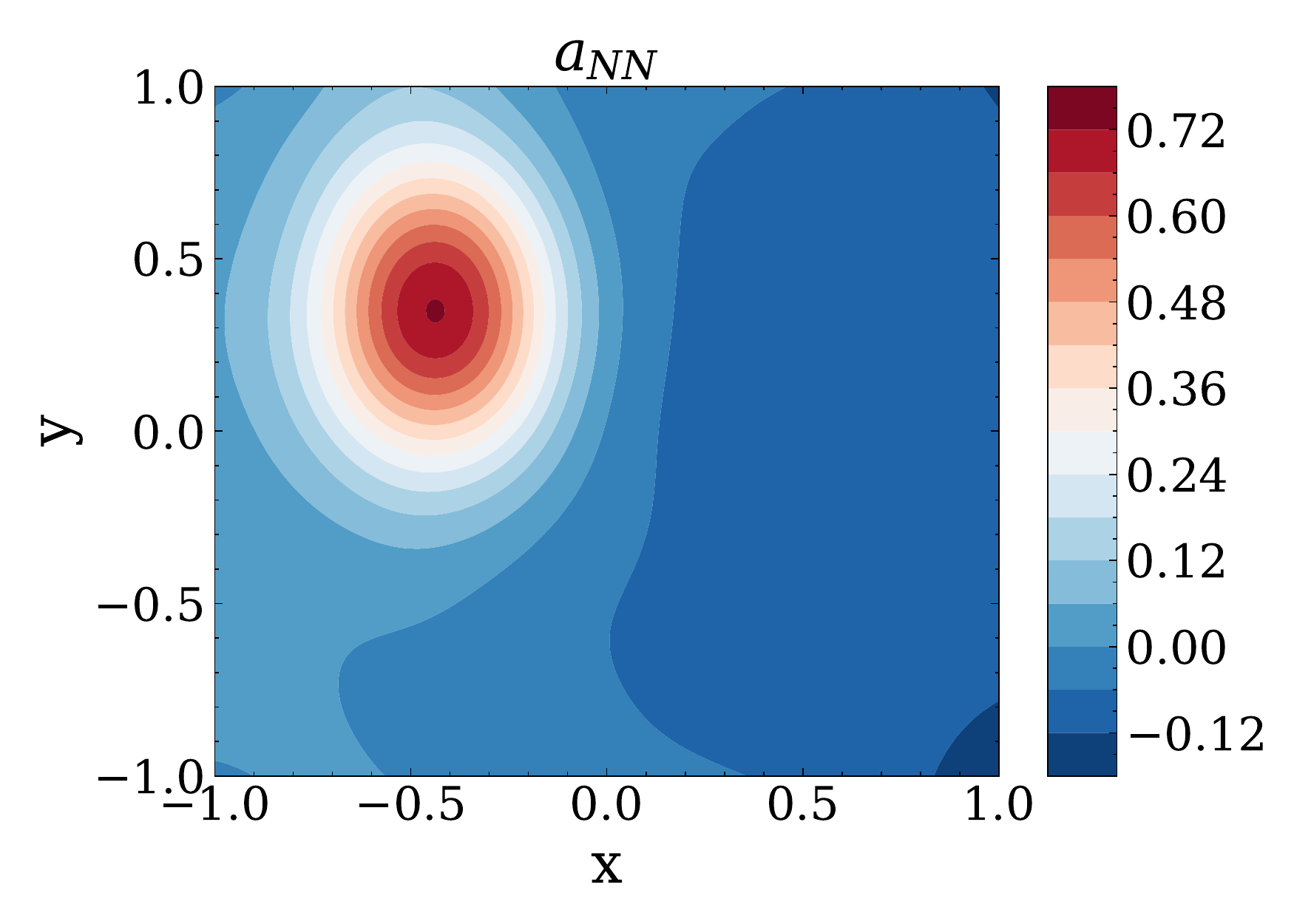}}
    \subfigure[$u_{NN}$ obtained by vanilla PINN (Noise level 0.1)]{\label{fig:inverse_u_pinn_01}
        \includegraphics[width=0.235\textwidth]{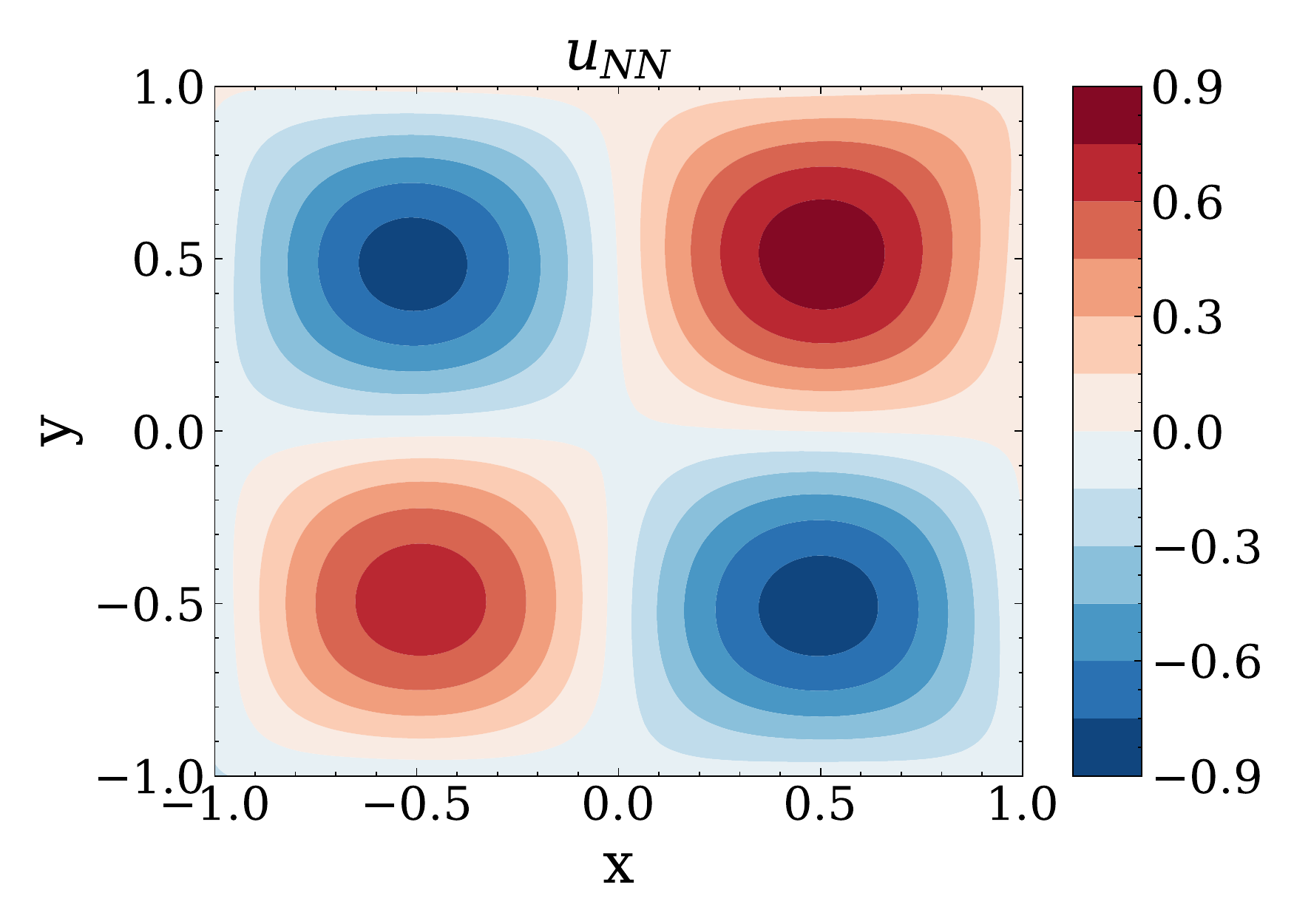}}
    \subfigure[$a_{NN}$ obtained by vanilla PINN (Noise level 0.1)]{\label{fig:inverse_a_pinn_01}
        \includegraphics[width=0.235\textwidth]{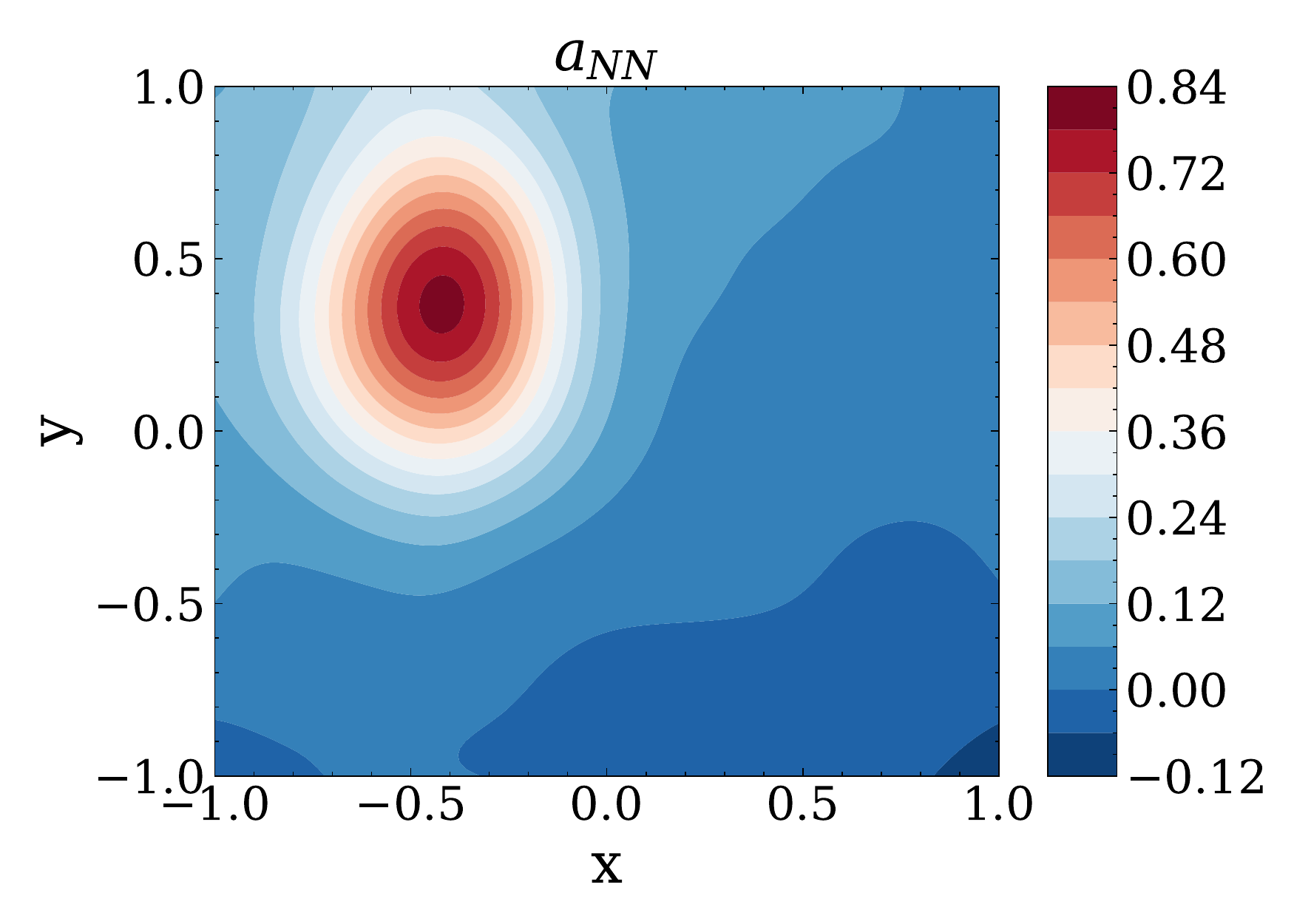}}
    \subfigure[$u_{NN}$ obtained by VPINN (Noise level 0.01)]{\label{fig:inverse_u_vpinn_001}
        \includegraphics[width=0.235\textwidth]{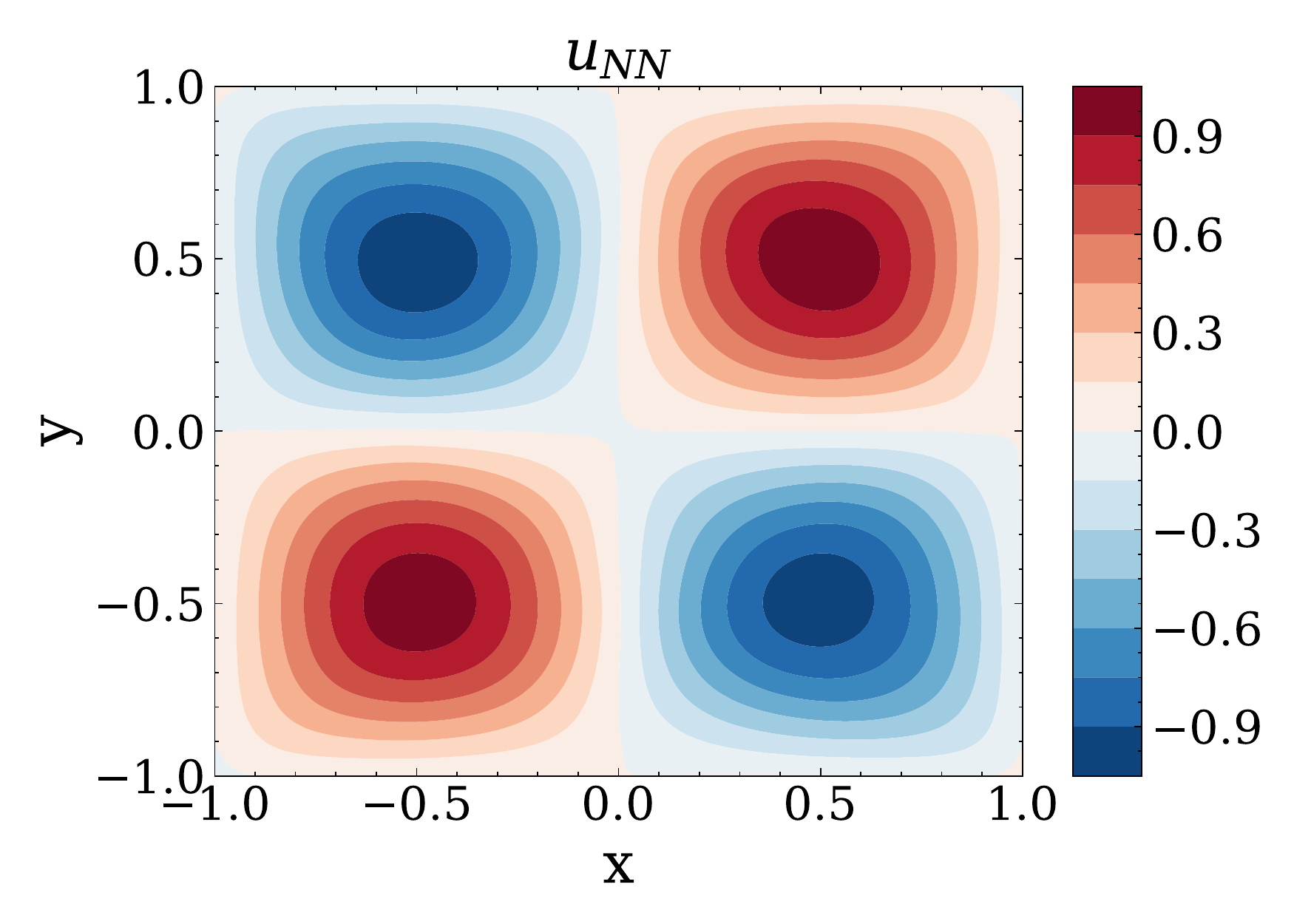}}
    \subfigure[$a_{NN}$ obtained by VPINN (Noise level 0.01)]{\label{fig:inverse_a_vpinn_001}
        \includegraphics[width=0.235\textwidth]{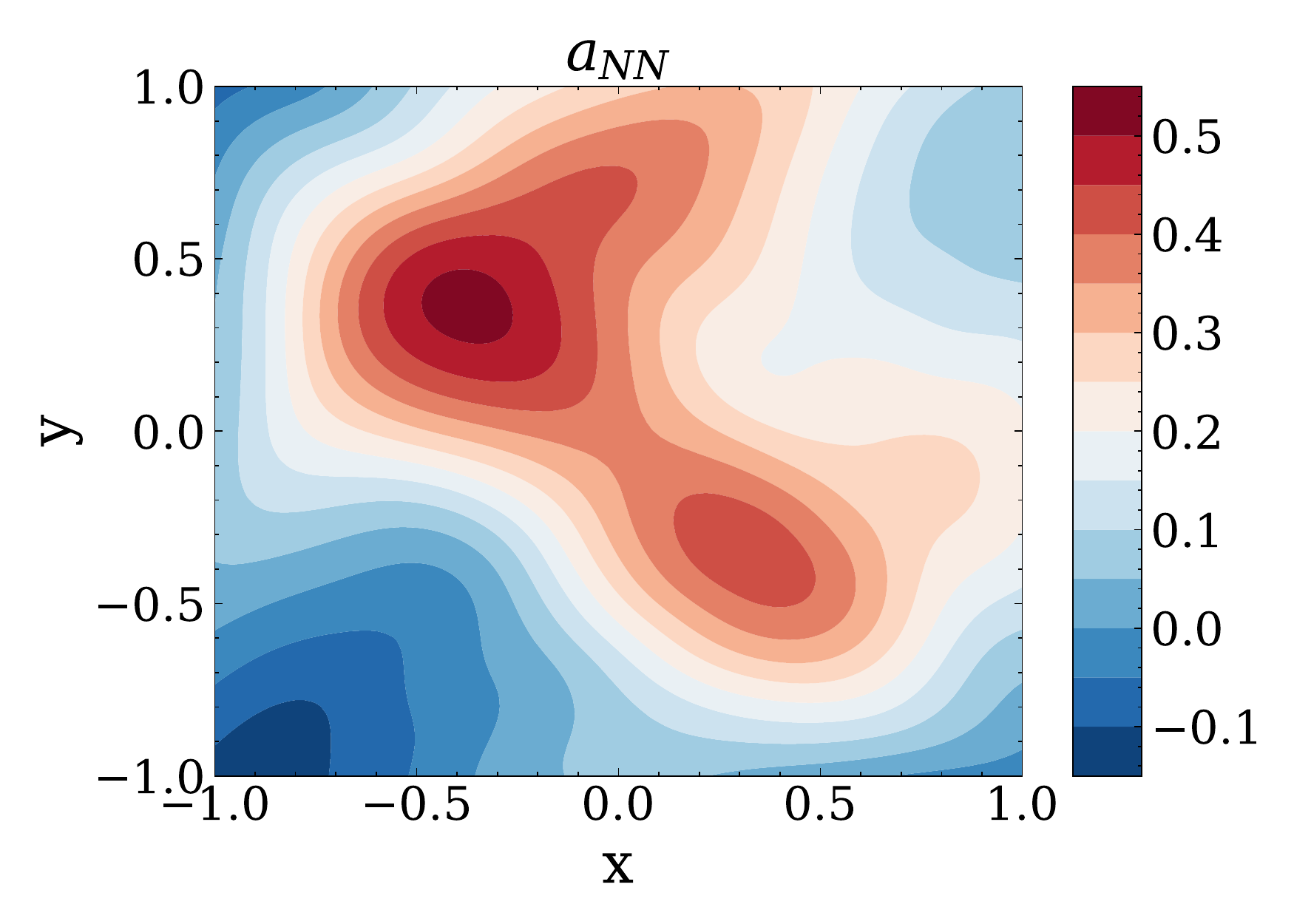}}
    \subfigure[$u_{NN}$ obtained by VPINN (Noise level 0.1)]{\label{fig:inverse_u_vpinn_01}
        \includegraphics[width=0.235\textwidth]{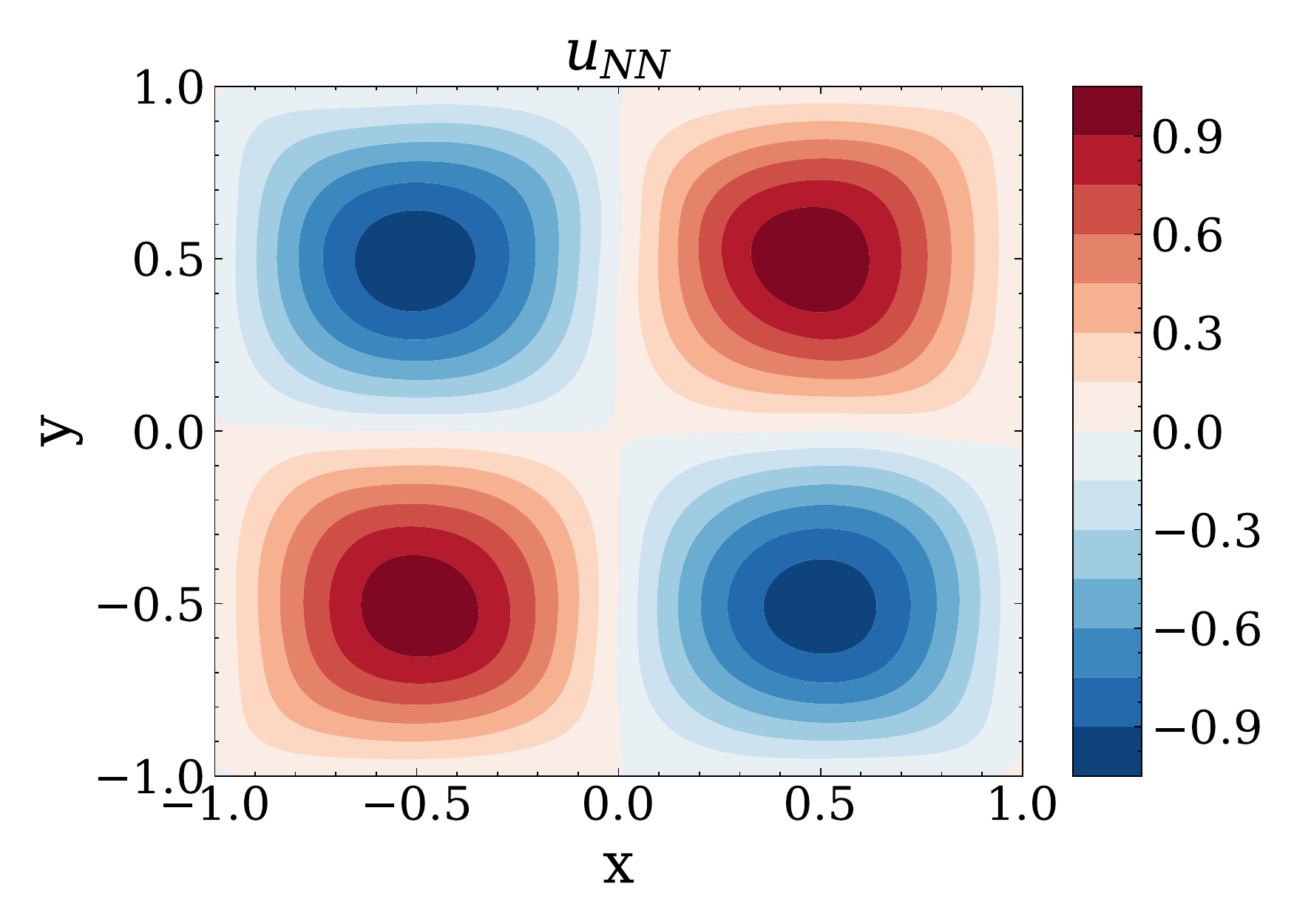}}
    \subfigure[$a_{NN}$ obtained by VPINN (Noise level 0.1)]{\label{fig:inverse_a_vpinn_01}
        \includegraphics[width=0.235\textwidth]{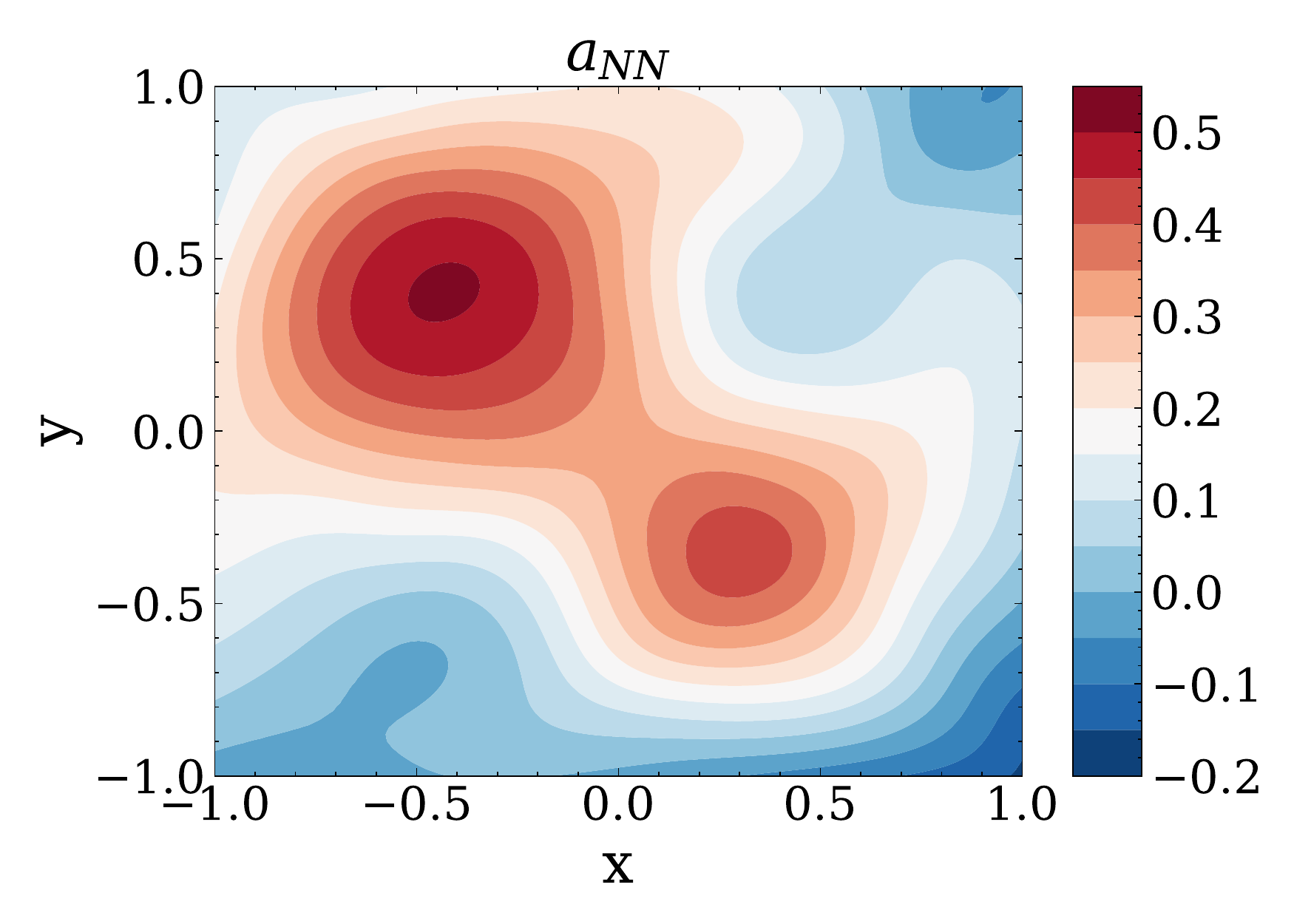}}
    \vspace{-0.25cm}
    \caption{The performance of different methods in solving the inverse problem \eqref{eq:poisson2d_inverse}. 
    (a), (b), (c), (d) The average $u_{NN}$, $a_{NN}$ obtained by the ParticleWNN at noise level 0.01 and 0.1, respectively; 
    (e), (f), (g), (h) The average $u_{NN}$, $a_{NN}$ obtained by the vanilla PINN at noise level 0.01 and 0.1, respectively; 
    (i), (j), (k), (l) The average $u_{NN}$, $a_{NN}$ obtained by the VPINN at noise level 0.01 and 0.1, respectively.
    } 
    \label{fig:inverse_pred}
\end{figure}

We consider two different noise levels $\delta\sim\mathcal{N}(0,0.01^2),\ \delta\sim\mathcal{N}(0,0.1^2)$.
The experiment results are recorded in Table \ref{tab:inverse_problem}, and plots of the Relative error, 
MAE, and loss with respect to time are shown in Figure \ref{fig:inverse_coef}. 
The predicted coefficients and solutions obtained by different methods are illustrated in 
Figure \ref{fig:inverse_pred}.
From Table \ref{tab:inverse_problem}, we observe that the vanilla PINN and the VPINN are very sensitive to noise, 
while the ParticleWNN achieves good results at both noise levels.
Moreover, ParticleWNN converges much faster among these methods and takes less than half of the 
computation time of the vanilla PINN. 
This is also evident in Figure \ref{fig:inverse_coef}.
Figure \ref{fig:inverse_pred} also indicates that, although VPINN can obtain the solution very well, 
it struggles to invert the coefficients effectively.

\begin{table}[tbp]
    \centering
    \caption{Experiment results for the inverse problem \eqref{eq:poisson2d_inverse}.}
    \begin{tabular}{c|c|ccc}\bottomrule
        \multicolumn{2}{c|}{}      & ParticleWNN & vanilla PINN  & VPINN           \\ \hline
    \multirow{3}{*}{Noise level 0.01} & {Relative error for $a$} & $0.030\pm 0.003$ & $0.663\pm 1.224$ & $0.858\pm 0.475$\\ 
                               {} &  {MAE for $a$ }         & $0.057\pm 0.006$ & $0.384\pm 0.612$ & $0.656\pm 0.145$\\ 
                              {}  & {Relative error for $u$} & $0.009\pm 0.001$ & $0.199\pm 0.367$ & $0.022\pm 0.007$\\ 
                              {}  & {MAE for $u$}            & $0.018\pm 0.003$ & $0.274\pm 0.477$ & $0.082\pm 0.037$\\ 
                              {}   & {Time (s)}           & $232.82\pm 4.29$ & $463.48\pm 2.09$ & $308.96\pm 1.28$ \\ \hline
    \multirow{3}{*}{Noise level 0.1} & {Relative error for $a$} & $0.058\pm 0.006$ &$0.337\pm 0.519$ & $0.645\pm 0.080$\\ 
                              {} & {MAE for $a$}           & $0.086\pm 0.008$ &$0.309\pm 0.382$ & $0.600\pm 0.059$\\ 
                              {} & {Relative error for $u$} & $0.024\pm 0.002$  &$0.135\pm 0.218$ & $0.031\pm 0.003$\\ 
                              {} & {MAE for $u$}            & $0.049\pm 0.007$  &$0.335\pm 0.569$ & $0.093\pm 0.036$\\ 
                              {} & {Time (s)}              & $228.15\pm 2.04$ &$456.52\pm 0.54$ & $305.46\pm 0.13$\\ \toprule
    \end{tabular}
    \label{tab:inverse_problem}
\end{table}
\subsection{A high-dimension example}
Finally, we demonstrate ParticleWNN's capability to handle high-dimensional problems 
by solving the following PDE with Dirichlet boundary conditions:
\begin{equation}\label{eq:high_dim}
    -\Delta u = f, \quad \bm{x}\in[-1,1]^d
\end{equation}
where $f=\sum^{d}_{i=1}x_i\sin(\pi x_{i+1})$ and $d=5$. 
In the ParticleWNN method, we employ $N_p=80$, $K_{int}=1233$, and $\tilde{N}_{bd}=100$.
We use Tanh as the activation for the DNN model and keep 
other settings consistent with those in Section \ref{sec:poisson1d}.
For fairness, we generate $1233*80$ integration points for the DeepRitz method and
$1233*80$ collocation points for the vanilla PINN method.
Additionally, we use $N_{test}=50$ test functions and $N_{int}=1973$ integration points 
for the VPINN method, while keeping other parameters consistent with the ParticleWNN method.

After $maxIter=20000$ iterations, the performance of each method is detailed 
in Table \ref{tab:high_dimension}.
From this table, we can see that the ParticleWNN method achieves both the smallest error 
and the shortest computation time.
\begin{table}[!ht]\small
\centering
\caption{Experiment results for the high-dimension problem \eqref{eq:high_dim} with $d=5$.}
\begin{tabular}{c|c|c|c|cc} \bottomrule
                    {} & ParticleWNN &  vanilla PINN  & VPINN & DeepRitz \\ \hline
    {Relative error} & $0.033\pm0.002$ & $0.036\pm0.002$ & $0.038\pm 0.001$ & $0.431\pm 0.001$   \\
    {MAE} & $0.305\pm 0.031$ & $0.270\pm 0.024$ &$0.358\pm 0.043$ & $1.364\pm 0.049$\\ 
    {Time(s)} & $559.79\pm 1.61$ & $3994.45\pm 2.64$ & $589.31\pm 2.50$  & $643.71\pm 15.74$ \\\toprule
\end{tabular}
\label{tab:high_dimension}
\end{table}
We also test ParticleWNN's capacity to address higher-dimensional problems by 
considering the problem \eqref{eq:high_dim} with $d=10$.  
To adapt to this higher dimension, we reduce the number of neurons in each layer 
of the DNN to $25$ while keeping the network's overall structure unchanged.
We then set $N_p=80$, $N_{int}=1161$, and $\tilde{N}_{bd}=100$, 
with all other parameters remaining the same.
After $maxIter=50000$ iterations, the Relative error and MAE obtained by the ParticleWNN 
are $0.049\pm 0.001$ and $0.783\pm 0.079$, respectively. 
This result demonstrates ParticleWNN's robust performance in handling higher-dimensional problems.
%
\subsection{Additional ablations}
\label{sec:add_ablation}
\begin{figure}[!htbp]
    \centering  
    \subfigure[Relative error]{\label{fig:Np_Kint_l2}
        \includegraphics[width=0.32\textwidth]{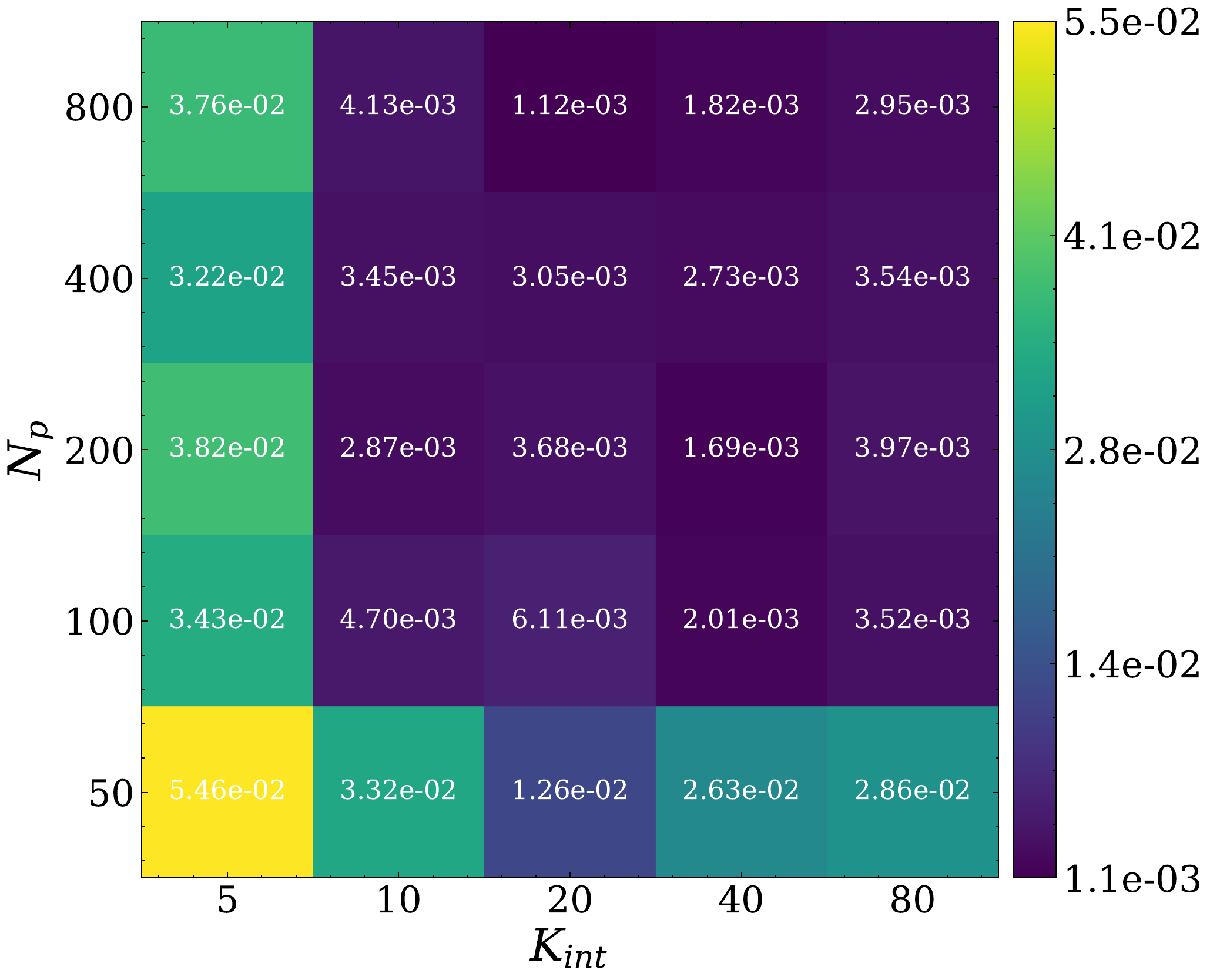}}
    \subfigure[MAE]{\label{fig:Np_Kint_abs}
        \includegraphics[width=0.32\textwidth]{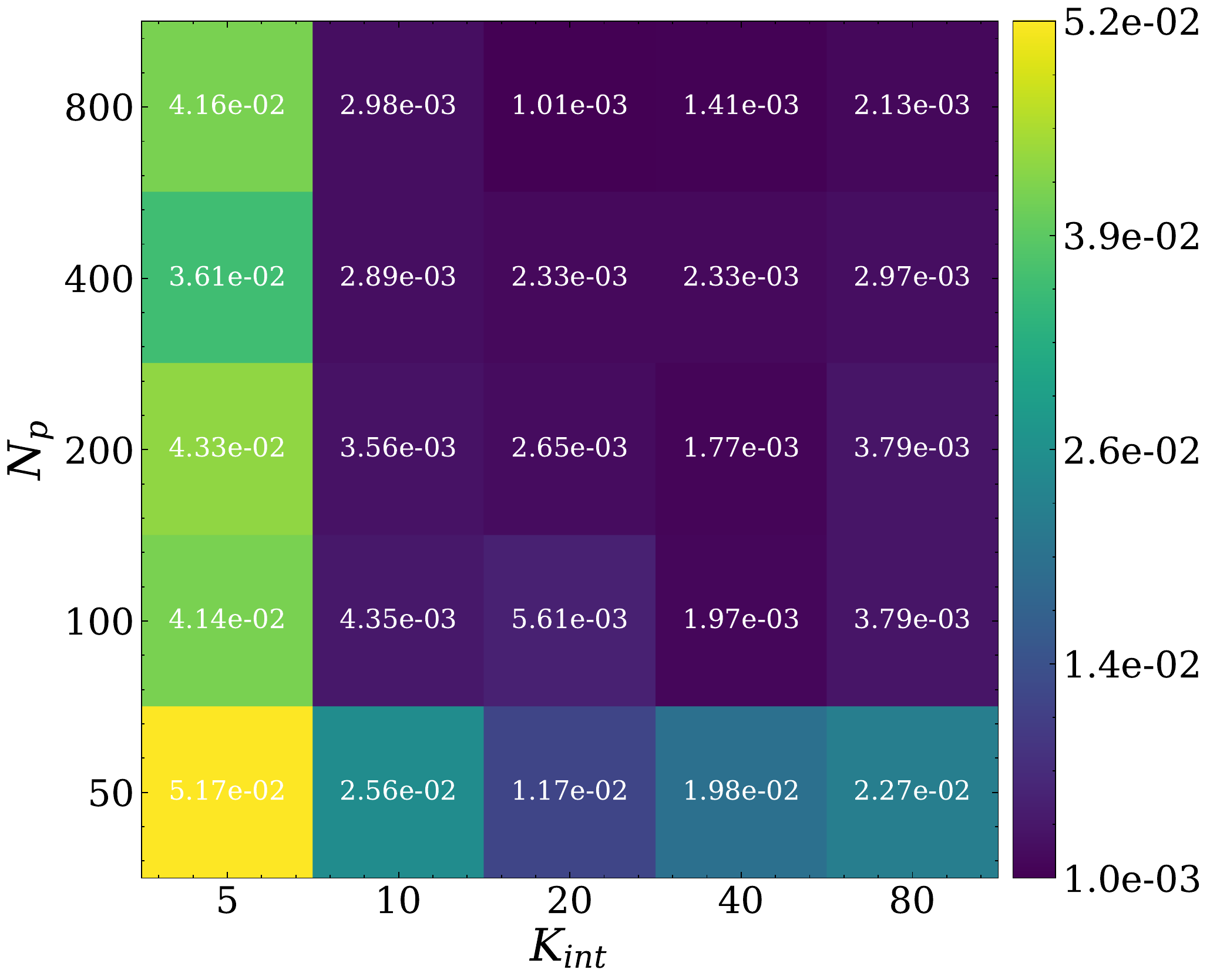}}
    \subfigure[Time(s) vs. $N_p K_{int}$]{\label{fig:ablation_time}
        \includegraphics[width=0.3\textwidth]{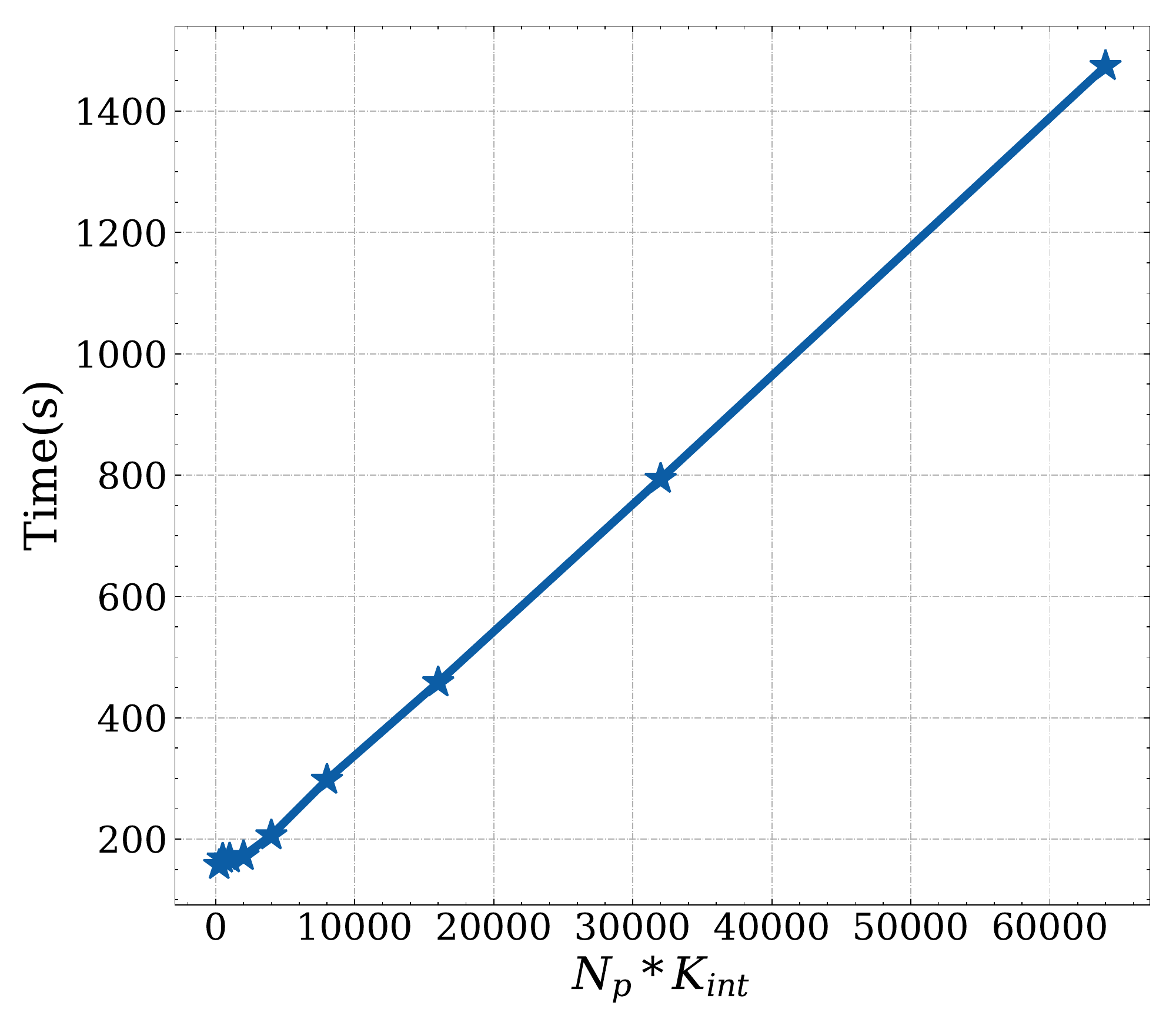}}
    \vspace{-0.25cm}
    \caption{Performance of the ParticleWNN with different combinations of $N_p$ and $K_{int}$.
    (a) Average Relative errors; (b) Average MAEs; (c) Average computation times vs. $N_p K_{int}$.} 
    \label{fig:ablation_Np_Kint}
\end{figure}

For weak-form DNN-based methods, the choice of the number of integration points and/or particles 
plays a pivotal role in algorithm performance. 
This decision involves finding the delicate balance between maximizing accuracy and 
minimizing computational effort when utilizing weak-form DNN-based methods for solving PDEs. 
Therefore, in this section, we delve into the impact of varying $N_p$ and $K_{int}$ 
on the ParticleWNN method.
To conduct this investigation, we apply the ParticleWNN approach to solve 
the Poisson problem \eqref{eq:poisson_1d} with $\omega=15\pi$. 
We set Since as the activation function and 
explore different combinations of $N_p$ and $K_{int}$. 
Experimental results are presented in Figure \ref{fig:ablation_Np_Kint}, and detailed metrics 
such as relative errors, MAEs, and computation times can be found in Appendix \ref{sec:app_ablation}.
From Figures \ref{fig:Np_Kint_l2} to \ref{fig:Np_Kint_abs}, we observe that, 
within the context of our problem settings, achieving a satisfactory level of accuracy 
requires a minimum of $100$ particles and a minimum of $10$ integration points 
(summing up to $1000$ points). 
This number of points is significantly lower than what other methods typically demand under 
similar conditions (as discussed in Section \ref{sec:poisson1d}). 
However, it is important to note that the required number of particles and integration points 
may vary for different problems and parameter settings. 
For instance, problems with lower solution regularity may necessitate more points, and vice versa.

In summary, our ablation study indicates that increasing the number of points leads to higher accuracy, 
albeit with a linear increase in computation time (as illustrated in Figure \ref{fig:ablation_time}). 
Moreover, both the number of particles and integration points should not be set too small 
for effective performance.
\section{Conclusion}
\label{sec:conclusion}
In this work, we propose a novel framework, ParticleWNN, for solving PDEs in weak-form with DNNs. 
The novelty of the framework lies in the use of test functions compactly supported in extremely small regions. 
The sizes and locations of these support regions can be selected arbitrarily. In this way, the method 
not only inherits the advantages of weak-form methods such as requiring less regularity of the solution 
and a small number of quadrature points for computing the integrals, but also outperforms them in solving 
complicated solution problems. Additionally, the flexible definition of the test function makes ParticleWNN 
easily applicable for solving high-dimensional problems and complex domain problems. To improve the 
training of the ParticleWNN, several training strategies are developed, such as the R-descending strategy 
and adaptive selection of particles. Finally, we demonstrate the advantages of the ParticleWNN over other 
state-of-the-art methods in solving PDE problems and inverse problems.

While ParticleWNN avoids integrals over the entire domain or subdomains, it still encounters 
limitations imposed by integral calculations, such as errors caused by integral approximations. 
Therefore, an important future direction is to develop more efficient integral calculation techniques. 
Another future direction is to investigate new techniques to improve the ParticleWNN method, such as 
smarter particle selection rules \cite{tang2023pinns,gao2023failure}, 
more efficient training strategies \cite{krishnapriyan2021characterizing,wang2022respecting}, 
and adaptive weighting strategies \cite{wang2020understanding,van2022optimally}.
%
\section*{Acknowledgement}
We are grateful to Qian Huang for providing valuable feedbacks on the draft. 
The research of GB was supported in part by a National Natual Science of China grant (No. U21A20425) 
and a Key Laboratory of Zhejiang Province.
\bibliographystyle{elsarticle-num}
\bibliography{ParticleWNN_jcp}
\newpage
\begin{appendices}
\section{The effect of the R-adaptive training strategy on the algorithm}
\label{sec:app_R_adaptive}
\begin{figure}[!htbp]
    \centering  
    \subfigure[Relative error]{\label{fig:poisson2d_R_l2}
        \includegraphics[width=0.425\textwidth]{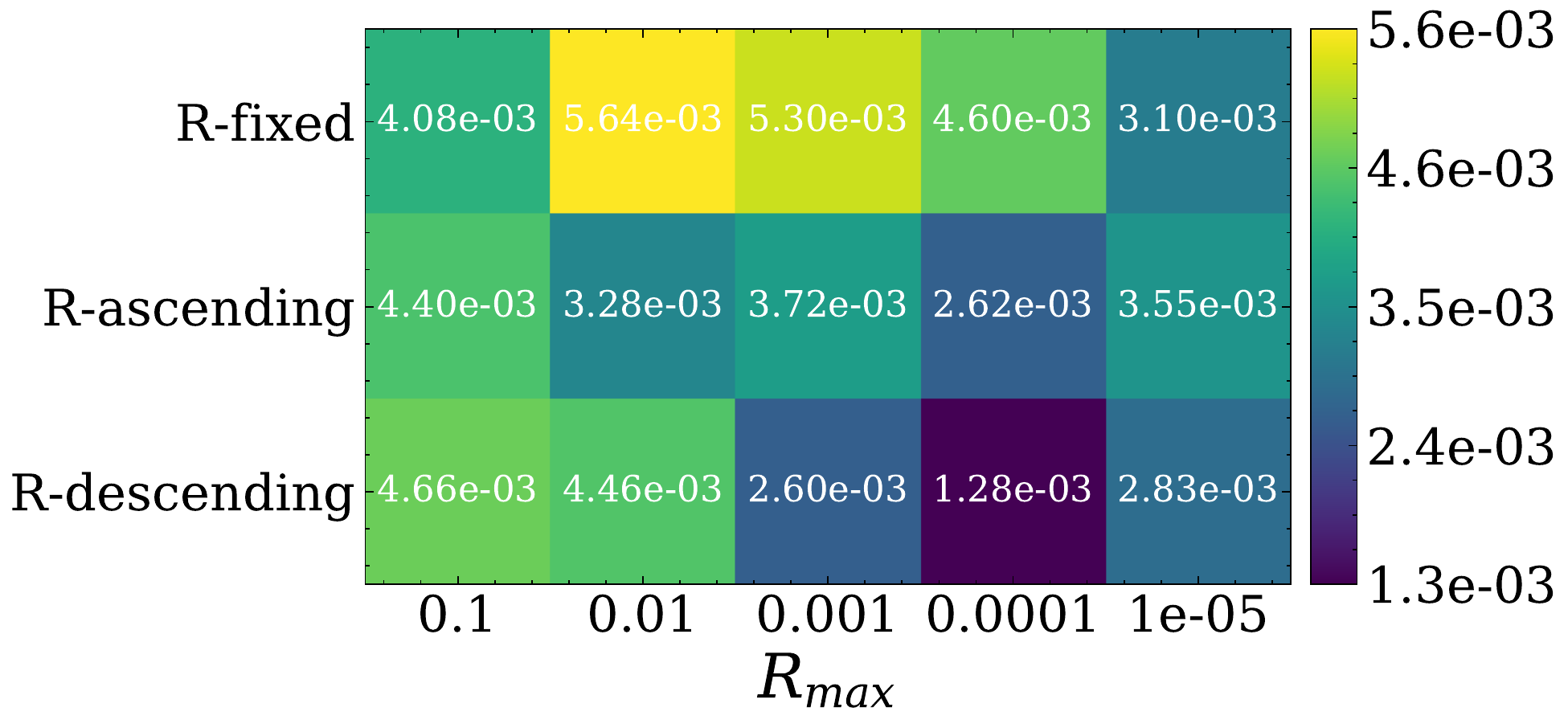}}
    \subfigure[MAE]{\label{fig:poisson2d_R_abs}
        \includegraphics[width=0.45\textwidth]{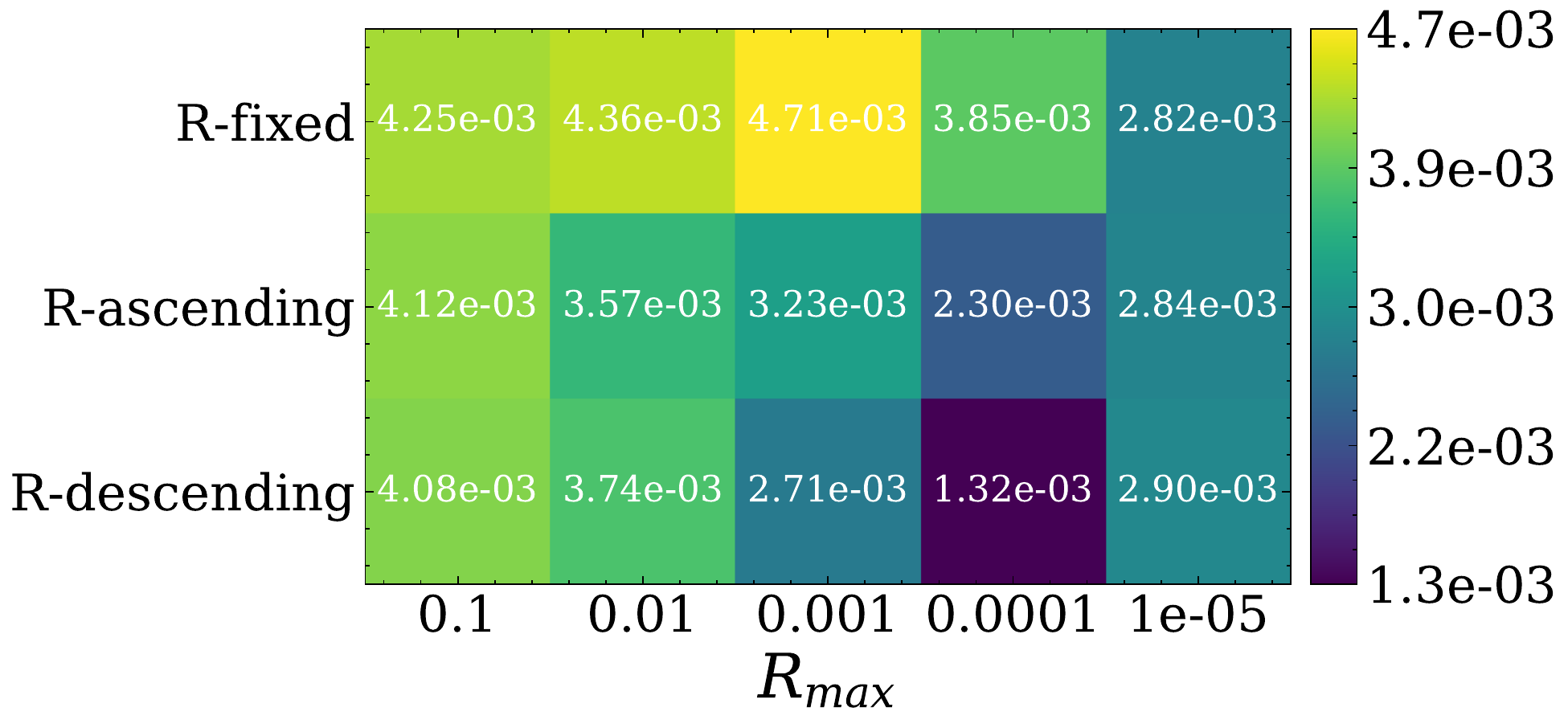}}
    \vspace{-0.25cm}
    \caption{The performance of the ParticleWNN under different R-adaptive strategies 
    and $R_{max}$.
    (a) Average Relative errors; (b) Average MAEs.} 
    \label{fig:poisson1d_R}
\end{figure}

To determine the most effective R-adaptive training strategy, we solve 
the Poisson problem with frequency $\omega=15\pi$ in Section \ref{sec:poisson1d} 
using ParticleWNN with different R-adaptive strategies.
For each strategy, we fix $R_{min}=10^{-6}$, and then investigate the impact of 
varying $R_{max}$ on the algorithm.
The activation function is selected as Since and the remaining experimental settings 
are consistent with those in Section \ref{sec:poisson1d}.
We record the Relative errors and MAEs obtained by the ParticleWNN 
in Table \ref{tab:poisson1d_Rmax_l2} and 
Table \ref{tab:poisson1d_Rmax_mae}, respectively.
To gain a more intuitive understanding of the impact of the R-adaptive strategy 
and $R_{max}$ on ParticleWNN, we visualize the Relative errors and the MAEs in 
Figure \ref{fig:poisson2d_R_l2} and Figure \ref{fig:poisson2d_R_abs}, respectively.
From \ref{fig:poisson2d_R_l2} and \ref{fig:poisson2d_R_abs}, we can see that the ParticleWNN 
performs best with the R-descending strategy.
Particularly, when the R-descending strategy is applied, ParticleWNN achieves the smallest 
average Relative error and MAE at $R_{max}=10^{-4}$.
\begin{table}[!ht]\small
  \centering
  \caption{Relative errors obtained by the ParticleWNN under different combinations of R-adaptive strategy and $R_{max}$.}
  \begin{tabular}{c|ccccc} \bottomrule
          {} & \multicolumn{5}{c}{$R_{max}$} \\\cline{2-6}
          {} & {$1e^{-1}$} & {$1e^{-2}$} & {$1e^{-3}$} & {$1e^{-4}$} & {$1e^{-5}$} \\ \hline
      {R-fixed} & {$4.1e^{-3}\pm 1.5e^{-3}$}  & {$5.6e^{-3}\pm 5.6e^{-3}$} & {$5.3e^{-3}\pm 3.1e^{-3}$}  & {$4.6e^{-3}\pm 4.6e^{-3}$} & {$3.1e^{-3}\pm1.2e^{-3}$}  \\ 
      {R-ascending} & {$4.4e^{-3}\pm 3.0e^{-3}$}  & {$3.3e^{-3}\pm 1.6e^{-3}$} & {$3.7e^{-3}\pm3.0e^{-3}$} & {$2.6e^{-3}\pm2.6e^{-3}$}  & {$3.5e^{-3}\pm2.0e^{-3}$}  \\ 
      {R-descending} & {$4.6e^{-3}\pm 1.8e^{-3}$}  & {$4.5e^{-3}\pm 4.2e^{-3}$}   & {$2.6e^{-3}\pm1.6e^{-3}$}& {$1.3e^{-3}\pm1.1e^{-3}$} & {$2.8e^{-3}\pm2.3e^{-3}$} \\  \toprule
\end{tabular}
\label{tab:poisson1d_Rmax_l2}
\end{table}
\begin{table}[!ht]\small
  \centering
  \caption{MAEs obtained by the ParticleWNN under different combinations of R-adaptive strategy and $R_{max}$.}
  \begin{tabular}{c|ccccc} \bottomrule
          {} & \multicolumn{5}{c}{$R_{max}$} \\\cline{2-6}
          {} & {$1e^{-1}$} & {$1e^{-2}$} & {$1e^{-3}$} & {$1e^{-4}$} & {$1e^{-5}$} \\ \hline
      {R-fixed} & {$4.3e^{-3}\pm1.3e^{-3}$}  & {$4.4e^{-3}\pm3.1e^{-3}$} & {$4.7e^{-3}\pm2.2e^{-3}$}  & {$3.8e^{-3}\pm3.7e^{-3}$} & {$2.8e^{-3}\pm1.1e^{-3}$}  \\ 
      {R-ascending} & {$4.1e^{-3}\pm2.0e^{-3}$}  & {$3.6e^{-3}\pm2.3e^{-3}$} & {$3.2e^{-3}\pm2.4e^{-3}$}  & {$2.3e^{-3}\pm2.0e^{-3}$} & {$2.8e^{-3}\pm1.2e^{-3}$}  \\ 
      {R-descending} & {$4.1e^{-3}\pm1.7e^{-3}$}  & {$3.7e^{-3}\pm2.8e^{-3}$} & {$2.7e^{-3}\pm1.7e^{-3}$}  & {$1.3e^{-3}\pm9.8e^{-4}$} & {$2.9e^{-3}\pm2.4e^{-3}$} \\  \toprule
\end{tabular}
\label{tab:poisson1d_Rmax_mae}
\end{table}
\section{The effect of the adaptive particle selection technique on the algorithm}
\label{sec:app_topK}
\begin{figure}[!htbp]
  \centering  
  \subfigure[Relative error]{\label{fig:ablation_topk_l2}
      \includegraphics[width=0.3\textwidth]{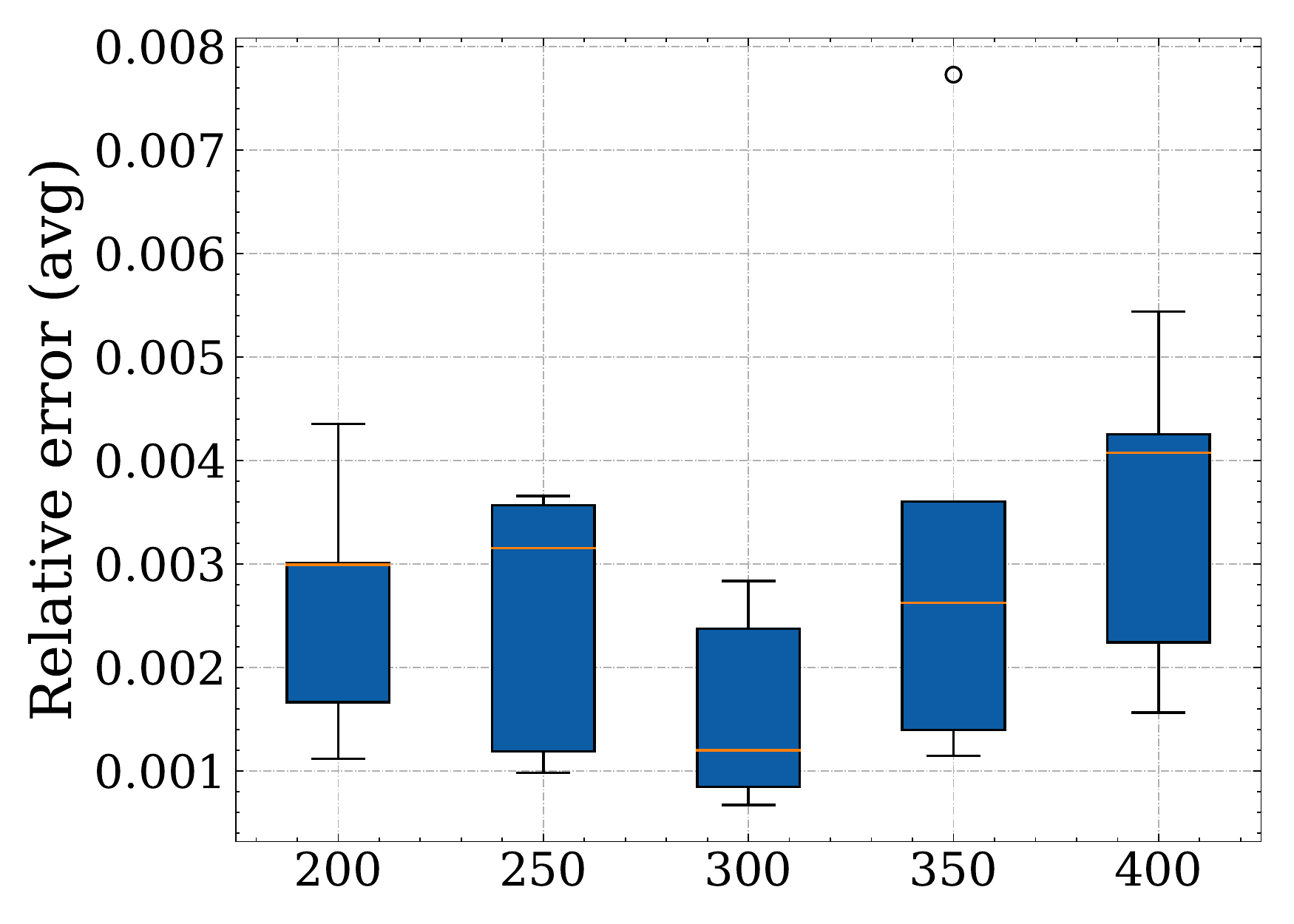}}
  \subfigure[MAE]{\label{fig:ablation_topk_abs}
      \includegraphics[width=0.3\textwidth]{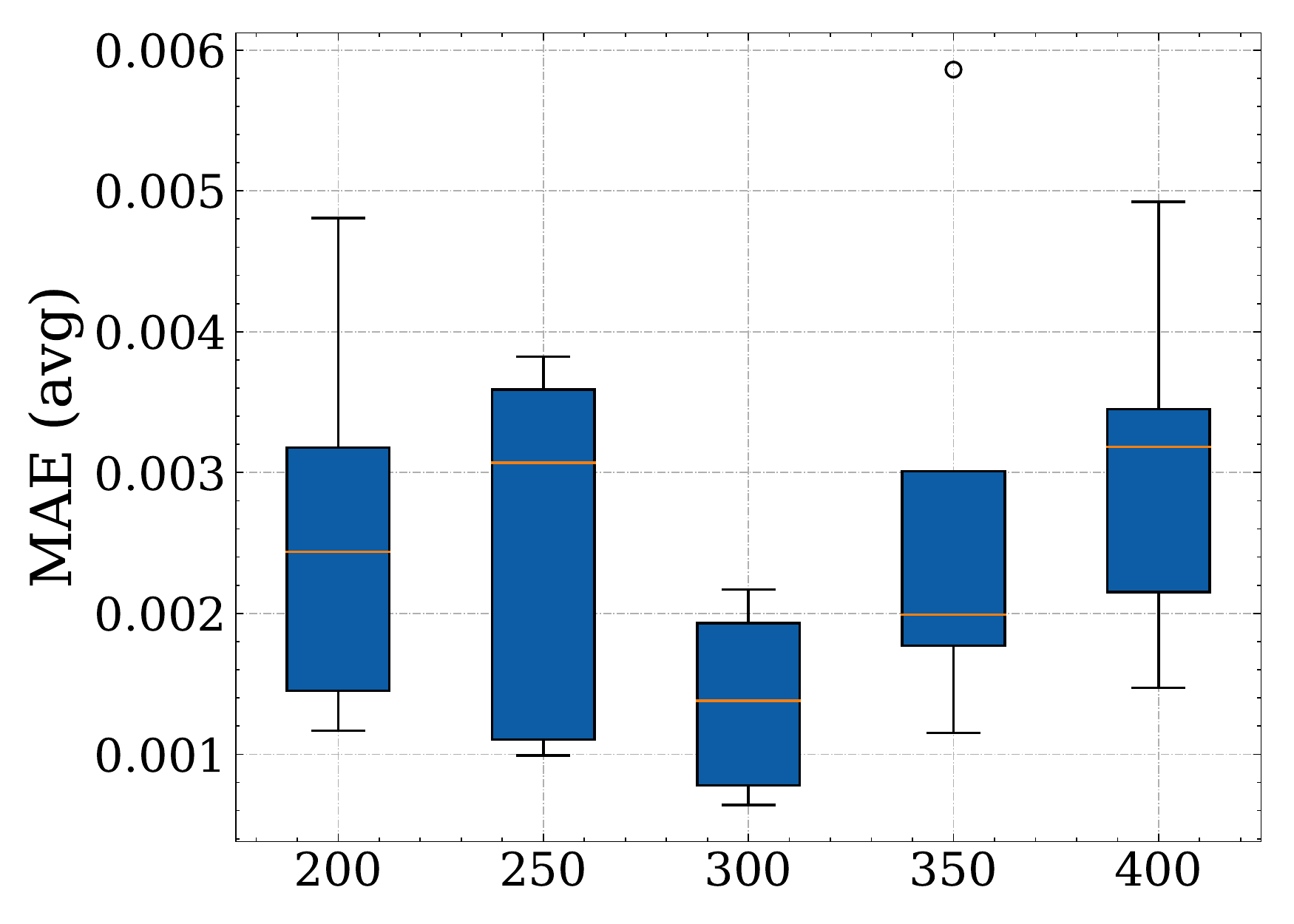}}
  \subfigure[Relative error vs. time]{\label{fig:ablation_topk_time}
      \includegraphics[width=0.3\textwidth]{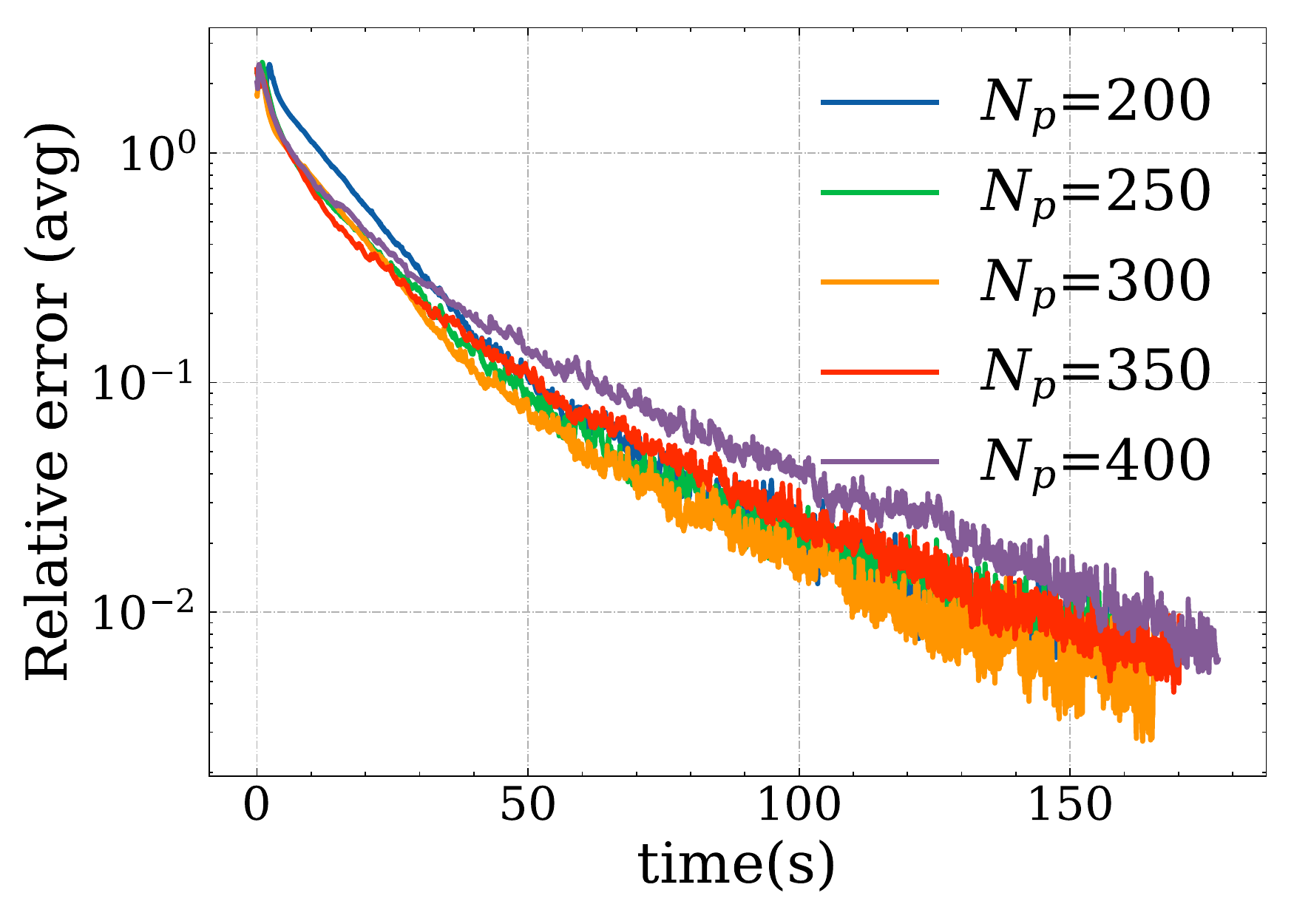}}
  \vspace{-0.25cm}
  \caption{Performance of the ParticleWNN under different $N_p$ ($topK=200$).
  (a) Average Relative errors; 
  (b) Average MAEs.
  (c) Average Relative error vs. Average computation times.} 
  \label{fig:ablation_topk}
\end{figure}

In this section, we examine the effect of the $topK$ adaptive particle selection technique on 
the ParticleWNN approach.
We set $topK=200$ and vary $N_p$ between $200, 250, 300, 350$ and $400$, 
and then employ ParticleWNN to solve the Poisson problem \eqref{eq:poisson_1d} with frequency $\omega=15\pi$
in Section \ref{sec:poisson1d}.
We select the activation as Since and keep other parameters consistent with 
those used in Section \ref{sec:poisson1d}.
The experimental results are illustrated in Figure \ref{fig:ablation_topk}, and the Relative errors and MAEs are 
recorded in Table \ref{tab:poisson1d_topK}.
From Figure \ref{fig:ablation_topk}, we can see that a suitable selection of $N_p$ (e.g., $N_p=300$) 
with $topK=200$ has a positive effect on the algorithm's convergence and accuracy.
When $N_P\leq 300$ ($topK=200$), Figure \ref{fig:ablation_topk_time} shows that the $topK$ technique obviously 
improves the convergence and accuracy of the algorithm.
\begin{table}[!ht]\small
  \centering
  \caption{Relative errors and MAEs obtained by the ParticleWNN under different $N_p$ ($topK=200$).}
  \begin{tabular}{c|ccccc} \bottomrule
          {} & {$N_p$ =200} & {$N_p$ =250} & {$N_p$ =300} & {$N_p$ =350} & {$N_p$ =400} \\ \hline
      {Relative error} & {$2.6e^{-3}\pm1.1e^{-3}$}  & {$2.5e^{-3}\pm1.2e^{-3}$} & {$1.6e^{-3}\pm8.6e^{-4}$}  & {$3.3e^{-3}\pm2.4e^{-3}$} & {$3.5e^{-3}\pm1.4e^{-3}$}  \\ 
      {MAE} & {$2.6e^{-3}\pm1.3e^{-3}$}  & {$2.5e^{-3}\pm1.2e^{-3}$} & {$1.4e^{-3}\pm6.1e^{-4}$}  & {$2.7e^{-3}\pm1.7e^{-3}$} & {$3.0e^{-3}\pm1.2e^{-3}$}  \\ \toprule
\end{tabular}
\label{tab:poisson1d_topK}
\end{table}
\section{Additional ablations}
\label{sec:app_ablation}
\begin{figure}[!htbp]
  \centering  
  \subfigure[Relative errors]{\label{fig:Np_Kint_l2_new}
      \includegraphics[width=0.32\textwidth]{./ablation_Np_Kint_l2.pdf}}
  \subfigure[MAEs]{\label{fig:Np_Kint_abs_new}
      \includegraphics[width=0.32\textwidth]{./ablation_Np_Kint_abs.pdf}}
  \subfigure[Times(s)]{\label{fig:Np_Kint_time_new}
      \includegraphics[width=0.32\textwidth]{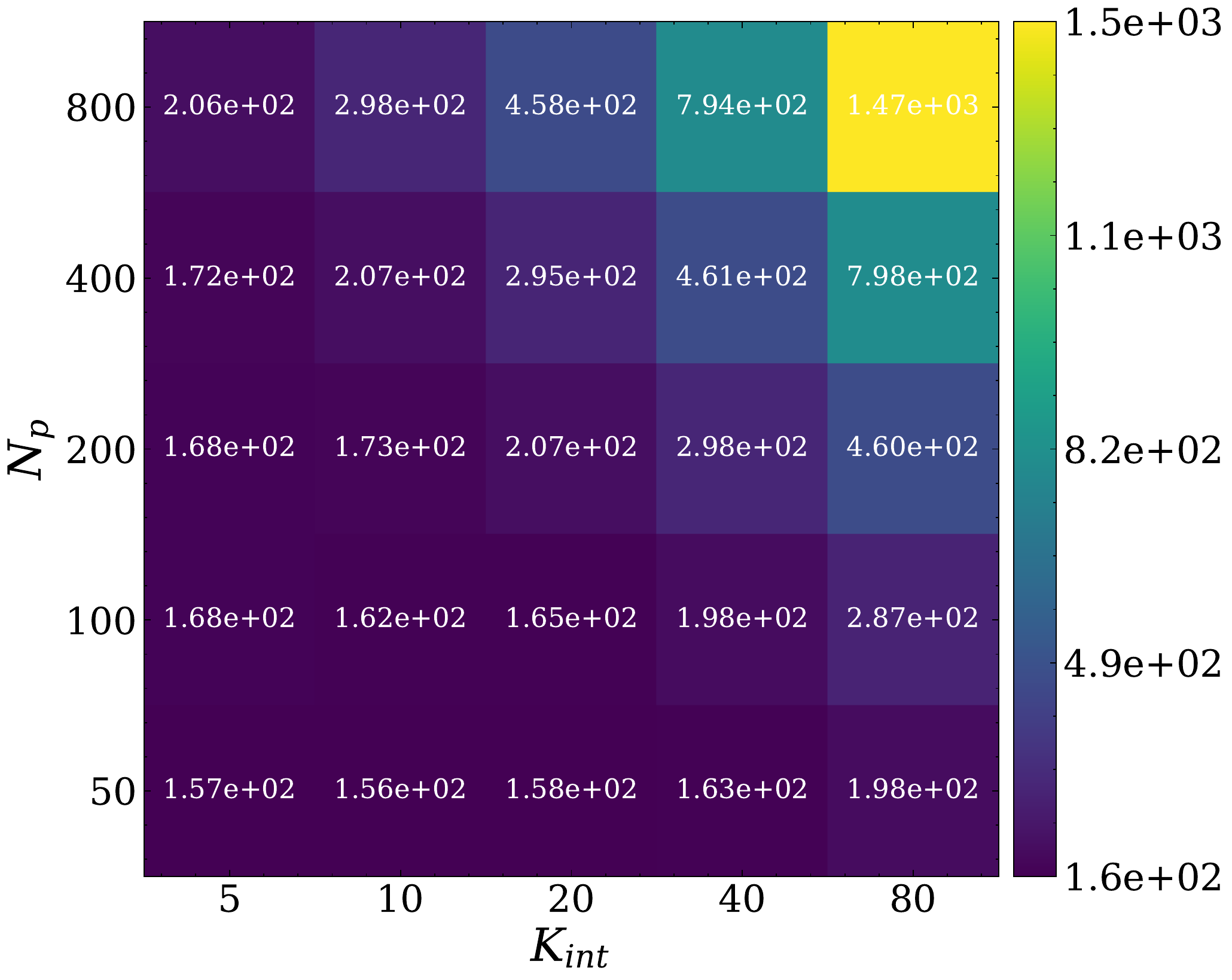}}
  \vspace{-0.25cm}
  \caption{Performance of the ParticleWNN with different combinations of $N_p$ and $K_{int}$.
  (a) Average Relative errors; (b) Average MAEs; (c) Average computation times.} 
  \label{fig:poisson1d_Np_Kint_new}
\end{figure}

The Relative errors, MAEs, and time consumptions obtained by the ParticleWNN 
in Section \ref{sec:add_ablation} are recorded in Table 
\ref{tab:Np_Kint_l2_error}, Table \ref{tab:Np_Kint_mae_error}, and Table \ref{tab:Np_Kint_time}, 
respectively.
Additionally, the visual representations of the Relative errors, MAEs, and computation times 
are provided in Figures \ref{fig:Np_Kint_l2_new}, \ref{fig:Np_Kint_abs_new}, and \ref{fig:Np_Kint_time_new}, respectively.

\begin{table}[!ht]\small
  \centering
  \caption{Relative errors obtained by the ParticleWNN under different combinations of $N_p$ and $K_{int}$.}
  \begin{tabular}{c|ccccc} \bottomrule
    {\diagbox{$N_p$}{$K_{int}$}} & {$5$} & {$10$} & {$20$} & {$40$} & {$80$} \\ \hline
{$800$} & {$3.7e^{-2}\pm3.4e^{-3}$}  & {$4.1e^{-3}\pm4.5e^{-3}$} & {$1.1e^{-3}\pm 0.3e^{-3}$}  & {$1.8e^{-3}\pm1.0e^{-3}$} & {$2.9e^{-3}\pm1.1e^{-3}$}  \\ 
{$400$} & {$3.2e^{-2}\pm5.2e^{-3}$}  & {$3.5e^{-3}\pm2.8e^{-3}$} & {$3.0e^{-3}\pm3.4e^{-3}$}  & {$2.7e^{-3}\pm1.9e^{-3}$} & {$3.5e^{-3}\pm4.2e^{-3}$}  \\ 
{$200$} & {$3.8e^{-2}\pm3.8e^{-3}$}  & {$2.9e^{-3}\pm1.3e^{-3}$} & {$3.7e^{-3}\pm2.1e^{-3}$}  & {$1.7e^{-3}\pm0.5e^{-4}$} & {$3.9e^{-3}\pm2.4e^{-3}$} \\ 
{$100$} & {$3.4e^{-2}\pm3.2e^{-3}$}  & {$4.7e^{-3}\pm3.7e^{-3}$} & {$6.1e^{-3}\pm3.2e^{-3}$}  & {$2.0e^{-3}\pm0.4e^{-4}$} & {$3.5e^{-3}\pm1.9e^{-3}$} \\ 
{$50$} & {$5.4e^{-2}\pm1.4e^{-2}$}  & {$3.3e^{-2}\pm2.1e^{-2}$} & {$1.3e^{-2}\pm5.6e^{-3}$}  & {$2.6e^{-2}\pm1.8e^{-2}$} & {$2.8e^{-2}\pm2.4e^{-2}$}   \\  \toprule
\end{tabular}
\label{tab:Np_Kint_l2_error}
\end{table}
\begin{table}[!ht]\small
  \centering
  \caption{MAEs obtained by the ParticleWNN under different combinations of $N_p$ and $K_{int}$.}
  \begin{tabular}{c|ccccc} \bottomrule
    {\diagbox{$N_p$}{$K_{int}$}} & {$5$} & {$10$} & {$20$} & {$40$} & {$80$} \\ \hline
{$800$} & {$4.2e^{-2}\pm 2.7e^{-3}$}  & {$2.9e^{-3}\pm 2.3e^{-3}$} & {$1.0e^{-3}\pm 4.7e^{-4}$}  & {$1.4e^{-3}\pm 5.3e^{-4}$} & {$2.1e^{-3}\pm 7.3e^{-4}$}  \\ 
{$400$} & {$3.6e^{-2}\pm 6.7e^{-3}$}  & {$2.8e^{-3}\pm 1.6e^{-3}$} & {$2.3e^{-3}\pm 2.1e^{-3}$}  & {$2.3e^{-3}\pm 1.2e^{-3}$} & {$2.9e^{-3}\pm 2.6e^{-3}$}  \\ 
{$200$} & {$4.3e^{-2}\pm 2.3e^{-3}$}  & {$3.5e^{-3}\pm 2.1e^{-3}$} & {$2.6e^{-3}\pm 1.3e^{-3}$}  & {$1.7e^{-3}\pm 5.4e^{-4}$} & {$3.8e^{-3}\pm 2.0e^{-3}$} \\ 
{$100$} & {$4.1e^{-2}\pm 3.1e^{-3}$}  & {$4.3e^{-3}\pm 2.9e^{-3}$} & {$5.6e^{-3}\pm 2.6e^{-3}$}  & {$1.9e^{-3}\pm 4.1e^{-4}$} & {$3.8e^{-3}\pm 2.2e^{-3}$} \\ 
{$50$} & {$5.2e^{-2}\pm 5.3e^{-3}$}  & {$2.5e^{-2}\pm 1.5e^{-2}$} & {$1.2e^{-2}\pm 4.2e^{-3}$}  & {$1.9e^{-2}\pm 1.6e^{-2}$} & {$2.3e^{-2}\pm 1.5e^{-2}$}   \\  \toprule
\end{tabular}
\label{tab:Np_Kint_mae_error}
\end{table}
\begin{table}[!ht]\small
  \centering
  \caption{Time consumptions of the ParticleWNN under different combinations of $N_p$ and $K_{int}$.}
  \begin{tabular}{c|ccccc} \bottomrule
    {\diagbox{$N_p$}{$K_{int}$}} &{$5$} & {$10$} & {$20$} & {$40$} & {$80$} \\ \hline
{$800$} & {$206.2\pm5.7$}  & {$297.7\pm3.2$} & {$458.3\pm2.6$}  & {$793.8\pm11.0$} & {$1474.2\pm12.2$}  \\ 
{$400$} & {$172.1\pm12.6$}  & {$206.5\pm13.1$} & {$294.6\pm7.9$}  & {$461.0\pm11.7$} & {$789.3\pm9.8$}  \\ 
{$200$} & {$168.3\pm10.6$}  & {$173.0\pm12.5$} & {$207.0\pm9.2$}  & {$297.8\pm9.7$} & {$460.2\pm7.5$} \\ 
{$100$} & {$167.9\pm8.5$}  & {$161.9\pm4.9$} & {$165.4\pm3.5$}  & {$198.2\pm3.2$} & {$287.2\pm2.0$} \\ 
{$50$}  & {$157.4\pm4.7$}  & {$156.0\pm3.2$} & {$157.6\pm3.9$}  & {$162.6\pm3.5$} & {$197.6\pm2.9$}   \\  \toprule
\end{tabular}
\label{tab:Np_Kint_time}
\end{table}
\end{appendices}

\end{document}